\documentclass[10pt,twocolumn,letterpaper]{article}

\usepackage[pagenumbers]{cvpr} 

\usepackage[accsupp]{axessibility}
\usepackage{graphicx}
\usepackage{amsmath}
\usepackage{amssymb}
\usepackage{booktabs}

\usepackage[pagebackref,breaklinks,colorlinks]{hyperref}

\usepackage{stackengine}
\usepackage{scalerel}

\usepackage{cite}
\usepackage{pifont}

\usepackage{empheq}
\usepackage{multirow}
\usepackage[table]{xcolor}
\usepackage[percent]{overpic}
\usepackage[numbers]{natbib}

\usepackage[capitalize]{cleveref}
\crefname{section}{Sec.}{Secs.}
\Crefname{section}{Section}{Sections}
\Crefname{table}{Table}{Tables}
\crefname{table}{Tab.}{Tabs.}

\def\cvprPaperID{8490} 
\def\confName{CVPR}
\def\confYear{2023}

\newcommand{\Paragraph}[1]{\vspace{0.6mm} \noindent \textbf{#1}
\hspace{0mm}}
\newcommand{\topvspace}{\vspace{0mm}}

\newcommand{\firstkey}[1]{\textcolor{red}{\textbf{#1}}}
\newcommand{\secondkey}[1]{\textcolor{blue}{\textbf{#1}}}

\begin{document}

\title{PMatch: Paired Masked Image Modeling for Dense Geometric Matching}

\author{%
  Shengjie Zhu and Xiaoming Liu \\
  Department of Computer Science and Engineering, \\
  Michigan State University, East Lansing, MI, 48824 \\
  zhusheng@msu.edu, liuxm@cse.msu.edu
}
\maketitle

\begin{abstract}
Dense geometric matching determines the dense pixel-wise correspondence between a source and support image corresponding to the same 3D structure.
Prior works employ an encoder of transformer blocks to correlate the two-frame features.
However, existing monocular pretraining tasks, \textit{e.g.}, image classification, and masked image modeling (MIM), can not pretrain the cross-frame module, yielding less optimal performance.
To resolve this, we reformulate the MIM from reconstructing a single masked image to reconstructing a pair of masked images, enabling the pretraining of transformer module.
Additionally, we incorporate a decoder into pretraining for improved upsampling results.
Further, to be robust to the textureless area, we propose a novel cross-frame global matching module (CFGM).
Since the most textureless area is planar surfaces, we propose a homography loss to further regularize its learning.
Combined together, we achieve the State-of-The-Art (SoTA) performance on geometric matching.
Codes and models are available at \href{https://github.com/ShngJZ/PMatch}{https://github.com/ShngJZ/PMatch}.

\end{abstract}

\section{Introduction}
\label{sec:intro}
When a 3D structure is viewed in both a source and a support image, for a pixel (or keypoint) in the source image, the task of geometric matching identifies its corresponding pixel in the support image.
This task is a cornerstone for many downstream vision applications, {\it e.g.}~homography estimation~\cite{dubrofsky2009homography}, structure-from-motion~\cite{schonberger2016structure}, visual odometry estimation~\cite{engel2017direct} and visual camera localization~\cite{brahmbhatt2018geometry}.

\begin{figure}[t!]
  \captionsetup{font=small}
  \centering
  \includegraphics[width=1.0\linewidth]{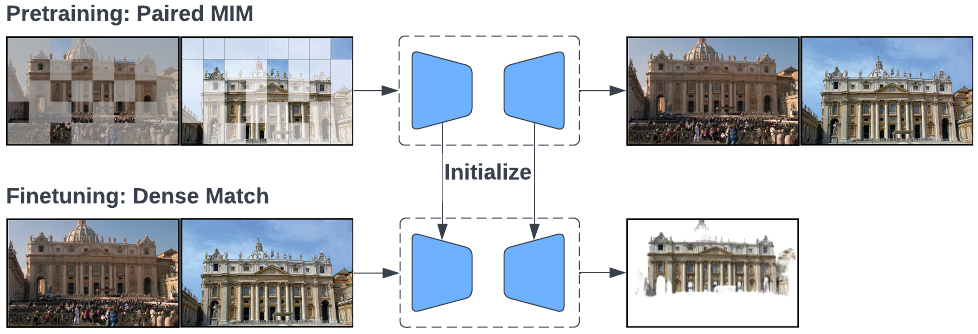}
  \caption{\small 
  Most vision tasks start with a pretrained network.
  In geometric matching, the unique network components processing two-view features cannot benefit from the monocular pretraining task, \textit{e.g.}, image classification, and masked image modeling (MIM).
  As in the figure, this work enables the pretraining of a matching model via reformulating MIM from reconstructing a single masked image to reconstructing a pair of masked images.
  \vspace{-4mm}}
  \label{fig:teaser}
\end{figure}

There exist both sparse and dense methods for geometric matching.
The sparse methods~\cite{dusmanu2019d2, revaud2019r2d2, tyszkiewicz2020disk, lowe2004distinctive, detone2018superpoint, rocco2020efficient, liu2022drc, sun2021loftr, sun2021loftr} only yield correspondence on sparse or semi-dense locations while the dense methods~\cite{truong2021learning, truong2021pdc, edstedt2023dkm} estimate pixel-wise correspondence.
They primarily differ in that the sparse methods embed a keypoint detection or a global matching on discrete coordinates, which underlyingly assumes a unique mapping between source and support frames.
Yet, the existence of textureless surfaces introduces multiple similar local patches,
disabling keypoint detection or causing ambiguous matching results. 
Dense methods, though facing similar challenges at the coarse level, alleviate it with the additional fine-level local context and smoothness constraint.
Until recently, the dense methods demonstrate a comparable or better geometric matching performance over the sparse methods~\cite{truong2021learning,truong2021pdc, edstedt2023dkm}.

A relevant task to dense geometric matching is the optical flow estimation~\cite{teed2020raft}.
Both tasks estimate dense correspondences, whereas the optical flow is applied over consecutive frames with the constant brightness assumption.

In geometric matching~\cite{sun2021loftr, chen2022aspanformer}, apart from the encoder encodes source and support frames into feature maps, there exist transformer blocks which correlate two-frame features, \textit{e.g.}, the LoFTR module~\cite{sun2021loftr}.
Since these network components consume two-frame inputs, the monocular pretraining task, \textit{e.g.}, the image classification and masked image modeling (MIM) defined on ImageNet dataset, is unable to benefit the network.
This limits both the geometric matching performance and its generalization capability.

To address this, we reformulate the MIM from single masked image reconstruction to paired masked images reconstruction, \textit{i.e.}, pMIM.
Paired MIM benefits the geometric matching as both tasks rely on the cross-frame module to correlate two frames inputs for prediction.

With a pretrained encoder, the decoder in dense geometric matching is still randomly initialized.
Following the idea of pretraining encoder, we extend pMIM pretraining to the decoder.
As part functionality of decoder is to upsample the coarse-scale initial prediction to the same resolution as input, we also task the decoder in pMIM to upsample the coarse-scale reconstruction to its original resolution.
Correspondingly, we consist the decoder as stacks of the depth-wise convolution except for the last prediction head.
With the depth-wise decoder, when transferring from pMIM to geometric matching, we duplicate the decoder along the channel dimension to finish the initialization.
To this end, there exists only a small number of components in the decoder randomly initialized, we pretrain the rest network components using synthetic image pair augmentation~\cite{truong2021pdc}.

To further improve the dense geometric matching performance, we propose a cross-frame global matching module (CFGM).
In CFGM, we first compute the correlation volume.
We model the correspondences of coarse scale pixels as a summation over the discrete coordinates in the support frame, weighted by the softmaxed correlation vector.
However, this modeling fails when multiple similar local patches exit.
As a solution, we impose positional embeddings to the discrete coordinates and decode with a deep architecture to avoid ambiguity.
Meanwhile, we notice that the textureless surfaces are mostly planar structures described by a low-dimensional $8$ degree-of-freedom (DoF) homography matrix.
We thus design a homography loss to augment the learning of the low DoF planar prior.

\begin{figure*}[h!]
  \captionsetup{font=small}
  \centering
  \vspace{8mm}
  \includegraphics[width=\linewidth]{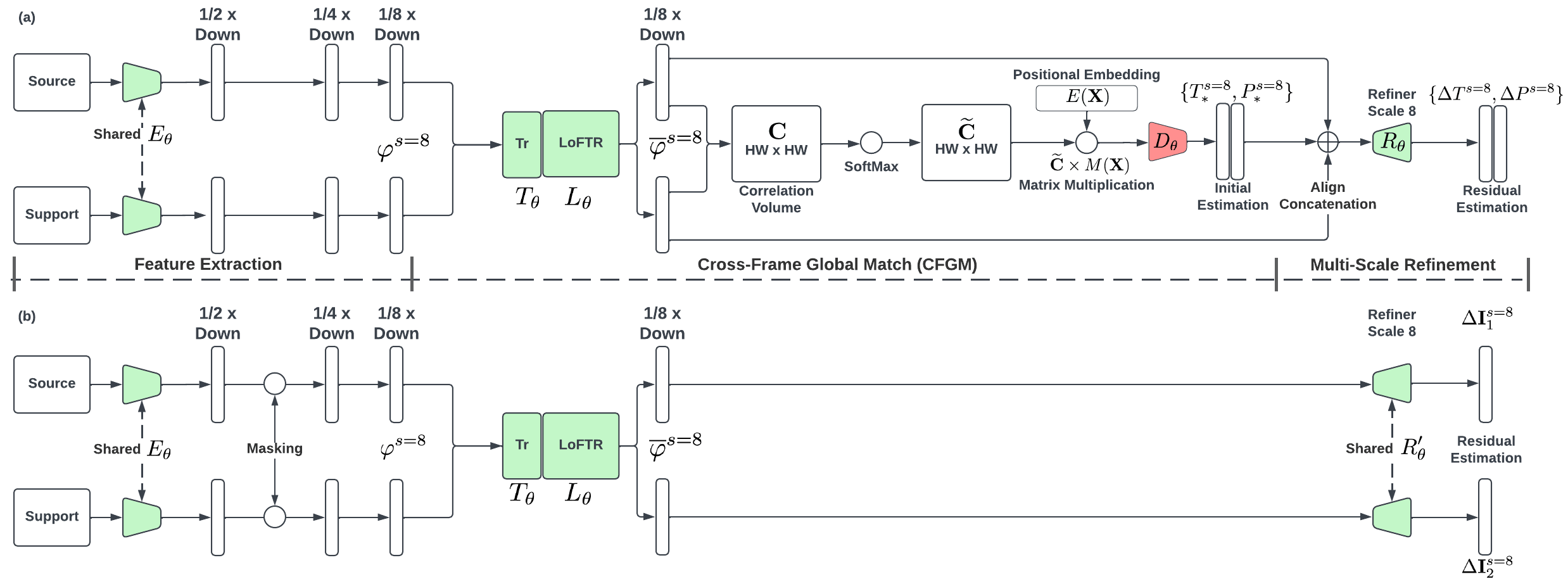}
  \vspace{-4mm}
  \caption{\small 
  \textbf{Methodology Overview.}
  In (a), we illustrate the proposed dense geometric matching network.
  After extracting the multi-scale feature with the encoder $E_\theta$, we extend the LoFTR module with (1) Transformer blocks $T_\theta$ and (2) positional embeddings with an appended decoder $D_\theta$ to remove the ambiguity when multiple local patches exist.
  In (b), we show the proposed paired MIM pretext task.
  We apply image masking at the scale $s=2$, and recover the masked images with the transformer blocks.
  In (a), network $D_\theta $ (in \textcolor{red}{red}) is not included in pMIM pretraining.
  In dense matching, $R_\theta$ takes in the stack of source and the aligned support frame feature.
  In the pretext task, $R_\theta'$ only takes in the source frame feature. 
  Thus, $R_\theta'$ is a sub-graph of $R_\theta$.
  We detail how to initialize $R_\theta$ using $R_\theta'$ in Fig.~\ref{fig:sepconv}.
  The residual refinement at other scales repeats the process at scale $s = 8$ but consumes feature embeddings of other scales, skipped for simplicity.
  \vspace{-3mm}  }
  \label{fig:framework}
\end{figure*}

We summarize our contributions as follows:

\noindent $\bullet$
We introduce the paired masked image modeling pretext task, pretraining both the encoder and decoder of a dense geometric matching network.

\noindent $\bullet$
We propose a novel cross-frame global matching module that is robust to textureless local patches. 
Since the most textureless patches are planar structures, we augment their learning with a homography loss.

\noindent $\bullet$
We outperform dense and sparse geometric matching methods on diverse datasets.

\section{Related works}
\subsection{Pretraining and Finetuning}
Pretraining and finetuning is an effective paradigm in vision tasks.
Supervised image classification has been one of the most widely adopted pretraining methods.
An encoder~\cite{he2016deep, simonyan2014very, huang2017densely}, \textit{e.g.}, ResNet~\cite{he2016deep}, together with a few fully connected (FC) layers is trained for image classification using a large-scale dataset, \textit{e.g.}, ImageNet~\cite{imagenet_cvpr09}.
After converging, the encoder is used as the initialization in the downstream vision tasks. 

Apart from supervised classification tasks, there are self-supervised methods producing discriminative feature representation.
Inspired by BYOL~\cite{grill2020bootstrap}, DINO~\cite{caron2021emerging} introduces a self-supervised mean-teacher knowledge distillation task.
It encourages the prediction consistency between a student and teacher model where the teacher is an exponential moving average of the student model.
The pretrained ViT model embeds explicit information of semantic segmentation, which is not observed in a supervised counterpart.
Other self-supervised pretraining methods include color transformation~\cite{chen2020simple}, geometric transformation~\cite{chen2020simple}, Jigsaw Puzzle~\cite{misra2020self}, feature frame prediction~\cite{oord2018representation}, \textit{etc}.

Among the self-supervised learning tasks, masked image modeling (MIM)~\cite{vincent2010stacked, xiao2021early, bao2021beit, zhou2021ibot, yang2021instance, he2022masked} achieves SoTA finetuning performance on ImageNet~\cite{imagenet_cvpr09}.
The task introduces Masked Language Modeling used in NLP domain to vision, reconstructing an image from its masked input.
While iGPT~\cite{chen2020generative}, ViT~\cite{dosovitskiy2020image}, and BEiT~\cite{bao2021beit} adopt sophisticated paradigm in modeling, MAE~\cite{he2022masked} and SimMIM~\cite{xie2022simmim} show that directly regressing the masked continuous RGB pixels can achieve competitive results.
Typically, they focus on pretraining the encoder, adopting an asymmetric design where only a shallow decoder head is appended.

In this paper, we reformulate MIM from reconstructing a single image to the paired images, reducing the domain gap between the pretexting task and the downstream geometric matching.
As a result, we extend the benefit of MIM pretraining to the task of dense geometric matching.

\subsection{Sparse Geometric Matching}
There are detector-based and detector-free sparse geometric matching methods.
Classic works are detector based, and employ the nearest neighbor (NN) match using the hand-crafted feature on detected keypoints, \textit{e.g.}, SIFT~\cite{lowe2004distinctive}, SURF~\cite{bay2008speeded}, and ORB~\cite{rublee2011orb}.
Both keypoint detection and feature extraction are improved by data-driven deep models~\cite{detone2018superpoint, dusmanu2019d2, ono2018lf, revaud2019r2d2, yi2016lift, detone2018superpoint}.
Later, \cite{sarlin2020superglue, rocco2020efficient, tyszkiewicz2020disk} propose to replace the naive NN match by graph neural network based differentiable matching.

While the detector based methods operate on keypoints, the detector free methods, \textit{e.g.}~LoFTR~\cite{sun2021loftr} and ASpanFormer~\cite{chen2022aspanformer} operate all-to-all matching on coarse-scale discrete grid locations.
Still, their matching depends on the correlation between features, yielding ambiguous results when multiple local patches exist.
We improve LoFTR from two perspectives. 
First, we extend the LoFTR module to the proposed cross-frame global matching module to benefit from the MIM pretexting task.
Second, we alleviate the ambiguity caused by similar local patches by imposing positional embeddings over the low-dimensional 2D coordinates.
A decoder is then employed to resolve the ambiguity.

\subsection{Dense Geometric Matching}
DGC-Net~\cite{melekhov2019dgc} regresses dense correspondences from a global correlation volume at a limited resolution.
GLU-Net~\cite{truong2020glu} increases the resolution with a global-local correlation layer.
GOCor~\cite{truong2020gocor} further improves GLU-Net~\cite{truong2020glu} by replacing the correlation layer with online optimization.
Other methods, such as RANSAC Flow~\cite{shen2020ransac}, iteratively recover a homography transformation to reduce the visual difference between the source and support images.

Though dense methods estimate more correspondences than sparse methods, it is less favored for geometric matching.
Until recently, PDC Net+~\cite{truong2021pdc} and DKM~\cite{edstedt2023dkm} close the gap between dense and sparse methods.
Both methods model the dense match as probability functions.
PDC Net+ adopts a mixture Laplacian distribution while DKM models with the Gaussian Process (GP).
Furthermore, they estimate a confidence score to remove false positive results.
We follow \cite{truong2021pdc, edstedt2023dkm} in the confidence estimation.
However, instead of applying probabilistic regression, we keep the correlation based explicit matching process.
This saves the computation of the inverse matrix required in the GP Regression of DKM.
Also, we apply a unique architecture design to benefit from the MIM pretexting task.

\section{Method}

In this section, we first introduce the proposed dense geometric matching method.
Then we discuss how to pretext the network via the paired masked image modeling.
Fig.~\ref{fig:framework} depicts our framework in finetuning and pretexting stages.

\subsection{Dense Geometric Matching}
Dense geometric matching computes the dense correspondences between the source image $\mathbf{I}_1$ and support image $\mathbf{I}_2$.
Under the estimated correspondences $T$, source image $\mathbf{I}_1$ can be recovered from support image $\mathbf{I}_2$ by applying bilinear sampling at $T$. 
Since the dense correspondences between $\mathbf{I}_1$ and $\mathbf{I}_2$ is not guaranteed to exist at each pixel location, we follow ~\cite{edstedt2023dkm} in estimating confidence $P$ to indicate the fidelity of the prediction. 

\Paragraph{Feature Extraction.}
As shown in Fig.~\ref{fig:framework}, we adopt a multi-scale ResNet-based~\cite{he2016deep} feature extractor $E_\theta$.
Taking the source frame $\mathbf{I}_1$ as an example, we produce the multiscale feature embeddings as:
\begin{equation}
    \{\varphi_1^{s=2}, \varphi_1^{s=4}, \varphi_1^{s=8} \} = E_\theta (\mathbf{I}_1).
    \label{eqn:resnet}
\end{equation}
For the input image $\mathbf{I}_1$ of resolution $H \times W$, the scale $s$ indicates a feature map of resolution ${H}/{s} \times {W}/{s}$. 

\Paragraph{Cross-Frame Global Matching}
The cross-frame global matching module (CFGM) is designed to accomplish coarse-scale geometric matching.
To benefit from the MIM pretext task, we first process the scale $s = 8$ feature map $\varphi_1^{s=8}$ with the transformer block~\cite{katharopoulos2020transformers}:
\begin{equation}
    \{{\overline{\varphi}_1^{s=8}}', {\overline{\varphi}_2^{s=8}}' \} = T_\theta (\varphi_1^{s=8}, \varphi_2^{s=8}).
    \label{eqn:pvt}
\end{equation}
In the pretraining stage, the masked feature map is recovered by the appended transformer blocks.
Then, we follow LoFTR~\cite{sun2021loftr} in using linear transformer blocks to correlate the source and support frame feature:
\begin{equation}
    \{\overline{\varphi}_1^{s=8}, \overline{\varphi}_2^{s=8} \} = L_\theta ({\varphi_1^{s=8}}', {\varphi_2^{s=8}}').
    \label{eqn:loftr}
\end{equation}
To compute the global matching results, we first compute the 4D correlation volume $\mathbf{C}\left(\overline{\varphi}_1^{s=8}, \overline{\varphi}_2^{s=8} \right) \in \mathbb{R}^{H/8 \times W/8 \times H/8 \times W/8}$, where:
\begin{equation}
C_{i j k l}=\sum_{h} \frac{1}{\gamma} \left(\overline{\varphi}_1^{s=8}\right)_{i j h} \cdot \left(\overline{\varphi}_2^{s=8}\right)_{k l h},
\end{equation}
where $\gamma$ is a temperature scalar. 
The coarse matches are computed as a summation over pixel locations $\mathbf{X} \in \mathbb{R}^{(H/8) (W/8) \times 2}$ weighted by the softmaxed correlation volume. 
That is, after the correlation volume $\mathbf{C}$ being reshaped to $\mathbf{C} \in \mathbb{R}^{(H/8) (W/8) \times (H/8) (W/8)}$, we apply the softmax: 
\begin{equation}
\widetilde{C_{ij}} = \text{softmax}(C_{ij}).
\end{equation}
Here, element $C_{ij}$ is a size $(H/8) (W/8) \times 1$ vector. We conclude the coarse global matching results as:
\begin{equation}
    T^{s=8}_* = \widetilde{\mathbf{C}}  \times \mathbf{X}.
    \label{eqn:naive_match}
\end{equation}
Note, Eqn.~\ref{eqn:naive_match} will cause ambiguous results when multiple similar textureless local patches exist, \textit{i.e.}, multiple peak values in softmaxed correlation vector $\widetilde{C_{ij}}$.
To resolve this, we modify Eqn.~\ref{eqn:naive_match} with:
\begin{equation}
    T^{s=8}_*, P^{s=8}_* = D_\theta \left( \widetilde{\mathbf{C}}  \times M(\mathbf{X}) \right),
    \label{eqn:coarse_match}
\end{equation}
where $M (\mathbf{X})$ is cosine positional embeddings with learnable tokens~\cite{sun2021loftr, edstedt2023dkm}, projecting the 2D pixel locations to a high dimensional space to avoid ambiguity when multiple similar patches exist.
The decoder $D_\theta $ decodes $T^{s=8}_*$, initial correspondences estimation at scale $s= 8$, and  $P^{s=8}_*$, initial confidence estimation.

\begin{figure}[t!]
  \captionsetup{font=small}
  \centering
  \includegraphics[width=0.9\linewidth]{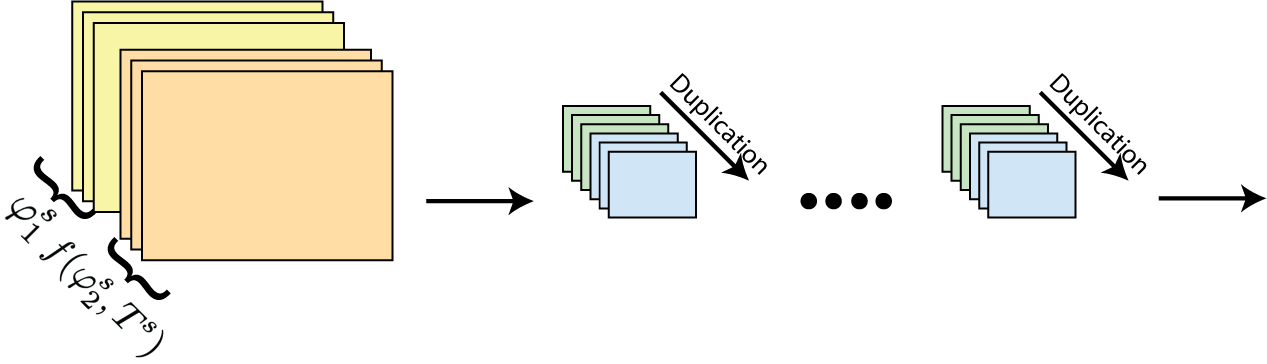}
  \vspace{-2mm}
  \caption{\small 
  \textbf{Resolution of the Discrepancy between $R_\theta$  and $R_\theta'$.}
  We adopt stacks of the depth-wise convolution in the refinement module, \textit{i.e.}, each convolution kernel only works with one channel of the input feature maps.
  This makes refiner $R_\theta'$ in pretexting a sub-graph of refiner $R_\theta$ in finetuning.
  While transferring from the pretexting task to finetuning task, the input feature map concatenates an extra aligned support frame feature $f(\varphi_2^s, T^s)$. 
  As the bilinear sampling $f$ imposes minimal distribution change, we duplicate the kernel weight along the channel dimension.
  \vspace{-3mm} 
  }
  \label{fig:sepconv}
\end{figure}

\Paragraph{Multi-Scale Refinement}
We follow~\cite{edstedt2023dkm} in using the multi-scale refinement module:
\begin{equation}
    \Delta T^{s}, \Delta P^{s} = R_\theta (\varphi_1^s, f(\varphi_2^s, T^{s})),
    \label{eqn:flowrefinement}
\end{equation}
where function $f(\cdot)$ indicates the bilinear interpolation to align the support frame feature using the current estimated correspondences $T^{s}$, shown in Fig.~\ref{fig:framework}.
To accommodate the transfer between pretexting and finetuning stage, we apply depth-wise convolution~\cite{edstedt2023dkm} in $R_\theta$. 
We detail the discussion in Fig.~\ref{fig:sepconv} and Sec.\ref{sec:MIM}.
The correspondences and confidence on the next scale are initialized with the bilinear upsampling.

\subsection{Paired MIM Pretraining}
\label{sec:MIM}
\begin{figure*}[t!]
    \centering
    \topvspace
    \subfloat{\includegraphics[height=3.9cm]{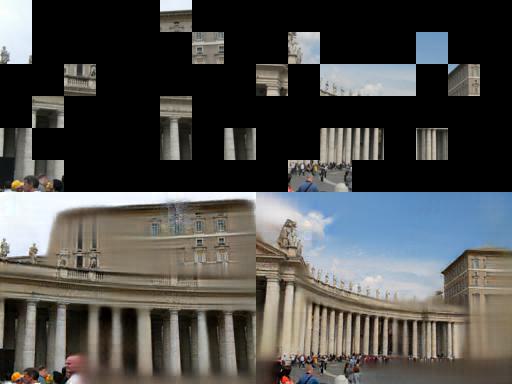}} \,
    \subfloat{\includegraphics[height=3.9cm]{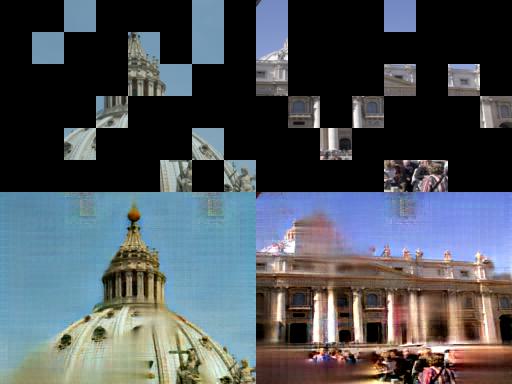}}\,
    \subfloat{\includegraphics[height=3.9cm]{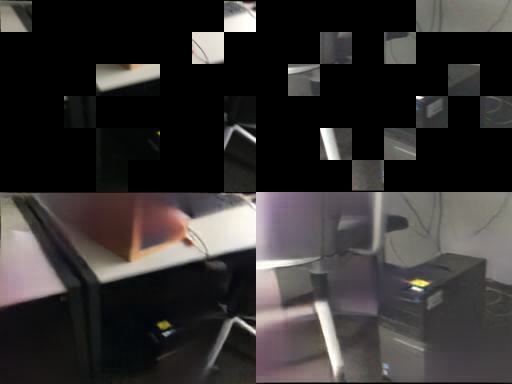}}
    \vspace{0mm}
    \caption{\small 
    \textbf{Visual Quality of the paired MIM pretext task.}
    Visualized cases are from the MegaDepth and the ScanNet dataset.}
    \vspace{-2mm}
    \label{fig:pretext_vls}
\end{figure*}

\Paragraph{Paired Masked Image Modeling (MIM)} 
MIM is extensively adopted in image classification task~\cite{he2022masked, xie2022simmim}. 
An image classification network can be further improved after MIM pretexting.
As shown in Fig.~\ref{fig:teaser} and ~\ref{fig:pretext_vls}, the network reconstructs the input from randomly masked feature embeddings at a specific scale.
In this work, we investigate the benefit of pretraining both the encoder and decoder under MIM.
Compared to only pretraining the encoder, pretraining the whole network further reduces the domain gap between pretexting and finetuning tasks.

\Paragraph{Masking Strategy} 
We follow SimMIM~\cite{xie2022simmim} in using randomly selected $32 \times 32$ mask patches with a predefined masking ratio $r_1$ and $r_2$ for source and support frames.
For source view, given the feature embeddings $\varphi_1^{s=2}$ output by the extractor $E_\theta$ at scale $s = 2$, we apply the randomly generated mask $\mathbf{w}$ to mask out the feature embeddings, {\it i.e.}:
\begin{equation}
    {\varphi_1^{s=2}}' = \varphi_1^{s=2} * (1 - \mathbf{w}) + \mathbf{x} * \mathbf{w},
\end{equation}
where $\mathbf{x}$ is the learnable mask tokens. Note, our extractor $E_\theta$  starts from a $3 \times 3$ convolution kernel to avoid leakage of the masked patches.

\Paragraph{Prediction Heads}
Different from SimMIM~\cite{xie2022simmim}, our prediction heads include most network components of the decoder.
We complete the masked feature embeddings with the transformer as:
\begin{equation}
    {\varphi_1^{s=8}}' = T_\theta ({\varphi_1^{s=8}}).
    \label{eqn:pvt_mim}
\end{equation}
Here, we use the same notation as Eqn.~$\ref{eqn:pvt}$ since both indicate image features at the scale $s = 8$.
Note that the subsequent network component LoFTR is a series of linear transformer blocks~\cite{katharopoulos2020transformers} which reduce the quadratic computational complexity to linear.
However, empirically we find the linear transformer poorly recovers the masked patches.
We thus append the transformer blocks.

As shown in Fig.~\ref{fig:framework}, after Eqn.~\ref{eqn:pvt_mim}, we feed the completed feature map to CFGM.
Note the refiner between the two stages is different.
Instead of taking a stacked feature map (Eqn.~\ref{eqn:flowrefinement}), in pretexting we only take in a single feature map:
\begin{equation}
     \Delta \mathbf{I}_1^s =  R_\theta' (\varphi_1^s), \quad \Delta \mathbf{I}_2^s = R_\theta' (\varphi_2^s).
    \label{eqn:rgbrefinement}
\end{equation}
To account for the difference between Eqn.~\ref{eqn:flowrefinement} and Eqn.~\ref{eqn:rgbrefinement}, we apply depth-wise convolution, where each convolution kernel operates on one channel of the feature map, shown in Fig.~\ref{fig:sepconv}.
Since $f(\varphi_2^s, T^{s})$ in Eqn.~\ref{eqn:flowrefinement} is a resampled support frame feature, it imposes minimal distribution difference to $\varphi_2^s$.
Then, while transferring from the pretexting task to the downstream task, we only need to duplicate the channel of $R_\theta$ to complete the initialization.
We follow SimMIM~\cite{xie2022simmim} in estimating full resolution residual RGB images in each scale of the decoder.
We visualize the reconstructed paired masked images in Fig.~\ref{fig:pretext_vls}.

\Paragraph{Network Components not included in pMIM}
Since the feature map at $s = 2$ contains little information about masked patches, the pretraining only includes refinement modules at scale $s = 4$ and $s = 8$. 
Furthermore, the CFGM decoder $D_\theta$ and part of $R_\theta$ are not included.
We pretrain the rest network component with synthetic image pairs~\cite{truong2021pdc}.

\Paragraph{Prediction Objective}
Set the accumulated reconstruction at each scale $s$ as $\mathbf{I}^{s}$, we regress the raw pixel value with an $l_1$ loss:
\begin{equation}
    \mathcal{L}_{M}= \sum_s \frac{1}{N}(| \mathbf{I}_1^{s} - \mathbf{I}_1 |_{1} + | \mathbf{I}_2^{s} - \mathbf{I}_2 |_{1}),
\end{equation}
where $N$ is the number of unmasked pixels.


\subsection{Dense Geometric Matching Loss }
\Paragraph{Homography Loss}
The image correspondences between two planar structures are constrained by a $3 \times 3$ homography matrix $\mathbf{H}$ with $8$ DoF.
Compared to correspondences estimation over arbitrary shapes, the correspondences in planar structures possess a lower rank.
Given a surface normal $\mathbf{n}$ computed using the depth gradient~\cite{nakagawa2015estimating}, the homography of the pixel can be computed as:
\begin{equation}
    \mathbf{H} = \begin{bmatrix} \mathbf{h}_1^\intercal \\ \mathbf{h}_2^\intercal \\ \mathbf{h}_3^\intercal \end{bmatrix} = \mathbf{K}_1\left(\mathbf{R}+\frac{\mathbf{t}^{\top}}{d} \mathbf{n}\right)\mathbf{K}_2^{-1},
\end{equation}
where the $\mathbf{K}_1$ and $\mathbf{K}_2$ are intrinsic matrices of $\mathbf{I}_1$ and $\mathbf{I}_2$, $\mathbf{R}$ and $\mathbf{t}$ are camera rotation and translation, and $d$ is the pixel depth. We randomly sample $K$ anchor points $\{\mathbf{p}_{m} \mid 1\leq m \leq K \}$. For each anchor point $\mathbf{p}_{m}$, we sample $K$ candidate points $\{\mathbf{q}_{n}^m \mid 1\leq n \leq K \}$. 
We determine a co-planar indicator matrix $\mathcal{O}^+$ of size $K \times K$ to suggest all co-planar pairs. 
We use the normal consistency, point-to-plane distance, and homography consistency to compute the co-planar groundtruth, detailed in Supp.
Finally, we apply a gradient-based penalty, penalizing the correspondences difference between the estimation and the groundtruth. 
\begin{equation}
    \mathcal{L}_h^s = \frac{1}{|\mathcal{O}^+|} \sum_{\mathcal{O}^+_{\mathbf{p}, \mathbf{q}} = 1} | \left(T^s_\mathbf{p} - T^s_\mathbf{q}\right) - \left(\overline{T}_{\mathbf{p}}^{s} - \overline{T}_{\mathbf{q}}^{s}\right) |_1.
\end{equation}



\Paragraph{Global Matching Loss}
Following \cite{sun2021loftr}, we minimize a binary cross-entropy loss over the correlation volume $\mathbf{C}$ after a dual-softmax operation:
\begin{equation}
    \widetilde{C_{ijkl}}' = \text{softmax}(C_{ij}) \cdot \text{softmax}(C_{kl}),
\end{equation}
where $C_{ij}$ and $C_{kl}$ are $(H/8)(W/8) \times 1$ vectors. The loss is defined as:
\begin{equation}
\small
\begin{aligned}
\mathcal{L}_{g}=&-\frac{1}{\left|\mathcal{M}^{+}\right|} \sum_{ijkl \in \mathcal{M}^{+}} \log \widetilde{C_{ijkl}}' \\ &-\frac{1}{\left|\mathcal{M}^{-}\right|} \sum_{ijkl \in \mathcal{M}^{-}} \log \left(1 - \widetilde{C_{ijkl}}'\right),
\end{aligned}
\end{equation}
where $\mathcal{M}^{+}$ and $\mathcal{M}^{-}$ are groundtruth indicator matrix of size $H \times W \times H \times W$ indicating whether a source frame pixel $(i,j)$ pairs with a target frame pixel $(k, l)$.

\Paragraph{Refinement Loss}
Following~\cite{edstedt2023dkm}, we supervise both correspondences and confidence on each scale of the predictions, 
\begin{equation}
    \mathcal{L}_r^s =\frac{1}{\left|P^+ \right|}\sum_{ij \in P^+} \left| T_{ij}^s - \overline{T}_{ij}^{s} \right|_{2},
\end{equation}
where $P^+_{ij}$ is a $H \times W$ matrix that indicates whether a valid pair is found at pixel location $ij$ in the source frame.
Similarly, the loss of confidence is defined as:
\begin{equation}
    \mathcal{L}_c^s =- \frac{1}{\left|\mathcal{P}^{+}\right|}\sum_{ij \in P^+} \log (P_{ij}) - \frac{1}{\left|\mathcal{P}^{-}\right|}\sum_{ij \in P^-} \log (1 - P_{ij}).
\end{equation}

\Paragraph{Total Loss}
The total loss is a weighted summation of proposed losses:
\begin{equation}
    \mathcal{L} = \frac{1}{4} \sum_s ({L}_r^s + w_c \mathcal{L}_c^s) + w_g\cdot  \mathcal{L}_{g} +  \frac{1}{4} w_h  \sum_s\mathcal{L}_{h}^s.
\end{equation}
The constant $4$ comes from the four scales $s = \{1, 2, 4, 8\}$ set in our paper.

\section{Experiments}
We first compare with other SoTA dense matching methods on the MegaDepth dataset.
Then, to comprehensively reflect the contributions from both the density and accuracy of geometric matching, we follow~\cite{sun2021loftr, edstedt2023dkm} in using the two-view relative camera pose estimation performance as the metric.
We report on both the outdoor scenario MegaDepth~\cite{li2018megadepth} dataset and the indoor scenario ScanNet~\cite{dai2017scannet} dataset. 
We additionally evaluate on the HPatches~\cite{balntas2017hpatches} and the YFCC100m~\cite{thomee2016yfcc100m} datasets to demonstrate the generalizability of the model.

\subsection{Implementation Details}
\Paragraph{Pretext stage} 
From DeMoN~\cite{ummenhofer2017demon}, BlendedMVS~\cite{yao2020blendedmvs}, HyperSim~\cite{roberts2021hypersim}, ARKitScenes~\cite{baruch2021arkitscenes}, and TartanAir~\cite{wang2020tartanair} datasets, we collect a pretraining dataset of $1,281,167$ image pairs, \textit{i.e.}, the same size as ImageNet~\cite{imagenet_cvpr09}.
Each pair is collected with a fixed frame index interval.
In the pretraining dataset, we train the model using a batchsize of $128$ under the resolution $192 \times 256$.
We use the Adam optimizer~\cite{kingma2014adam} with a learning rate $2 e^{-4}$, running for $250$k steps on $2 \times$ A100 GPUs.
We stack $1$ transformer layer.
We initialize the masking ratio $r_1 = 75\%$ and $r_2 = 75 \%$. 
The masking operation applies to the ResNet, causing significantly different batch statistics between masked and unmasked inputs.
Since the downstream task takes the unmasked image, we linearly reduce the support frame masking ratio $r_2$ to $0$ and use a different batch normalization layer for support view, resolving the batch statistics difference.
We also apply the synthetic image pair augmentation introduced in \cite{truong2021pdc}.

\Paragraph{Finetuning stage} 
Our model trains with a batchsize of $16$ at the resolution $544 \times 720$. 
The learning rate is set to $4 e^{-4}$, running $250$k steps with a warmup of $25$k steps.
On $4 \times$ A100 GPUs, we train for $5$ days with the Adam optimizer.
We follow~\cite{sun2021loftr} in sampling the paired images, weighted by the sequence length and overlap ratio. 
The softmax temperature $\gamma$ is  $0.1$.
We set loss weight $w_g$ to 0.7 and $w_h$ to $0.02$. 
We sample $600 \times 600$ points for homography loss $L_h$.

\subsection{Datasets}
\Paragraph{MegaDepth}
MegaDepth~\cite{li2018megadepth} collects over $10$ thousand images of worldwide landmarks from the Internet. 
The collected images are processed by COLMAP~\cite{schonberger2016structure} to produce groundtruth poses and depthmaps.
The dataset collects images of significant visual contrast due to lighting conditions, view angles, and imaging devices.
This imposes challenges to geometric matching.

\Paragraph{ScanNet~\cite{dai2017scannet}} is a large-scale indoor dataset with $1,613$ videos captured by RGB-D cameras.
There are challenging textureless indoor scenes for geometric matching.

\Paragraph{YFCC100m~\cite{thomee2016yfcc100m}} is a large multi-media dataset.
A subset of $72$ reconstructions of tourist landmarks is generated with groundtruth poses and depthmap.

\Paragraph{Hpatches~\cite{hpatches_2017_cvpr}} provides the pair of one source and five support images taken under different view angles and lighting conditions with groundtruth homography transformation.

\begin{table}[t!]
\centering
\captionsetup{font=small}
\resizebox{\linewidth}{!}{
\begin{tabular}{cccccc}
\hline
Methods & Venue & \multicolumn{3}{c}{Dense Match PCK $\uparrow$} & Run-      \\
& & @$1 \, \text{px}$     & @$3 \, \text{px}$    & @$5 \, \text{px}$   & time (ms) \\
\hline
RANSAC-FLow~\cite{shen2020ransac}  & ECCV'20 & $53.47$ & $83.45$ & $86.81$ & $3,596$\\
PDC-Net~\cite{zhang2019learning} & CVPR'21 & $71.81$ &$89.36$ & $91.18$ & $1,017$ \\
PDC-Net+~\cite{truong2021pdc} & Arxiv'21 & $\secondkey{74.51}$ & $\secondkey{90.69}$ & $\secondkey{92.10}$ & $1,017$ \\
\hline
LIFE~\cite{huang2021life} & Arxiv'21 & $39.98$ & $76.14$ & $83.14$  & $\secondkey{78}$\\
GLU-Net-GOCor~\cite{truong2020gocor} & NeurIPS'20 & $57.77$ & $78.61$ & $82.24$ &  $\firstkey{71}$\\
PDC-Net~\cite{zhang2019learning} & CVPR'21 & $68.95$ &$84.07$ & $85.72$ & $88$ \\
PDC-Net+~\cite{truong2021pdc} & Arxiv'21 & $72.41$ & $86.70$ & $88.12$ & $88$ \\
PMatch (Ours) & CVPR'23 & $\firstkey{79.83}$  & $\firstkey{95.18}$ & $\firstkey{96.52}$  & ${124}$ \\
\hline
\end{tabular}
}
\vspace{0mm}
\caption{\small \textbf{MegaDepth Dense Geometric Matching.} 
The running time of all methods is measured at the resolution $480 \times 480$. The upper and lower groups are methods running multiple or single times. [Key: \firstkey{Best}, \secondkey{Second Best}]
\label{tab:mega_dense}
}
\end{table}

\begin{table}[t!]
\centering
\captionsetup{font=small}
\resizebox{\linewidth}{!}{
\begin{tabular}[width=\linewidth]{cccccc}
\hline
\multicolumn{1}{c}{Category} &  \multicolumn{1}{c}{Methods} & \multicolumn{1}{c}{Venue} & \multicolumn{3}{l}{Pose Estimation AUC $\uparrow$} \\
& &   & @$5^\circ$           & @$10^\circ$          & @$20^\circ$ \\
\hline
Sparse  & SuperGlue~\cite{sarlin2020superglue} & CVPR'19 & $42.2$ & $61.2$ & $75.9$ \\
 W/ Detector & SGMNet~\cite{li2020sgm} & Pattern'20 & $40.5$ & $59.0$ & $72.6$ \\
\hline
 & DRC-Net~\cite{liu2022drc} & ICASSP'22 & $27.0$ & $42.9$ & $58.3$ \\
& LoFTR~\cite{sun2021loftr} & CVPR'21 & $52.8$ & $69.2$ & $81.2$ \\
Sparse & QuadTree~\cite{tang2022quadtree} & ICLR'22 & $54.6$ & $70.5$ & $82.2$ \\
Wo/ Detector & MatchFormer~\cite{wang2022matchformer} & ACCV'22 & $53.3$ & $69.7$ & $81.8$ \\
& ASpanFormer~\cite{chen2022aspanformer} & ECCV'22 & ${55.3}$ & ${71.5}$ & ${83.1}$ \\
\hline
\multirow{3}{*}{Dense} & PDC-Net+~\cite{truong2021pdc} & Arxiv'19 & $43.1$ & $61.9$ & $76.1$ \\
& DKM~\cite{edstedt2023dkm} & CVPR'23 & $\secondkey{60.5}$ & $\secondkey{74.9}$ & $\secondkey{85.1}$ \\
& PMatch (Ours) & CVPR'23 & $\firstkey{61.4}$ & $\firstkey{75.7}$ & $\firstkey{85.7}$      \\    
\hline
\end{tabular}
}
\vspace{0mm}
\caption{\small \textbf{MegaDepth Two-View Camera Pose Estimation.} 
We compare three groups of methods following SuperGlue~\cite{sarlin2020superglue} in evaluation.
The pose AUC error is reported. 
Our method shows substantial improvement.
[Key: \firstkey{Best}, \secondkey{Second Best}]
\label{tab:mega}
}
\end{table}

\begin{figure*}[t!]
    \centering
    \topvspace
    \subfloat{\includegraphics[height=2.0cm]{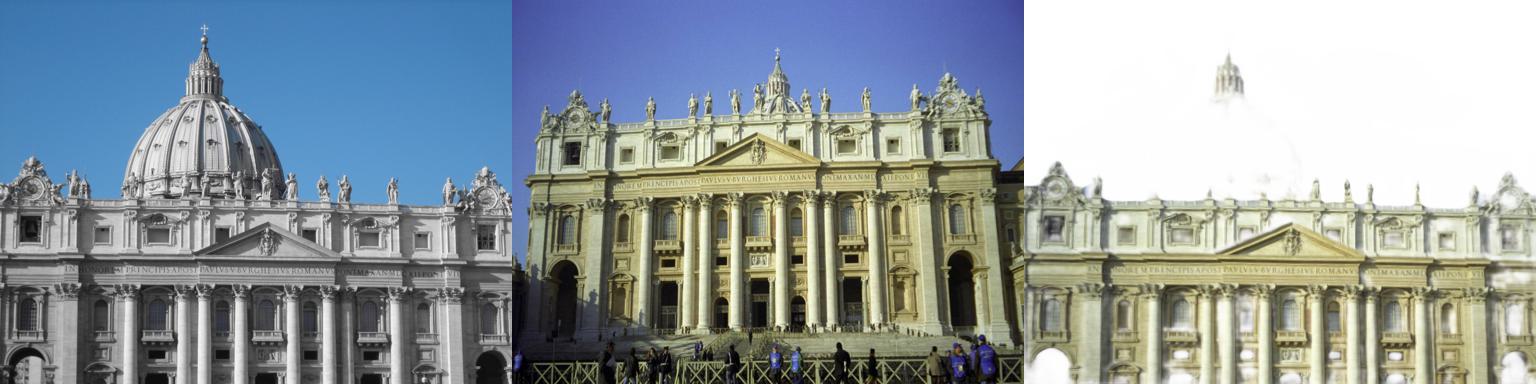}} \,
    \subfloat{\includegraphics[height=2.0cm]{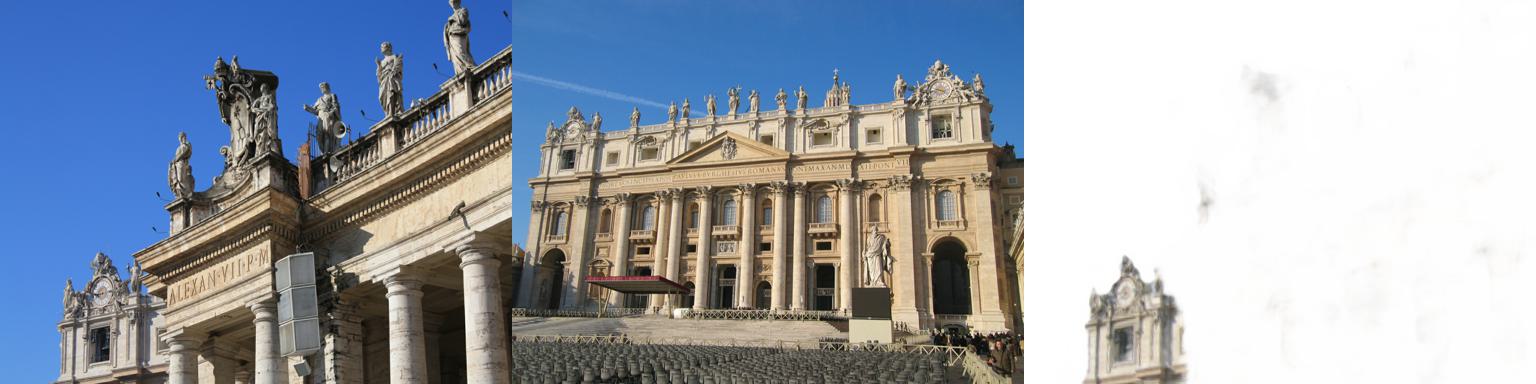}} \\
    \subfloat{\includegraphics[height=2.0cm]{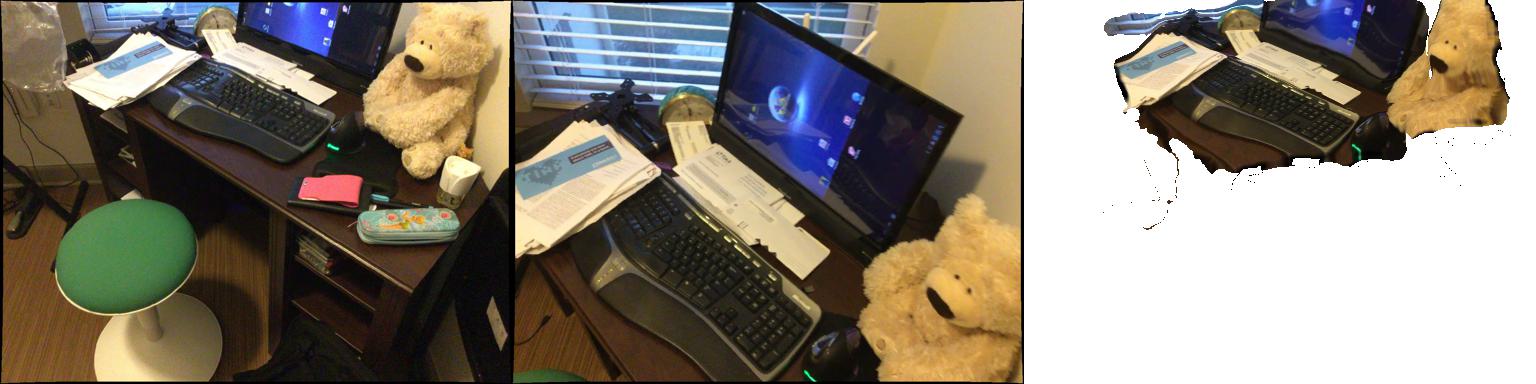}} \,
    \subfloat{\includegraphics[height=2.0cm]{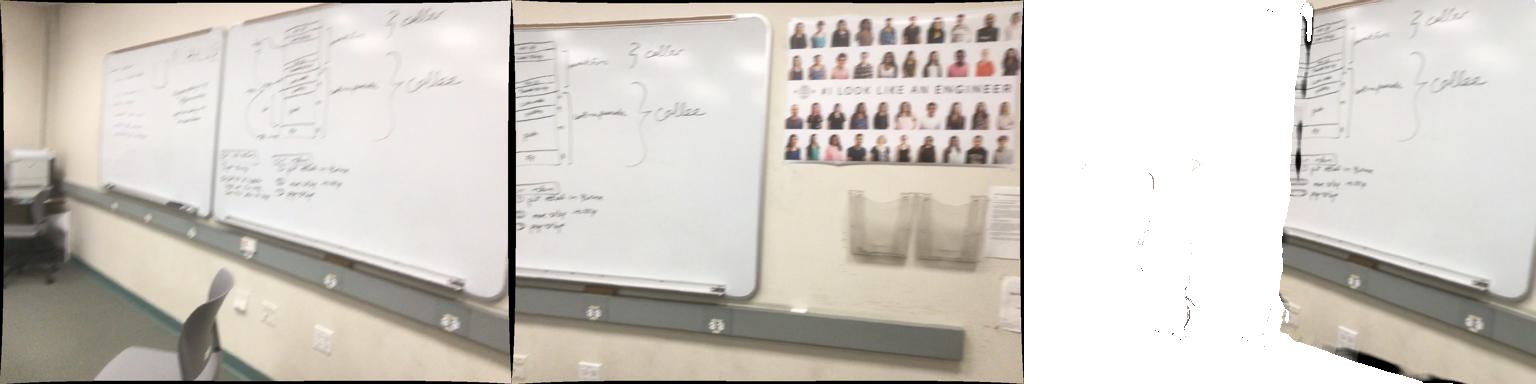}}
    \vspace{0mm}
    \caption{\small 
    \textbf{Visual Quality of the Reconstruction.} We visualize $4$ reconstructed images using estimated dense correspondences. 
    In each group, from left to right is the source image, support image, and the reconstructed image. 
    The areas of low confidence are filled with white color.
    In ScanNet where the confidence groundtruth is not available, we use forward-backward flow consistency mask as a replacement.
    }
    \vspace{-2mm}
    \label{fig:reconstruction}
\end{figure*}

\subsection{Dense Geometric Matching}
We follow the RANSAC-Flow~\cite{shen2020ransac} in training and testing split on the MegaDepth dataset.
The PCK scores in Tab.~\ref{tab:mega_dense} refer to the thresholded keypoints accuracy.
We divide the baseline methods into single and multiple run methods.
Note, the baseline methods PDC Net~\cite{truong2021learning} and PDC Net+~\cite{truong2021pdc} consume the additional synthetic data generated using COCO~\cite{lin2014microsoft} instance segmentation label.
For PCK @$1 \text{px}$, we outperform the SoTA single and multiple run methods by an absolute margin of $4.89\%$ and $6.99\%$ respectively.  
Meanwhile, we are about $\mathbf{8\times}$ {\bf faster} than SoTA baselines while suppassing SoTA performance.

\subsection{Two-View Camera Pose Estimation}

\Paragraph{Evaluation Protocol}
In the MegaDepth, ScanNet, and Hpatches datasets, we follow the evaluation protocol of~\cite{sarlin2020superglue,sun2021loftr, edstedt2023dkm} in reporting the pose accuracy AUC curve thresholded at $5$, $10$, and $20$ degrees.
In the YFCC100m dataset, we follow the  protocol of RANSAC-Flow~\cite{shen2020ransac}, additionally reporting the pose mAP value.
The pose estimation is considered an outlier if its maximum degree error of translation or rotation exceeds the threshold.
The two-view relative pose is estimated using the five-point algorithm~\cite{nister2004efficient} with RANSAC~\cite{derpanis2010overview} via the OpenCV implementation~\cite{bradski2000opencv}.

\Paragraph{Baseline Methods} We compare with three groups of the methods, \textit{i.e.},  sparse methods with detector~\cite{sarlin2020superglue, li2020sgm},  sparse methods without detector~\cite{liu2022drc, sun2021loftr, tang2022quadtree, wang2022matchformer, chen2022aspanformer} and  dense methods~\cite{truong2021pdc, edstedt2023dkm, shen2020ransac, truong2021learning, dai2021learning, wiles2021co}.
For sparse detector based methods, we use SuperPoint~\cite{detone2018superpoint} as the keypoint detector.
For dense methods, we further categorize them into single-run  and multiple-run methods. 
For multiple-run methods, \textit{e.g.},  RANSAC-Flow~\cite{shen2020ransac}, it repeats the prediction while reducing the visual difference with an estimated homography transformation.
Among baselines, AspanFormer~\cite{chen2022aspanformer} is a recent publicly available sparse detector-free method, improving LofTR with a sophisticated attention mechanism.

\begin{figure*}[t!]
    \captionsetup[subfigure]{labelformat=empty}
    \centering
    \vspace{-2mm}
    \subfloat{\includegraphics[height=2.4cm]{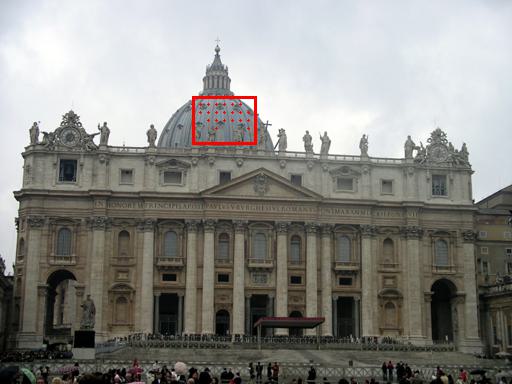}} \,
    \subfloat{\includegraphics[height=2.4cm]{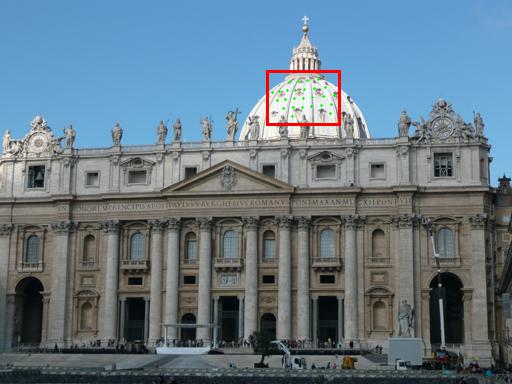}}\,
    \subfloat{\includegraphics[height=2.4cm]{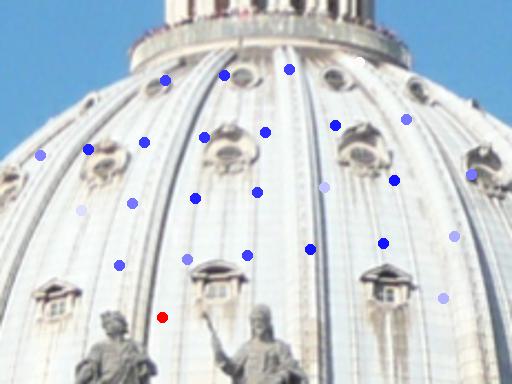}}\,
    \subfloat{\includegraphics[height=2.4cm]{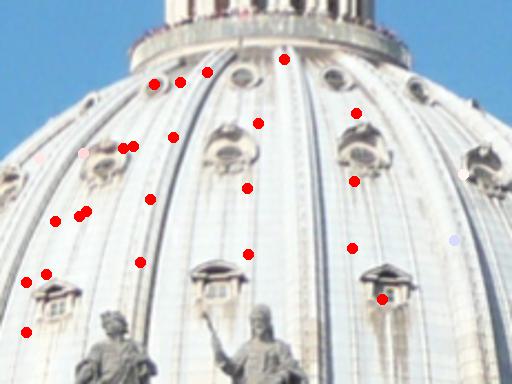}}\,
    \subfloat{\includegraphics[height=2.4cm]{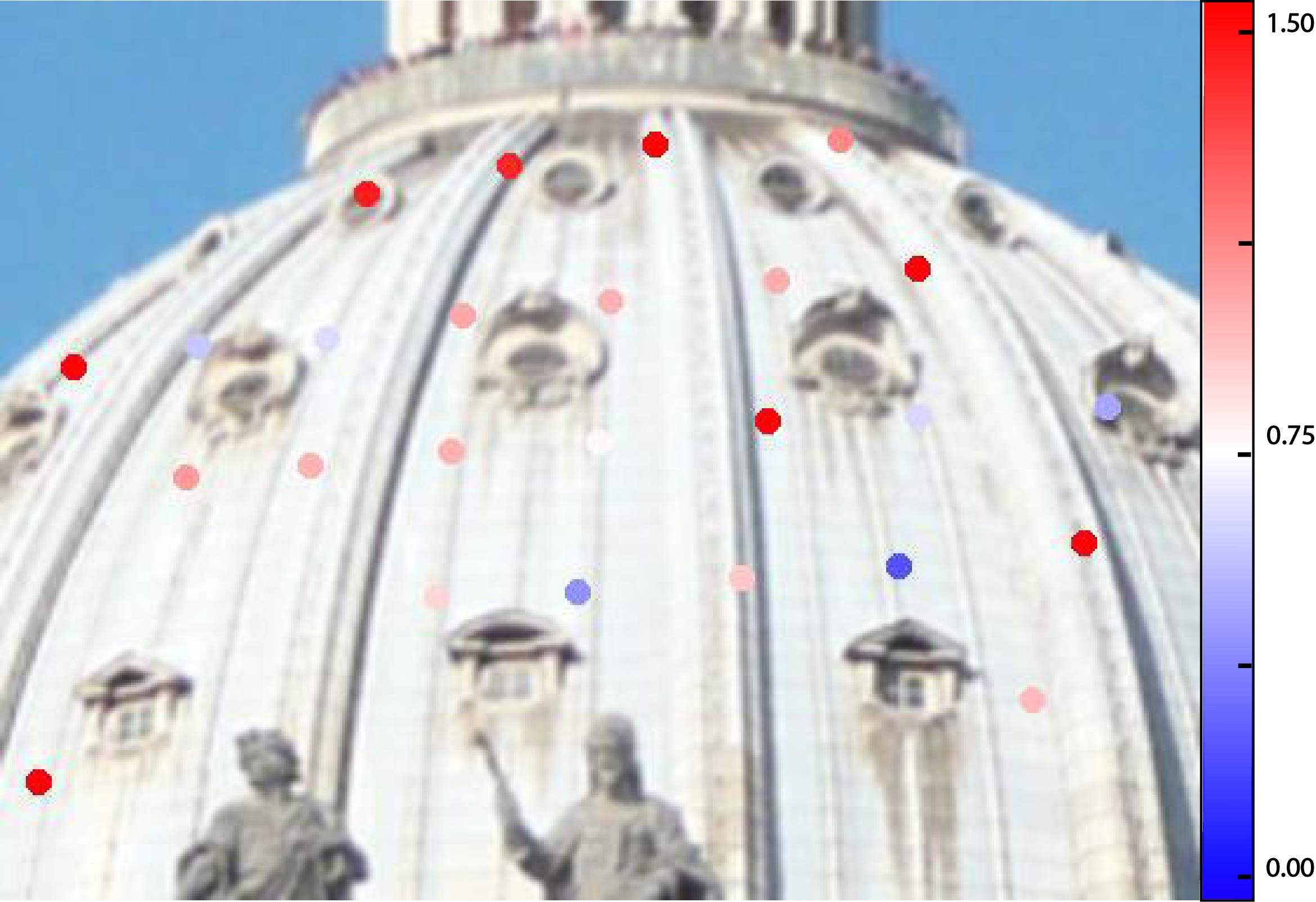}} \\
    \subfloat[(a) Source Frame $\mathbf{I}_1$]{\includegraphics[height=2.4cm]{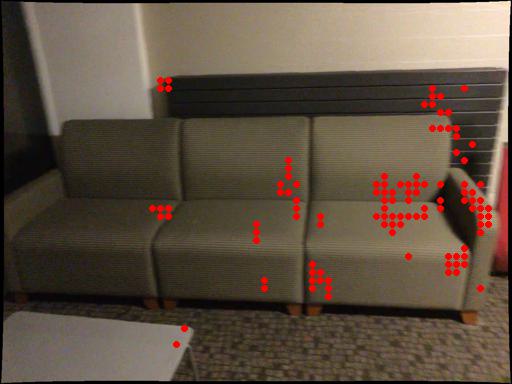}} \,
    \subfloat[(b) Support Frame $\mathbf{I}_2$]{\includegraphics[height=2.4cm]{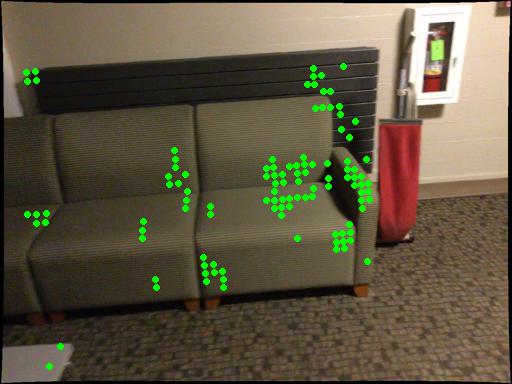}}\,
    \subfloat[(c) PMatch (Ours)]{\includegraphics[height=2.4cm]{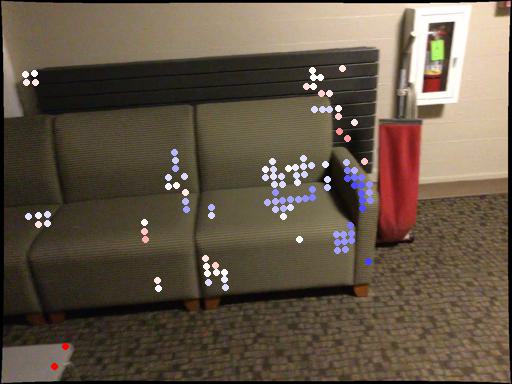}}\,
    \subfloat[(d) DKM~\cite{edstedt2023dkm}]{\includegraphics[height=2.4cm]{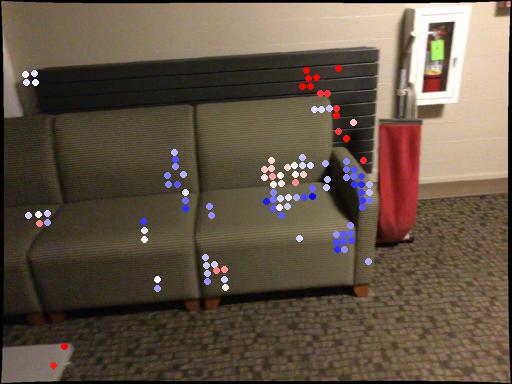}}\,
    \subfloat[(e) LoFTR~\cite{sun2021loftr}]{\includegraphics[height=2.4cm]{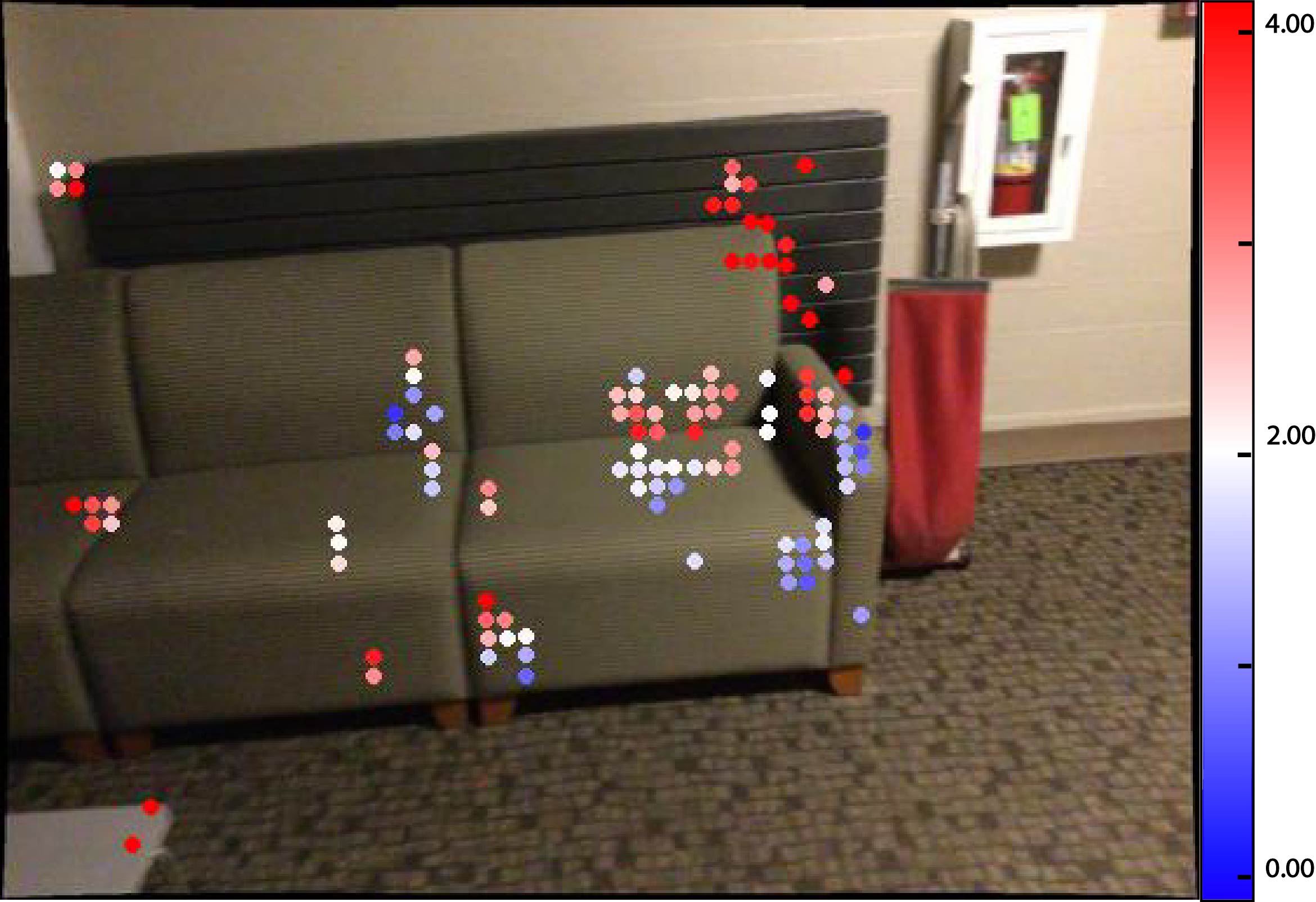}}
    \vspace{-1mm}
    \caption{\small 
    \textbf{Visual Comparisons.}
    We conduct the visual comparison against the SoTA dense~\cite{edstedt2023dkm} and sparse~\cite{sun2021loftr} methods on the MegaDepth and the ScanNet datasets.
    The color from \textcolor{blue}{blue} to \textcolor{red}{red} indicates an increment in the end-point-error (L2 error).
    }
    \vspace{-2mm}
    \label{fig:visual_cp}
\end{figure*}


\begin{table}[t!]
\centering
\captionsetup{font=small}
\resizebox{\linewidth}{!}{
\begin{tabular}[width=\linewidth]{cccccc}
\hline
\multicolumn{1}{c}{Category} & \multicolumn{1}{c}{Methods} & \multicolumn{1}{c}{Venue} & \multicolumn{3}{l}{Pose Estimation AUC $\uparrow$} \\
& &   & @$5^\circ$           & @$10^\circ$          & @$20^\circ$ \\
\hline
Sparse & SuperGlue~\cite{sarlin2020superglue} & CVPR'19 & $16.2$ & $33.8$ & $51.8$ \\
W/ Detector & SGMNet~\cite{li2020sgm} & PR'20 & $15.4$ & $32.1$ & $48.3$ \\
\hline
 & DRC-Net~\cite{liu2022drc} & ICASSP'22 & $7.7$ & $17.9$ & $30.5$ \\
  & LoFTR~\cite{sun2021loftr} & CVPR'21 & $22.0$ & $40.8$ & $57.6$ \\
Sparse & QuadTree~\cite{tang2022quadtree} & ICLR'22 & $24.9$ & $44.7$ & $61.8$ \\
Wo/ Detector & MatchFormer~\cite{wang2022matchformer} & ACCV'22 & $24.3$ & $43.9$ & $61.4$ \\
& ASpanFormer~\cite{chen2022aspanformer} & ECCV'22 & $\secondkey{25.6}$ & ${46.0}$ & ${63.3}$ \\
\hline
\multirow{3}{*}{Dense} & PDC-Net+~\cite{truong2021pdc} & Arxiv'19 & $20.2$ & $39.4$ & $57.1$ \\
& DKM~\cite{edstedt2023dkm} & CVPR'23 & $\firstkey{29.4}$ & $\firstkey{50.7}$ & $\firstkey{68.3}$ \\
& PMatch (Ours) & CVPR'23 & $\firstkey{29.4}$  & $\secondkey{50.1}$  &  $\secondkey{67.4}$      \\      
\hline
\end{tabular}
}
\vspace{0mm}
\caption{\small \textbf{ScanNet Two-View Camera Pose Estimation.} We follow SuperGlue~\cite{sarlin2020superglue} in the testing protocol. 
The pose AUC error is reported.
Our method achieves clear improvement over other baselines.
[Key: \firstkey{Best}, \secondkey{Second Best}]
}
\label{tab:scannet}
\end{table}

\begin{table}[t!]
\centering
\captionsetup{font=small}
\resizebox{ \linewidth}{!}{
\begin{tabular}{cccccccc}
\hline
\multicolumn{1}{c}{Methods} & \multicolumn{1}{c}{Venue} & \multicolumn{3}{l}{Pose Estimation AUC $\uparrow$} & \multicolumn{3}{l}{Pose Estimation mAP $\uparrow$} \\
&   & @$5^\circ$           & @$10^\circ$          & @$20^\circ$ & @$5^\circ$           & @$10^\circ$          & @$20^\circ$ \\
\hline
RANSAC-Flow~\cite{shen2020ransac} & ECCV'20 & - & - & - & $64.9$ & $73.3$ & $81.6$ \\
PDC-Net~\cite{truong2021learning}   & CVPR'21 & $35.7$ & $55.8$ & $72.3$ & $63.9$ & $73.0$ & $81.2$\\
PDC-Net+~\cite{truong2021pdc}   & Arxiv'21 & $37.5$ & $58.1$ & $74.5$ & $\secondkey{67.4}$ & $\secondkey{76.6}$ & $\secondkey{84.6}$\\
\hline
OANet~\cite{dai2021learning} & ICCV'19 & - & - & - & 52.2 & - & - \\
CoAM~\cite{wiles2021co} & CVPR'21 & - & - & - & $55.6$ & $66.8$ & - \\
PDC-Net~\cite{truong2021learning}   & CVPR'21 & $32.2$ & $52.6$ & $70.1$  & $60.5$ & $70.9$ & $80.3$ \\
PDC-Net+~\cite{truong2021pdc}   & Arxiv'21 & $34.8$ & $55.4$ & $72.6$ &  $63.9$ & $73.8$ & $82.7$\\
ASpanFormer~\cite{chen2022aspanformer} & ECCV'22 & $\secondkey{44.5}$ & $\secondkey{63.8}$ & $\secondkey{78.4}$ &- & -& -\\
PMatch (Ours)  & CVPR'23  & $\firstkey{45.7}$ & $\firstkey{65.2}$ & $\firstkey{79.8}$ & $\firstkey{75.9}$ & $\firstkey{83.1}$ & $\firstkey{89.3}$\\      
\hline
\end{tabular}
}
\vspace{0mm}
\caption{\small \textbf{YFCC100m Two-View Camera Pose Estimation.} 
The upper group runs multiple times, while the lower group runs a single time.
We follow \cite{zhang2019learning} in the evaluation and preprocessing, reporting both pose AUC and mAP errors.
[Key: \firstkey{Best}, \secondkey{Second Best}]
\vspace{-7mm}}
\label{tab:yfcc100m}
\end{table}


\Paragraph{Outdoor Dataset}
We test our method on the outdoor dataset MegaDepth. 
We follow the training and validation split of \cite{sarlin2020superglue, sun2021loftr, edstedt2023dkm}. 
The evaluation split contains $1,500$ paired images randomly selected from the scene $0015$ and $0022$. 
As shown in Tab.~\ref{tab:mega}, we achieve an absolute improvement of $0.9\%$ over the recent SoTA dense method DKM~\cite{edstedt2023dkm}.
Compared to the SoTA sparse method ASpanFormer~\cite{chen2022aspanformer}, we maintain an improvement of $6.1\%$.

\Paragraph{Indoor Dataset}
We test our method on the indoor dataset ScanNet.
We follow \cite{edstedt2023dkm} in training and testing protocol, resizing images to $480 \times 640$. 
The validation split of ScanNet consists of $1,500$ image pairs~\cite{sarlin2020superglue}.
In Tab.~\ref{tab:scannet}, we maintain competitive performance with the SoTA dense method DKM~\cite{edstedt2023dkm} and outperform SoTA sparse method by $1.4\%$.

\Paragraph{Generalization to YFCC100m}
We use the MegaDepth trained model to test on YFCC100m~\cite{thomee2016yfcc100m} dataset.
We follow the preprocessing steps of \cite{zhang2019learning}, evaluated on $4$ scenes with a total of $1,000$ images.
During the evaluation, we resample the input images of the shorter side to $480$.
Tab.~\ref{tab:yfcc100m} shows that our method can achieve a superior generalization ability, maintaining an improvement of $1.2\%$ over SoTA sparse methods~\cite{chen2022aspanformer}.


\Paragraph{Generalization to HPatches}
Following LoFTR~\cite{sun2021loftr}, we test the MegaDepth dataset trained model on HPatches.
In evaluation, the homography matrix is estimated using OpenCV's implementation.
We compare correspondences accuracy computed using the groundtruth and estimated homography.
The image pairs in HPatches have lighting differences or view differences.
The pattern is different from the training dataset MegaDepth.
Under the unseen testing scenario, our model generalizes best among baselines.

\begin{table}[t!]
\centering
\captionsetup{font=small}
\resizebox{\linewidth}{!}{
\begin{tabular}[width=\linewidth]{cccccc}
\hline
\multicolumn{1}{c}{Category} & \multicolumn{1}{c}{Methods} & \multicolumn{1}{c}{Venue} & \multicolumn{3}{l}{Pose Estimation AUC $\uparrow$} \\
&   & & @$3 \text{px}$           & @$5 \text{px}$          & @$10 \text{px}$ \\
\hline
 & D2Net~\cite{dusmanu2019d2} & CVPR'19 & $23.2$ & $35.9$ & $53.6$ \\
Sparse & R2D2~\cite{revaud2019r2d2} & NeurIPS'19 & $50.6$ & $63.9$ & $76.8$ \\
W/ Detector & DISK~\cite{tyszkiewicz2020disk}  & NeurIPS'20 & $52.3$ & $64.9$ & $78.9$ \\
& SuperGlue & CVPR'19 & $53.9$ & $68.3$ & $81.7$ \\
\hline
 & NCNet~\cite{rocco2020efficient} & ECCV'20 & $48.9$ & $54.2$ & $67.1$ \\
Sparse & DRC-Net~\cite{liu2022drc} & ICASSP'22 & $50.6$ & $56.2$ & $68.3$ \\
Wo/ Detector & LoFTR~\cite{sun2021loftr} & CVPR'21 & $65.9$ & $75.6$ & $\secondkey{84.6}$ \\
\hline
\multirow{2}{*}{Dense} & DKM~\cite{edstedt2023dkm} & CVPR'23 & $\secondkey{71.3}$ & $\secondkey{80.6}$ & $\firstkey{88.5}$ \\
& PMatch (Ours) & CVPR'23 & $\firstkey{71.9}$  & $\firstkey{80.7}$  & $\firstkey{88.5}$     \\      
\hline
\end{tabular}
}
\vspace{0mm}
\caption{\small \textbf{Hpatches Homography Estimation.} We follow \cite{sun2021loftr} in evaluation protocol. 
We report the corner point AUC error under the estimated homography matrix.
[Key: \firstkey{Best}, \secondkey{Second Best}]
\vspace{-2mm}}
\label{tab:hpatch}
\end{table}

\section{Ablation Study}
\Paragraph{Qualitative Comparison} 
The visual quality of reconstructed images using the predicted correspondences is visualized in Fig.~\ref{fig:reconstruction}.
We conduct a visual comparison with other SoTA dense and sparse methods in Fig.~\ref{fig:visual_cp}.
In Row $1$, (c), and (d), compared to DKM~\cite{edstedt2023dkm}, the proposed CFGM module achieves correct initial correspondences.
In Row $1$, (c), and (e), compared to LoFTR~\cite{sun2021loftr}, multi-scale dense refinement improves fine-scale correspondence accuracy.
In Row $2$, (c), (d), and (e), our CFGM  and homography loss achieve accurate correspondence estimation on textureless planar surface, \textit{e.g.}, the black wall behind the sofa.

\Paragraph{Running Time}
Evaluated on an RTX 2080 Ti GPU, we run $160$ ms for an image of $480 \times 640$ while LoFTR~\cite{sun2021loftr} runs $116$ ms and DKM~\cite{edstedt2023dkm} runs $148$ ms.
Our model runs similarly compared to the baselines.
The running time comparison to other dense methods is in Tab.~\ref{tab:mega_dense}.

\Paragraph{Benefit of the paired MIM pretraining} Shown in Tab.~\ref{tab:ablation}, with the paired MIM pretext task, the pose accuracy thresholded at $5^\circ$ improves by $3.5\% = 61.4\% - 57.9\%$.
A visual result of the paired MIM task is shown in Fig.~\ref{fig:pretext_vls}.

\Paragraph{CFGM and Homography Loss} The benefit of the proposed CFGM module and homography loss $L_h$ is included in Tab.~\ref{tab:ablation}. 
They help the network predict more accurate results in textureless planar surfaces.


\begin{table}[t!]
\centering
\captionsetup{font=small}
\resizebox{ \linewidth}{!}{
\begin{tabular}[width=\linewidth]{ccccccccc}
\hline
Baseline & CFGM & $L_H$ & pMIM Encoder & pMIM Decoder & \multicolumn{3}{l}{Pose Estimation AUC $\uparrow$} \\
 & & & ($E_\theta, T_\theta, L_\theta$) & ($R_\theta$)& @$5^\circ$           & @$10^\circ$          & @$20^\circ$ \\
\hline
$\checkmark$ & & & & &$56.1$ & $71.5$ & $83.0$ \\
$\checkmark$ & $\checkmark$ & & & & $57.5$ & $72.6$ & $83.9$ \\
$\checkmark$ & $\checkmark$ & $\checkmark$ & & & $57.9$ & $72.9$ & $84.1$  \\
$\checkmark$ & $\checkmark$ & $\checkmark$ & $\checkmark$ & & $60.6$ & $75.0$  & $85.3$  \\
$\checkmark$ & $\checkmark$ & $\checkmark$ & $\checkmark$ &$\checkmark$ & $\mathbf{61.4}$ & $\mathbf{75.7}$ & $\mathbf{85.7}$  \\
\hline
\end{tabular}
}
\vspace{-3mm}
\caption{\small \textbf{Ablation Studies  on MegaDepth.} 
The baseline method is the network in Fig.~\ref{fig:framework} with only a LoFTR module, \textit{i.e.}, without the other components of CFGM.
The ablation is conducted under the same training and testing resolution as Tab.~\ref{tab:mega}. Bold marks best.
\label{tab:ablation}
}
\label{tab:ablation}
\end{table}

\section{Conclusion}
This work investigates the benefit of pretraining the encoder and decoder of a dense geometric matching network under the paired MIM task.
We solve the discrepancy between the pretraining and finetuning tasks.
Also, we contribute an improved geometric matching network by reducing the ambiguity of textureless patches and augmenting the learning of local planar surfaces.

\Paragraph{Limitation}
Our method does not produce robust local descriptors.
When registering a keypoint, our method needs to run dense matching over all past frames, imposing latency for time-sensitive applications,  \textit{e.g.}, odometry estimation.

{\small
\bibliographystyle{ieee_fullname}
\bibliography{egbib}

\begin{thebibliography}{10}\itemsep=-1pt

\bibitem{balntas2017hpatches}
Vassileios Balntas, Karel Lenc, Andrea Vedaldi, and Krystian Mikolajczyk.
\newblock Hpatches: A benchmark and evaluation of handcrafted and learned local
  descriptors.
\newblock In {\em CVPR}, 2017.

\bibitem{hpatches_2017_cvpr}
Vassileios Balntas, Karel Lenc, Andrea Vedaldi, and Krystian Mikolajczyk.
\newblock Hpatches: A benchmark and evaluation of handcrafted and learned local
  descriptors.
\newblock In {\em CVPR}, 2017.

\bibitem{bao2021beit}
Hangbo Bao, Li Dong, and Furu Wei.
\newblock Beit: Bert pre-training of image transformers.
\newblock In {\em ICLR}, 2022.

\bibitem{baruch2021arkitscenes}
Gilad Baruch, Zhuoyuan Chen, Afshin Dehghan, Tal Dimry, Yuri Feigin, Peter Fu,
  Thomas Gebauer, Brandon Joffe, Daniel Kurz, Arik Schwartz, et~al.
\newblock Arkitscenes--a diverse real-world dataset for 3d indoor scene
  understanding using mobile rgb-d data.
\newblock {\em arXiv preprint arXiv:2111.08897}, 2021.

\bibitem{bay2008speeded}
Herbert Bay, Andreas Ess, Tinne Tuytelaars, and Luc Van~Gool.
\newblock Speeded-up robust features (surf).
\newblock {\em CVIU}, 2008.

\bibitem{bradski2000opencv}
Gary Bradski and Adrian Kaehler.
\newblock Opencv.
\newblock {\em Dr. Dobb’s journal of software tools}, 2000.

\bibitem{brahmbhatt2018geometry}
Samarth Brahmbhatt, Jinwei Gu, Kihwan Kim, James Hays, and Jan Kautz.
\newblock Geometry-aware learning of maps for camera localization.
\newblock In {\em CVPR}, 2018.

\bibitem{caron2021emerging}
Mathilde Caron, Hugo Touvron, Ishan Misra, Herv{\'e} J{\'e}gou, Julien Mairal,
  Piotr Bojanowski, and Armand Joulin.
\newblock Emerging properties in self-supervised vision transformers.
\newblock In {\em ICCV}, pages 9650--9660, 2021.

\bibitem{chen2022aspanformer}
Hongkai Chen, Zixin Luo, Lei Zhou, Yurun Tian, Mingmin Zhen, Tian Fang, David
  Mckinnon, Yanghai Tsin, and Long Quan.
\newblock Aspanformer: Detector-free image matching with adaptive span
  transformer.
\newblock In {\em ECCV}, 2022.

\bibitem{chen2020generative}
Mark Chen, Alec Radford, Rewon Child, Jeffrey Wu, Heewoo Jun, David Luan, and
  Ilya Sutskever.
\newblock Generative pretraining from pixels.
\newblock In {\em ICML}, 2020.

\bibitem{chen2020simple}
Ting Chen, Simon Kornblith, Mohammad Norouzi, and Geoffrey Hinton.
\newblock A simple framework for contrastive learning of visual
  representations.
\newblock In {\em ICML}, 2020.

\bibitem{dai2017scannet}
Angela Dai, Angel~X Chang, Manolis Savva, Maciej Halber, Thomas Funkhouser, and
  Matthias Nie{\ss}ner.
\newblock Scannet: Richly-annotated 3d reconstructions of indoor scenes.
\newblock In {\em CVPR}, 2017.

\bibitem{dai2021learning}
Luanyuan Dai, Xin Liu, Jingtao Wang, Changcai Yang, and Riqing Chen.
\newblock Learning two-view correspondences and geometry via local neighborhood
  correlation.
\newblock {\em Entropy}, 2021.

\bibitem{imagenet_cvpr09}
J. Deng, W. Dong, R. Socher, L.-J. Li, K. Li, and L. Fei-Fei.
\newblock {ImageNet: A Large-Scale Hierarchical Image Database}.
\newblock In {\em CVPR09}, 2009.

\bibitem{derpanis2010overview}
Konstantinos~G Derpanis.
\newblock Overview of the ransac algorithm.
\newblock {\em Image Rochester NY}, 2010.

\bibitem{detone2018superpoint}
Daniel DeTone, Tomasz Malisiewicz, and Andrew Rabinovich.
\newblock Superpoint: Self-supervised interest point detection and description.
\newblock In {\em CVPRW}, 2018.

\bibitem{dosovitskiy2020image}
Alexey Dosovitskiy, Lucas Beyer, Alexander Kolesnikov, Dirk Weissenborn,
  Xiaohua Zhai, Thomas Unterthiner, Mostafa Dehghani, Matthias Minderer, Georg
  Heigold, Sylvain Gelly, et~al.
\newblock An image is worth 16x16 words: Transformers for image recognition at
  scale.
\newblock In {\em ICLR}, 2021.

\bibitem{dubrofsky2009homography}
Elan Dubrofsky.
\newblock Homography estimation.
\newblock {\em Diplomov{\'a} pr{\'a}ce. Vancouver: Univerzita Britsk{\'e}
  Kolumbie}, 2009.

\bibitem{dusmanu2019d2}
Mihai Dusmanu, Ignacio Rocco, Tomas Pajdla, Marc Pollefeys, Josef Sivic,
  Akihiko Torii, and Torsten Sattler.
\newblock D2-net: A trainable cnn for joint detection and description of local
  features.
\newblock In {\em CVPR}, 2019.

\bibitem{edstedt2023dkm}
Johan Edstedt, Ioannis Athanasiadis, Mårten Wadenbäck, and Michael Felsberg.
\newblock {DKM}: Dense kernelized feature matching for geometry estimation.
\newblock In {\em CVPR}, 2023.

\bibitem{engel2017direct}
Jakob Engel, Vladlen Koltun, and Daniel Cremers.
\newblock Direct sparse odometry.
\newblock {\em PAMI}, 2017.

\bibitem{grill2020bootstrap}
Jean-Bastien Grill, Florian Strub, Florent Altch{\'e}, Corentin Tallec, Pierre
  Richemond, Elena Buchatskaya, Carl Doersch, Bernardo Avila~Pires, Zhaohan
  Guo, Mohammad Gheshlaghi~Azar, et~al.
\newblock Bootstrap your own latent-a new approach to self-supervised learning.
\newblock In {\em NeuriPS}, 2020.

\bibitem{he2022masked}
Kaiming He, Xinlei Chen, Saining Xie, Yanghao Li, Piotr Doll{\'a}r, and Ross
  Girshick.
\newblock Masked autoencoders are scalable vision learners.
\newblock In {\em CVPR}, 2022.

\bibitem{he2016deep}
Kaiming He, Xiangyu Zhang, Shaoqing Ren, and Jian Sun.
\newblock Deep residual learning for image recognition.
\newblock In {\em CVPR}, 2016.

\bibitem{huang2017densely}
Gao Huang, Zhuang Liu, Laurens Van Der~Maaten, and Kilian~Q Weinberger.
\newblock Densely connected convolutional networks.
\newblock In {\em CVPR}, 2017.

\bibitem{huang2021life}
Zhaoyang Huang, Xiaokun Pan, Runsen Xu, Yan Xu, Guofeng Zhang, Hongsheng Li,
  et~al.
\newblock Life: Lighting invariant flow estimation.
\newblock {\em arXiv preprint arXiv:2104.03097}, 2021.

\bibitem{katharopoulos2020transformers}
Angelos Katharopoulos, Apoorv Vyas, Nikolaos Pappas, and Fran{\c{c}}ois
  Fleuret.
\newblock Transformers are rnns: Fast autoregressive transformers with linear
  attention.
\newblock In {\em ICML}, 2020.

\bibitem{kingma2014adam}
Diederik~P Kingma and Jimmy Ba.
\newblock Adam: A method for stochastic optimization.
\newblock In {\em ICLR}, 2015.

\bibitem{li2020sgm}
Jianan Li, Xuemei Xie, Qingzhe Pan, Yuhan Cao, Zhifu Zhao, and Guangming Shi.
\newblock Sgm-net: Skeleton-guided multimodal network for action recognition.
\newblock {\em PR}, 2020.

\bibitem{li2018megadepth}
Zhengqi Li and Noah Snavely.
\newblock Megadepth: Learning single-view depth prediction from internet
  photos.
\newblock In {\em CVPR}, 2018.

\bibitem{lin2014microsoft}
Tsung-Yi Lin, Michael Maire, Serge Belongie, James Hays, Pietro Perona, Deva
  Ramanan, Piotr Doll{\'a}r, and C~Lawrence Zitnick.
\newblock Microsoft coco: Common objects in context.
\newblock In {\em ECCV}, 2014.

\bibitem{liu2022drc}
Jinjiang Liu and Xueliang Zhang.
\newblock Drc-net: Densely connected recurrent convolutional neural network for
  speech dereverberation.
\newblock In {\em ICASSP}, 2022.

\bibitem{lowe2004distinctive}
David~G Lowe.
\newblock Distinctive image features from scale-invariant keypoints.
\newblock {\em IJCV}, 2004.

\bibitem{melekhov2019dgc}
Iaroslav Melekhov, Aleksei Tiulpin, Torsten Sattler, Marc Pollefeys, Esa Rahtu,
  and Juho Kannala.
\newblock Dgc-net: Dense geometric correspondence network.
\newblock In {\em WACV}, 2019.

\bibitem{misra2020self}
Ishan Misra and Laurens van~der Maaten.
\newblock Self-supervised learning of pretext-invariant representations.
\newblock In {\em CVPR}, 2020.

\bibitem{nakagawa2015estimating}
Yosuke Nakagawa, Hideaki Uchiyama, Hajime Nagahara, and Rin-Ichiro Taniguchi.
\newblock Estimating surface normals with depth image gradients for fast and
  accurate registration.
\newblock In {\em 3DV}, 2015.

\bibitem{nister2004efficient}
David Nist{\'e}r.
\newblock An efficient solution to the five-point relative pose problem.
\newblock {\em PAMI}, 2004.

\bibitem{ono2018lf}
Yuki Ono, Eduard Trulls, Pascal Fua, and Kwang~Moo Yi.
\newblock Lf-net: Learning local features from images.
\newblock In {\em NeurIPS}, 2018.

\bibitem{oord2018representation}
Aaron van~den Oord, Yazhe Li, and Oriol Vinyals.
\newblock Representation learning with contrastive predictive coding.
\newblock {\em CoRR}, 2018.

\bibitem{revaud2019r2d2}
Jerome Revaud, Cesar De~Souza, Martin Humenberger, and Philippe Weinzaepfel.
\newblock R2d2: Reliable and repeatable detector and descriptor.
\newblock In {\em NeuriPS}, 2019.

\bibitem{roberts2021hypersim}
Mike Roberts, Jason Ramapuram, Anurag Ranjan, Atulit Kumar, Miguel~Angel
  Bautista, Nathan Paczan, Russ Webb, and Joshua~M Susskind.
\newblock Hypersim: A photorealistic synthetic dataset for holistic indoor
  scene understanding.
\newblock In {\em ICCV}, 2021.

\bibitem{rocco2020efficient}
Ignacio Rocco, Relja Arandjelovi{\'c}, and Josef Sivic.
\newblock Efficient neighbourhood consensus networks via submanifold sparse
  convolutions.
\newblock In {\em ECCV}, 2020.

\bibitem{rublee2011orb}
Ethan Rublee, Vincent Rabaud, Kurt Konolige, and Gary Bradski.
\newblock Orb: An efficient alternative to sift or surf.
\newblock In {\em ICCV}, 2011.

\bibitem{sarlin2020superglue}
Paul-Edouard Sarlin, Daniel DeTone, Tomasz Malisiewicz, and Andrew Rabinovich.
\newblock Superglue: Learning feature matching with graph neural networks.
\newblock In {\em CVPR}, 2020.

\bibitem{schonberger2016structure}
Johannes~L Schonberger and Jan-Michael Frahm.
\newblock Structure-from-motion revisited.
\newblock In {\em CVPR}, 2016.

\bibitem{shen2020ransac}
Xi Shen, Fran{\c{c}}ois Darmon, Alexei~A Efros, and Mathieu Aubry.
\newblock Ransac-flow: generic two-stage image alignment.
\newblock In {\em ECCV}, 2020.

\bibitem{simonyan2014very}
Karen Simonyan and Andrew Zisserman.
\newblock Very deep convolutional networks for large-scale image recognition.
\newblock In {\em ICLR}, 2015.

\bibitem{sun2021loftr}
Jiaming Sun, Zehong Shen, Yuang Wang, Hujun Bao, and Xiaowei Zhou.
\newblock Loftr: Detector-free local feature matching with transformers.
\newblock In {\em CVPR}, 2021.

\bibitem{tang2022quadtree}
Shitao Tang, Jiahui Zhang, Siyu Zhu, and Ping Tan.
\newblock Quadtree attention for vision transformers.
\newblock In {\em ICLR}, 2022.

\bibitem{teed2020raft}
Zachary Teed and Jia Deng.
\newblock Raft: Recurrent all-pairs field transforms for optical flow.
\newblock In {\em ECCV}, 2020.

\bibitem{thomee2016yfcc100m}
Bart Thomee, David~A Shamma, Gerald Friedland, Benjamin Elizalde, Karl Ni,
  Douglas Poland, Damian Borth, and Li-Jia Li.
\newblock Yfcc100m: The new data in multimedia research.
\newblock {\em Communications of the ACM}, 2016.

\bibitem{truong2020gocor}
Prune Truong, Martin Danelljan, Luc~V Gool, and Radu Timofte.
\newblock Gocor: Bringing globally optimized correspondence volumes into your
  neural network.
\newblock In {\em NeuriPs}, 2020.

\bibitem{truong2020glu}
Prune Truong, Martin Danelljan, and Radu Timofte.
\newblock Glu-net: Global-local universal network for dense flow and
  correspondences.
\newblock In {\em CVPR}, 2020.

\bibitem{truong2021pdc}
Prune Truong, Martin Danelljan, Radu Timofte, and Luc Van~Gool.
\newblock Pdc-net+: Enhanced probabilistic dense correspondence network.
\newblock {\em arXiv preprint arXiv:2109.13912}, 2021.

\bibitem{truong2021learning}
Prune Truong, Martin Danelljan, Luc Van~Gool, and Radu Timofte.
\newblock Learning accurate dense correspondences and when to trust them.
\newblock In {\em CVPR}, 2021.

\bibitem{tyszkiewicz2020disk}
Micha{\l} Tyszkiewicz, Pascal Fua, and Eduard Trulls.
\newblock Disk: Learning local features with policy gradient.
\newblock In {\em NeuriPS}, 2020.

\bibitem{ummenhofer2017demon}
Benjamin Ummenhofer, Huizhong Zhou, Jonas Uhrig, Nikolaus Mayer, Eddy Ilg,
  Alexey Dosovitskiy, and Thomas Brox.
\newblock Demon: Depth and motion network for learning monocular stereo.
\newblock In {\em CVPR}, 2017.

\bibitem{vincent2010stacked}
Pascal Vincent, Hugo Larochelle, Isabelle Lajoie, Yoshua Bengio, Pierre-Antoine
  Manzagol, and L{\'e}on Bottou.
\newblock Stacked denoising autoencoders: Learning useful representations in a
  deep network with a local denoising criterion.
\newblock {\em JMLR}, 2010.

\bibitem{wang2022matchformer}
Qing Wang, Jiaming Zhang, Kailun Yang, Kunyu Peng, and Rainer Stiefelhagen.
\newblock Matchformer: Interleaving attention in transformers for feature
  matching.
\newblock In {\em ACCV}, 2022.

\bibitem{wang2020tartanair}
Wenshan Wang, Delong Zhu, Xiangwei Wang, Yaoyu Hu, Yuheng Qiu, Chen Wang, Yafei
  Hu, Ashish Kapoor, and Sebastian Scherer.
\newblock Tartanair: A dataset to push the limits of visual slam.
\newblock In {\em IROS}, 2020.

\bibitem{wiles2021co}
Olivia Wiles, Sebastien Ehrhardt, and Andrew Zisserman.
\newblock Co-attention for conditioned image matching.
\newblock In {\em CVPR}, 2021.

\bibitem{xiao2021early}
Tete Xiao, Mannat Singh, Eric Mintun, Trevor Darrell, Piotr Doll{\'a}r, and
  Ross Girshick.
\newblock Early convolutions help transformers see better.
\newblock In {\em NeuriPs}, 2021.

\bibitem{xie2022simmim}
Zhenda Xie, Zheng Zhang, Yue Cao, Yutong Lin, Jianmin Bao, Zhuliang Yao, Qi
  Dai, and Han Hu.
\newblock Simmim: A simple framework for masked image modeling.
\newblock In {\em CVPR}, 2022.

\bibitem{yang2021instance}
Ceyuan Yang, Zhirong Wu, Bolei Zhou, and Stephen Lin.
\newblock Instance localization for self-supervised detection pretraining.
\newblock In {\em CVPR}, 2021.

\bibitem{yao2020blendedmvs}
Yao Yao, Zixin Luo, Shiwei Li, Jingyang Zhang, Yufan Ren, Lei Zhou, Tian Fang,
  and Long Quan.
\newblock Blendedmvs: A large-scale dataset for generalized multi-view stereo
  networks.
\newblock In {\em CVPR}, 2020.

\bibitem{yi2016lift}
Kwang~Moo Yi, Eduard Trulls, Vincent Lepetit, and Pascal Fua.
\newblock Lift: Learned invariant feature transform.
\newblock In {\em ECCV}, 2016.

\bibitem{zhang2019learning}
Jiahui Zhang, Dawei Sun, Zixin Luo, Anbang Yao, Lei Zhou, Tianwei Shen, Yurong
  Chen, Long Quan, and Hongen Liao.
\newblock Learning two-view correspondences and geometry using order-aware
  network.
\newblock In {\em ICCV}, 2019.

\bibitem{zhou2021ibot}
Jinghao Zhou, Chen Wei, Huiyu Wang, Wei Shen, Cihang Xie, Alan Yuille, and Tao
  Kong.
\newblock ibot: Image bert pre-training with online tokenizer.
\newblock In {\em ICLR}, 2022.

\end{thebibliography}


\begin{thebibliography}{1}\itemsep=-1pt

\bibitem{li2018megadepth}
Zhengqi Li and Noah Snavely.
\newblock Megadepth: Learning single-view depth prediction from internet
  photos.
\newblock In {\em CVPR}, 2018.

\end{thebibliography}
}

\end{document}


\title{
PMatch: Paired Masked Image Modeling for Dense Geometric Matching \\
==== Supplementary Material ====
}

\author{%
  Shengjie Zhu \\
  Michigan State University\\
  \texttt{zhusheng@msu.edu} \\
  \And
  Xiaoming Liu\\
  Michigan State University\\
  \texttt{liuxm@cse.msu.edu} \\
}
\maketitle

\begin{figure}[!t]
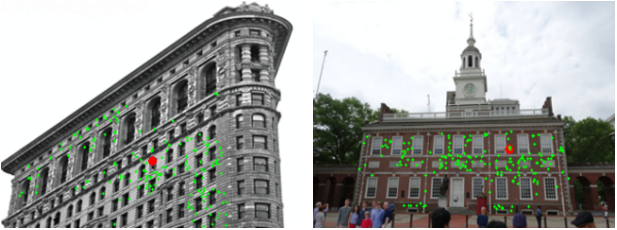

    \centering
    \vspace{-6mm}
    \subfloat{\includegraphics[height=2.8cm]{figure_supp/coplanar/[820].png}}\,
    \subfloat{\includegraphics[height=2.8cm]{figure_supp/coplanar/[5331].png}}
    \vspace{0mm}
    \caption{\small
    \textbf{Visualization of produced co-planar points in MegaDepth Dataset.}
    The \textcolor{red}{red} point is one anchor point $\mathbf{p}_m$, while the \textcolor{green}{green} dots are the co-planar pixels among $K$ sampled candidates $\{\mathbf{q}_{n}^m \mid 1\leq n \leq K \}$, computed following Eqn.~\ref{eqn:coplanar}.
    }
    \vspace{-4mm}
    \label{fig:coplanar}
\end{figure}

\section{Additional Implementation Details}
\Paragraph{Production of the indicator matrix $\mathcal{O}^+$.}
In the main paper Eqn.~$14$, we utilize an indicator matrix $\mathcal{O}^+$ to indicate the co-planar pairs between anchor and candidate points. 
Given the $K$ anchor points $\{\mathbf{p}_{m} \mid 1\leq m \leq K \}$ and $K \times K$ candidate points $\{\mathbf{q}_{n}^m \mid 1\leq n \leq K \}$, the indicator matrix $\mathcal{O}^+$ of size $K \times K$ is computed as:
\begin{equation}
    \mathcal{O}^+_{m, n} = 1 \quad \text{if}
    \begin{cases}
    1 - \arccos{ \left(\mathbf{n}_{\mathbf{p}_{m}}^\intercal \mathbf{n}_{\mathbf{q}_{n}^{m}} \right)} < k_1\\
    \text{dist}(\mathbf{n}_{\mathbf{p}_{m}}, \mathbf{p}_{m}, d_{\mathbf{p}_{m}}, \mathbf{q}_{n}^{m}, d_{\mathbf{q}_{n}^{m}}, \mathbf{K}_1)
    < k_2  \\
    \|
    \text{proj} (\mathbf{H}_{\mathbf{p}_{m}}^\intercal, \mathbf{q}_{n}^{m} )- \mathbf{q}_{n}^{m}\|_2 < k_3 .
\end{cases}
\label{eqn:coplanar}
\end{equation}
The function $\text{dist} (\cdot)$ computes the point to plane distance in 3D space.
The plane is spanned by the norm $\mathbf{n}_{\mathbf{p}_{m}}$ and the re-projected 3D point at $\mathbf{p}_{m}$.
The function $\text{proj}(\cdot)$ indicates the planar projection under pixel $\mathbf{p}_{m}$ homography matrix $\mathbf{\mathbf{H}_{\mathbf{p}_{m}}}$ (see main paper equation Eqn.~$13$).
The $k_1$, $k_2$ and $k_3$ are set to $0.002$, $0.02$, and $1$ respectively.
We visualize the produced groundtruth co-planar points in Fig.~\ref{fig:coplanar}.

\begin{table}[t!]
\centering
\captionsetup{font=small}
\vspace{0mm}
\resizebox{0.8\linewidth}{!}{
\begin{tabular}[width=\linewidth]{cccc}
\hline
\multicolumn{1}{c}{Methods} & \multicolumn{3}{l}{Pose Estimation AUC $\uparrow$} \\
  & @$3 \text{px}$           & @$5 \text{px}$          & @$10 \text{px}$ \\
\hline
Without Homography Loss & ${68.5}$ & ${78.3}$ & ${86.8}$ \\
With Homography Loss & $\mathbf{69.2}$  & $\mathbf{78.8}$  & $\mathbf{87.1}$     \\      
\hline
\end{tabular}
}
\vspace{0mm}
\caption{\small \textbf{Hpatches Homography Loss $L_H$ Ablation.} We follow \cite{sun2021loftr} in evaluation protocol.
We report the corner point AUC error under the estimated homography matrix using a model trained with and without the homography loss $L_h$.
The model is trained using the resolution of $240 \times 320$ on MegaDepth~\cite{li2018megadepth} Dataset.
\vspace{-2mm}}
\label{tab:hpatch_ablation}
\end{table}

\begin{table}[th!]
\centering
\captionsetup{font=small}
\resizebox{0.6 \linewidth}{!}{
\begin{tabular}[width=\linewidth]{lcc}
\hline
\multicolumn{1}{c}{Ablation} & \multicolumn{2}{l}{Image Reconstruction Error} \\
& L$1$ $\downarrow$          & $1$ - SSIM~\cite{dosselmann2011comprehensive}  $\downarrow$       \\
\hline
MIM & $0.696$ & ${0.005}$  \\
Paired MIM & $\mathbf{0.625}$ & $\mathbf{0.004}$ \\
\hline
\end{tabular}
}
\vspace{-2mm}
\caption{\small \textbf{Ablation on MIM Pretext Task.} 
We ablate the image reconstruction quality between the original MIM and the paired MIM.
The ablation is conducted on MegaDepth Dataset. The random sampling is kept identical across the comparison.
The image pixel ranges between $0$ and $255$ during the evaluation.
More details are specified in Tab.~\ref{tab:ablation}.
\label{tab:reconsttuction}
}
\vspace{-3mm}
\end{table}

\section{Additional Ablations}


\Paragraph{Homography Loss.}
Each pair of the images in Hpatches~\cite{hpatches_2017_cvpr} dataset follows the planar homography transformation with groundtruth matrix available.
We thus ablate homography loss $L_H$ on Hpatches dataset, as shown in Tab.~\ref{tab:hpatch_ablation}.
With $L_H$, we achieve an absolute improvement of $0.7\%$ on pose error thresholded at $3 \text{px}$.

\Paragraph{Benefit of the paired MIM.}
We compare between the original MIM and the proposed paired MIM.
For the original MIM, we follow SimMIM~\cite{xie2022simmim} in using the encoder and a lightweight decoder head.
The encoder includes the $E_\theta$ and $T_\theta$.
For the paired MIM, we include the $E_\theta$, $T_\theta$, $L_\theta$ and, $R_\theta'$ at scale $8$ and scale $4$ as shown in the main paper Fig.~$2$.
Other training specifications are kept identical to the main paper Tab.~$6$.
We ablate on two perspectives.
On pretext task image reconstruction, we measure the reconstruction quality between the two, as shown in Tab.~\ref{tab:reconsttuction}.
In Tab.~\ref{tab:reconsttuction}, paired MIM improves the reconstruction quality by reduce the metric $L_1$ from $0.696$ to $0.625$.
On dense matching, as shown in Tab.~\ref{tab:ablation},
we see performance improvement by including more network components.
On pose metric AUC thresholded at $5^\circ$, we achieve an absolute improvement of $0.7 \%$ over the MIM pretraining.

\begin{figure*}[htb!]
    \centering
    \subfloat{\includegraphics[height=4.1cm]{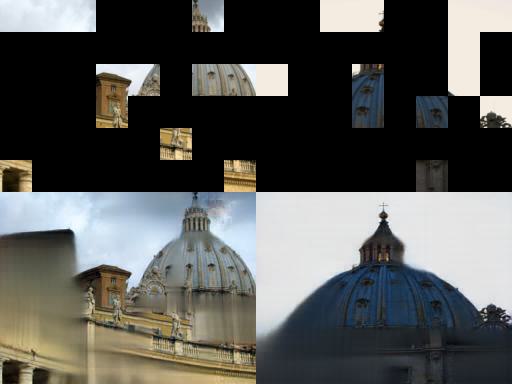}}\,
    \subfloat{\includegraphics[height=4.1cm]{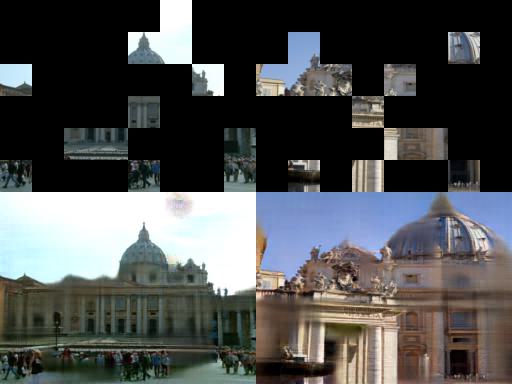}}\,
    \subfloat{\includegraphics[height=4.1cm]{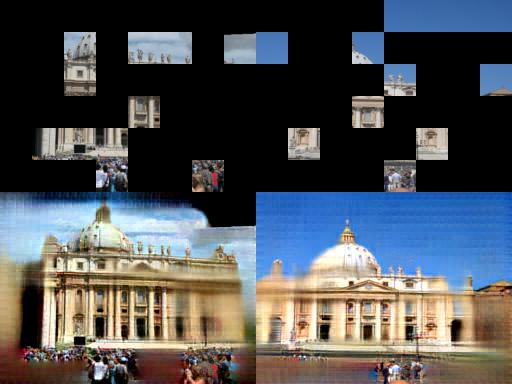}}\\
    \subfloat{\includegraphics[height=4.1cm]{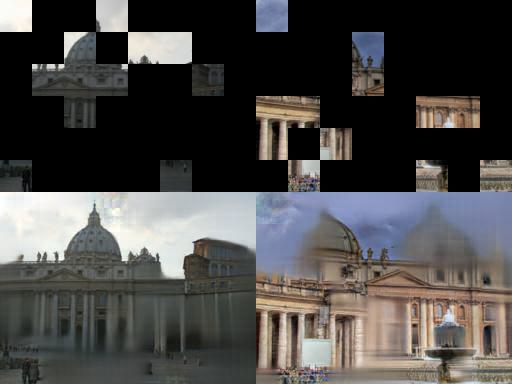}}\,
    \subfloat{\includegraphics[height=4.1cm]{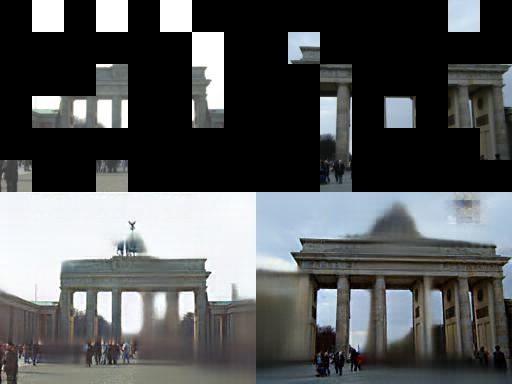}}\,
    \subfloat{\includegraphics[height=4.1cm]{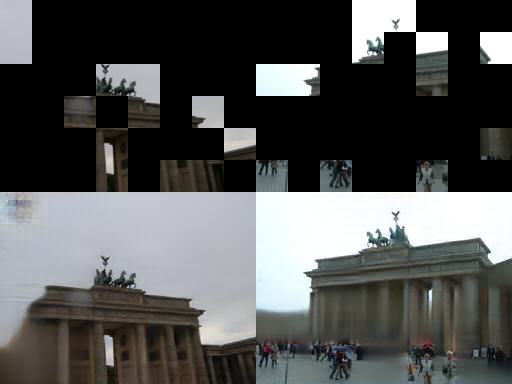}}\\
    \subfloat{\includegraphics[height=4.1cm]{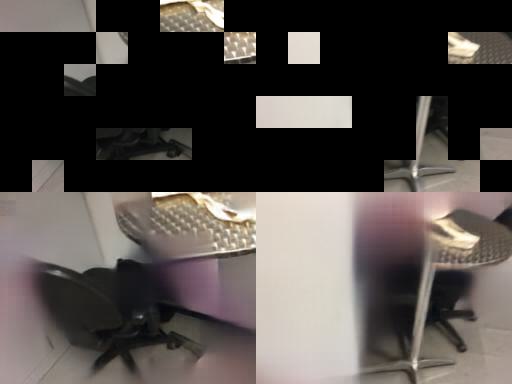}}\,
    \subfloat{\includegraphics[height=4.1cm]{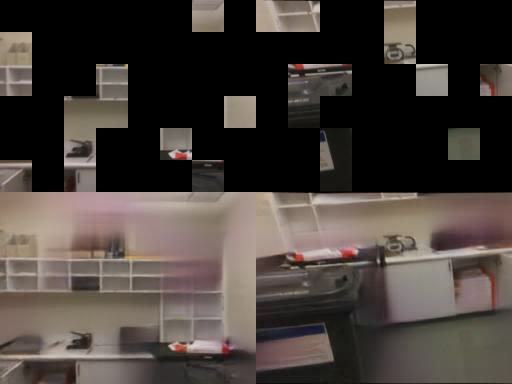}}\,
    \subfloat{\includegraphics[height=4.1cm]{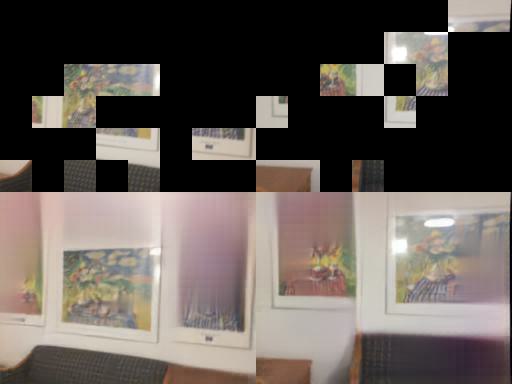}}\\
    \subfloat{\includegraphics[height=4.1cm]{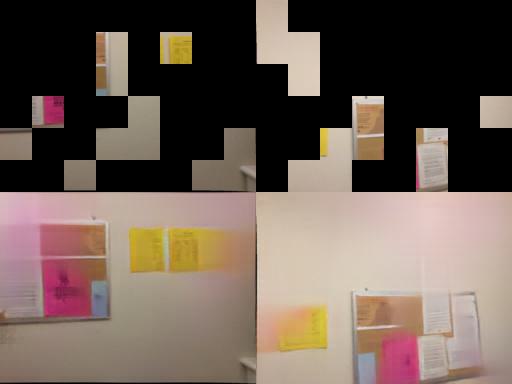}}\,
    \subfloat{\includegraphics[height=4.1cm]{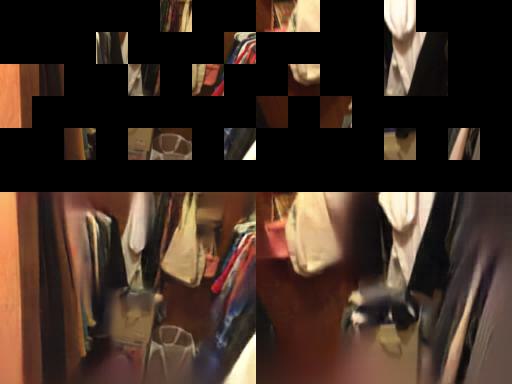}}\,
    \subfloat{\includegraphics[height=4.1cm]{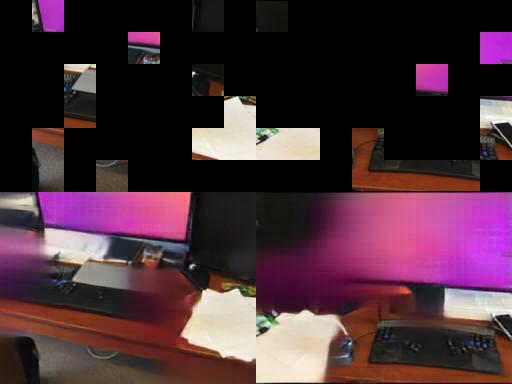}}\\
    \vspace{0mm}
    \caption{\small 
    \textbf{Visual Quality of the paired MIM pretext task.}
    Visualized cases are from the MegaDepth and the ScanNet dataset.}
    \vspace{-2mm}
    \label{fig:pretext_vls}
\end{figure*}

\begin{figure*}[t!]
    \centering
    \subfloat{\includegraphics[height=2.1cm]{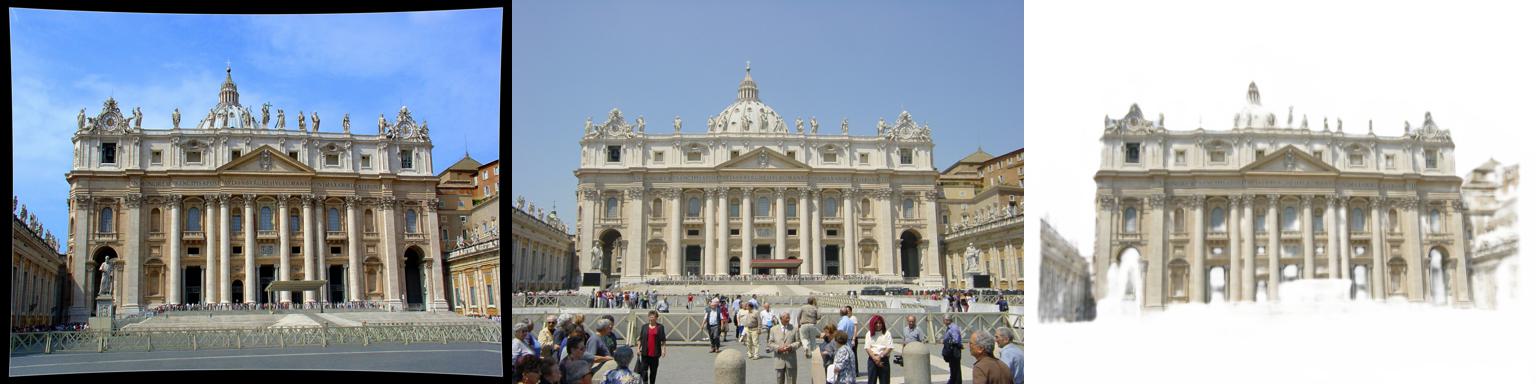}} \,
    \subfloat{\includegraphics[height=2.1cm]{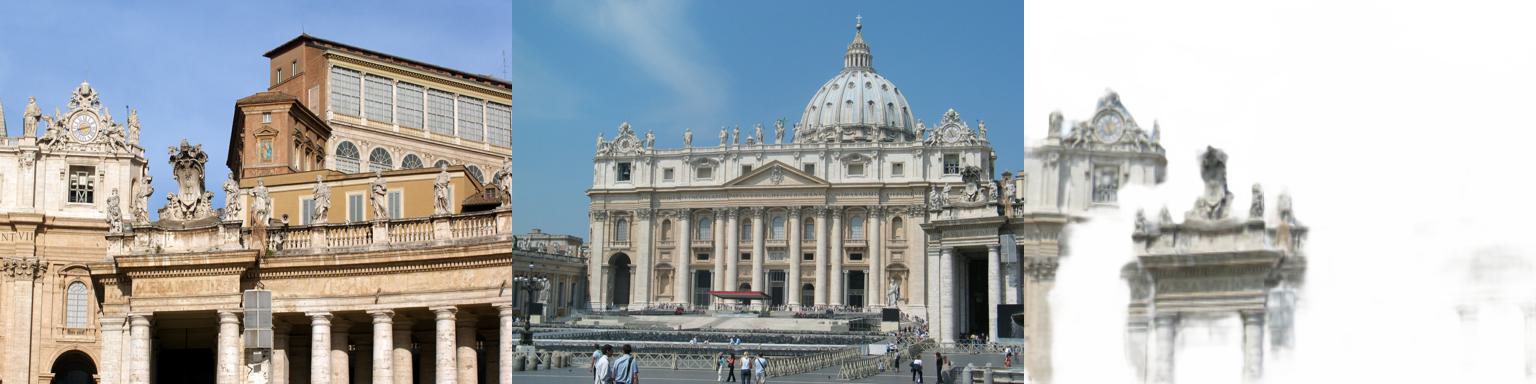}} \\
    \subfloat{\includegraphics[height=2.1cm]{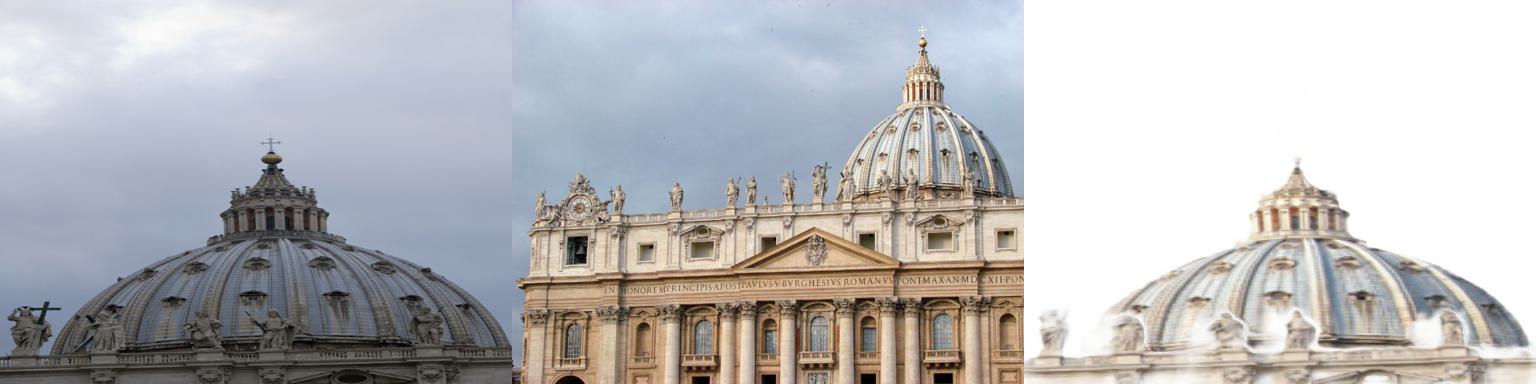}} \,
    \subfloat{\includegraphics[height=2.1cm]{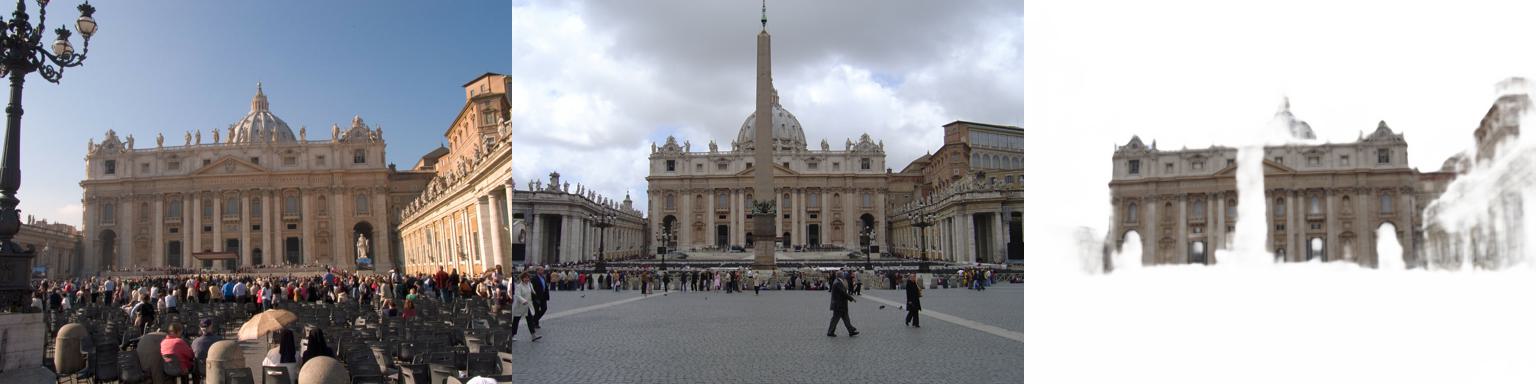}} \\
    \subfloat{\includegraphics[height=2.1cm]{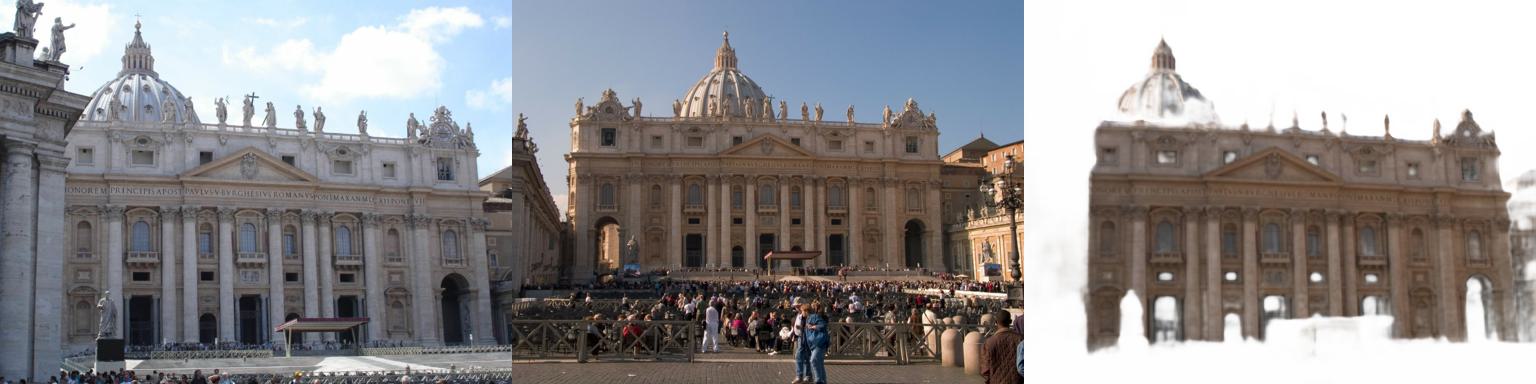}} \,
    \subfloat{\includegraphics[height=2.1cm]{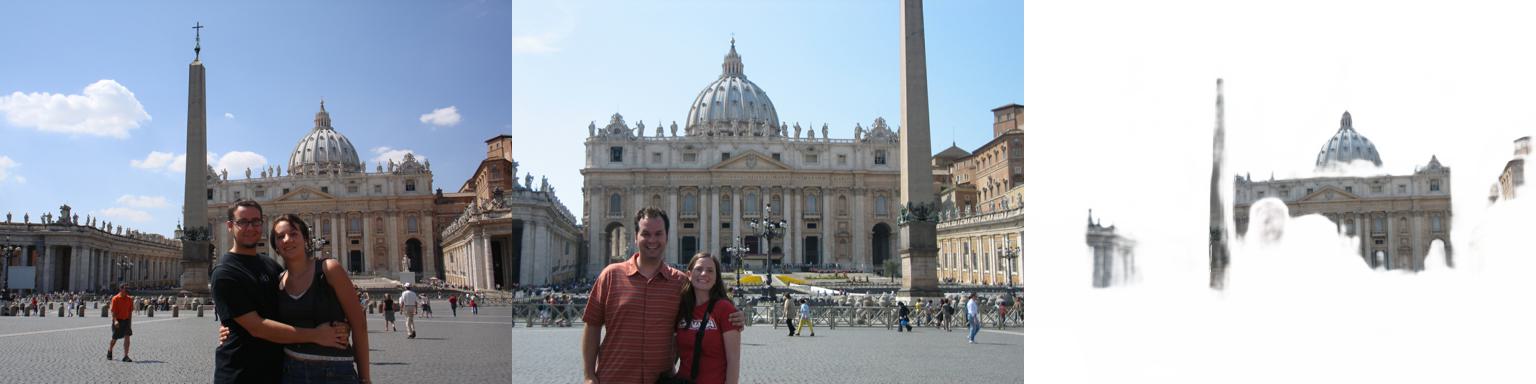}} \\
    \subfloat{\includegraphics[height=2.1cm]{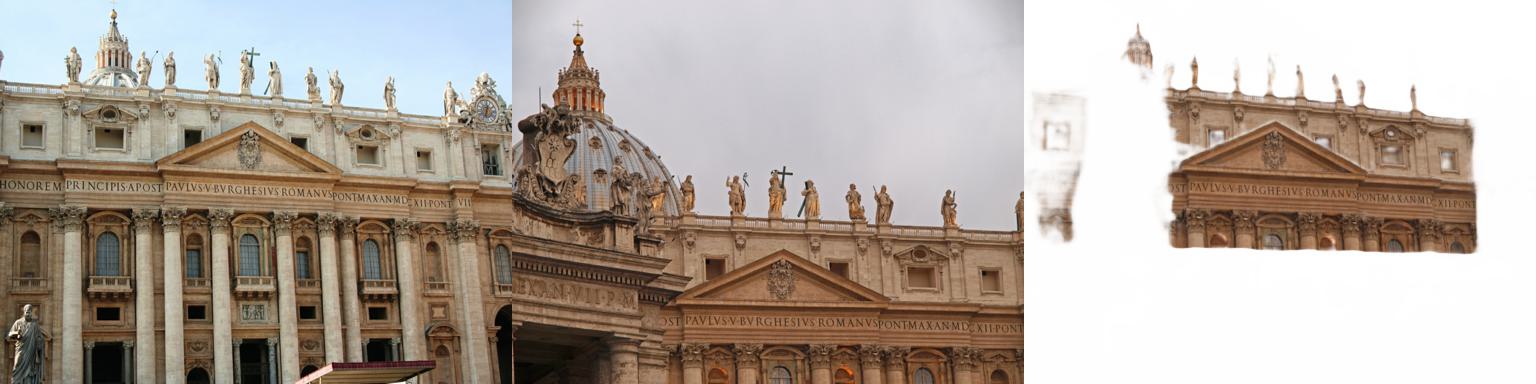}} \,
    \subfloat{\includegraphics[height=2.1cm]{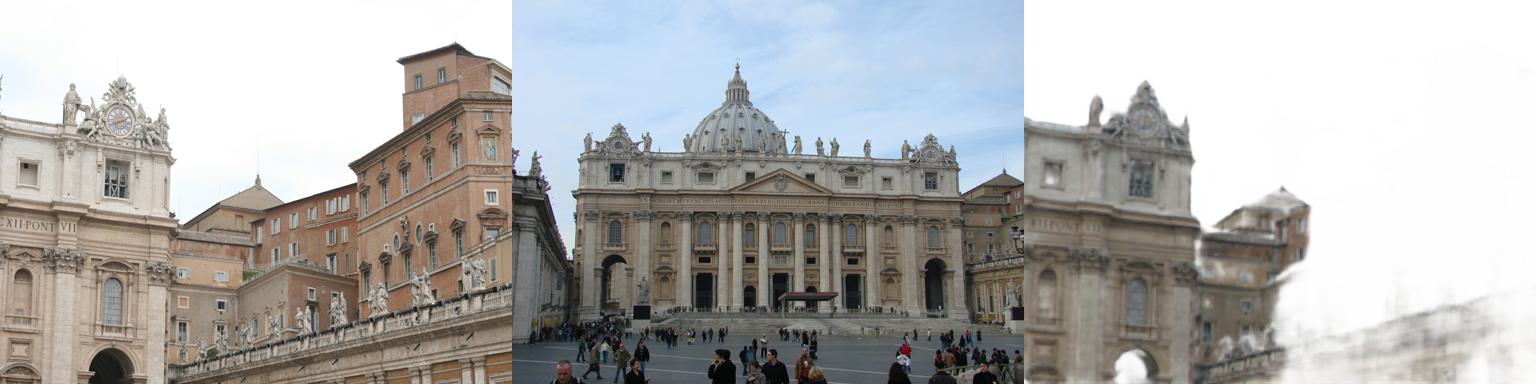}} \\
    \subfloat{\includegraphics[height=2.1cm]{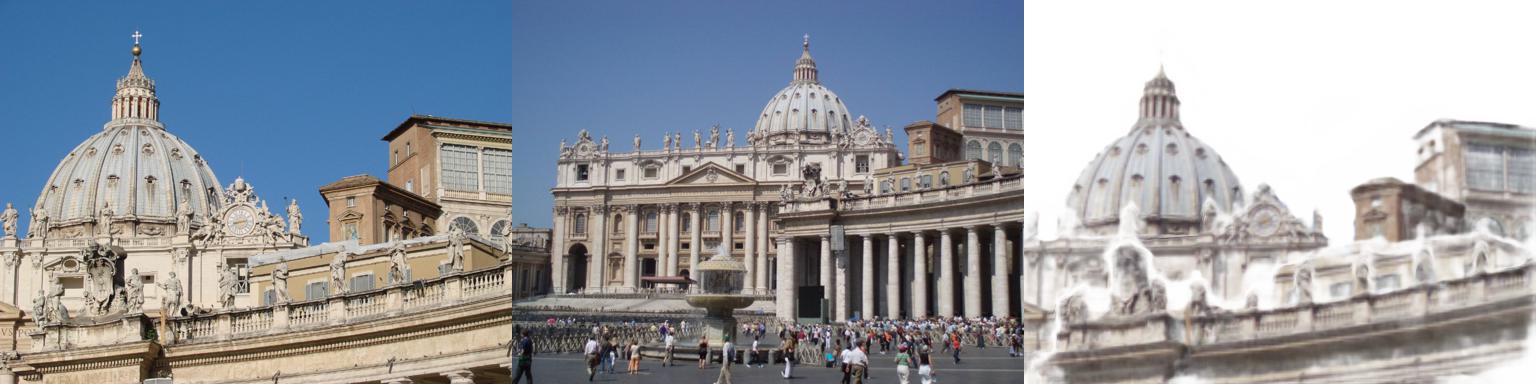}} \,
    \subfloat{\includegraphics[height=2.1cm]{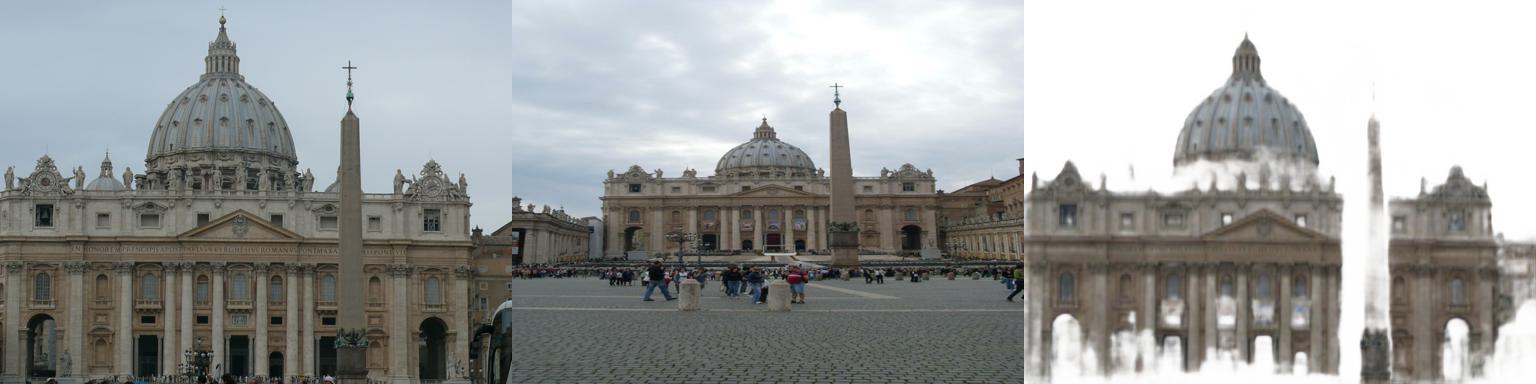}} \\
    \subfloat{\includegraphics[height=2.1cm]{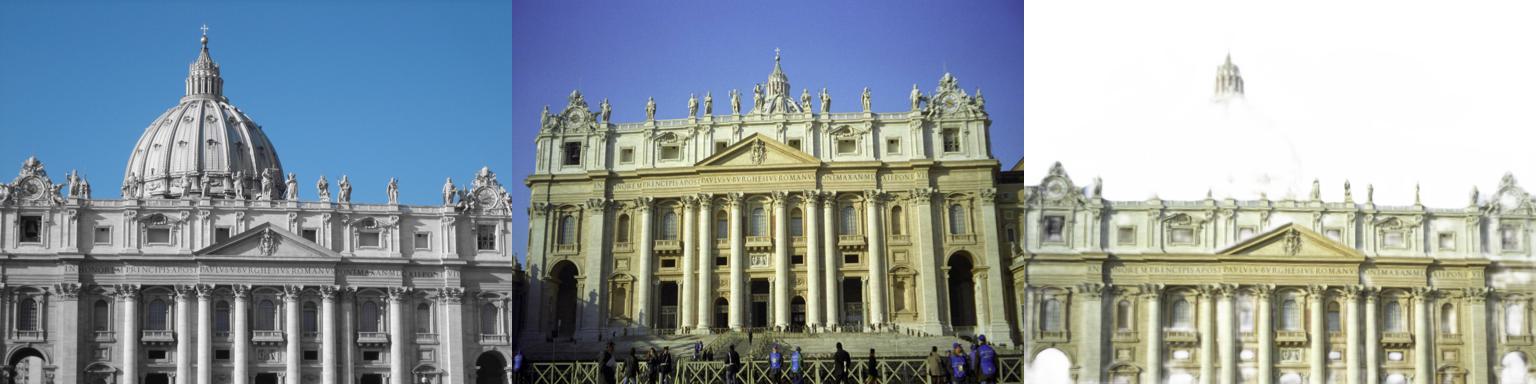}} \,
    \subfloat{\includegraphics[height=2.1cm]{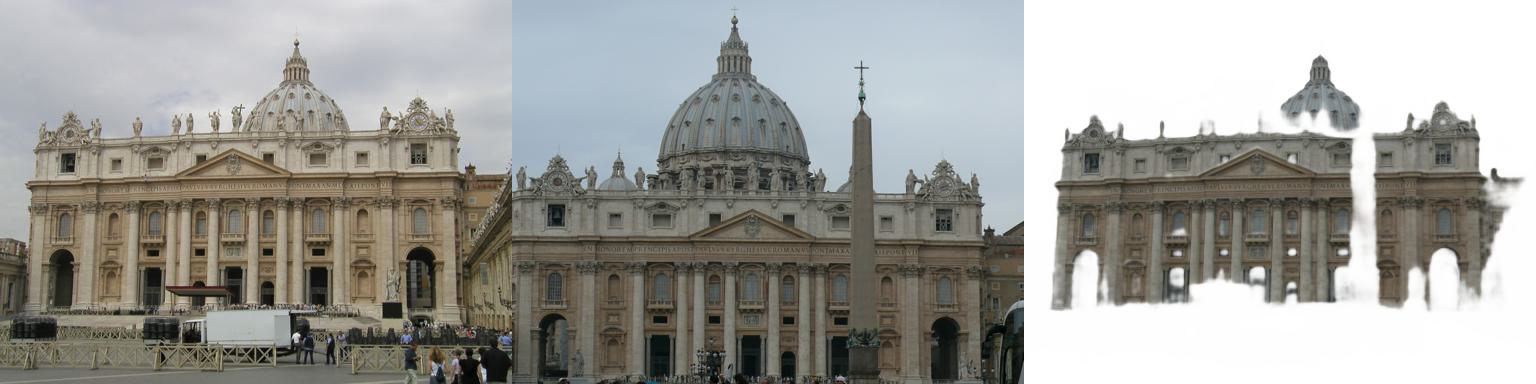}} \\
    \subfloat{\includegraphics[height=2.1cm]{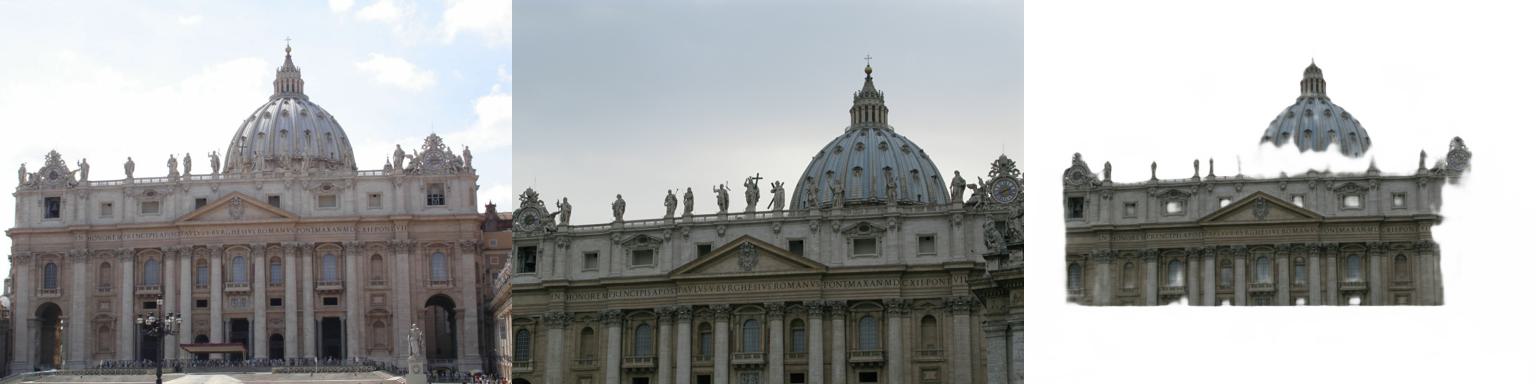}} \,
    \subfloat{\includegraphics[height=2.1cm]{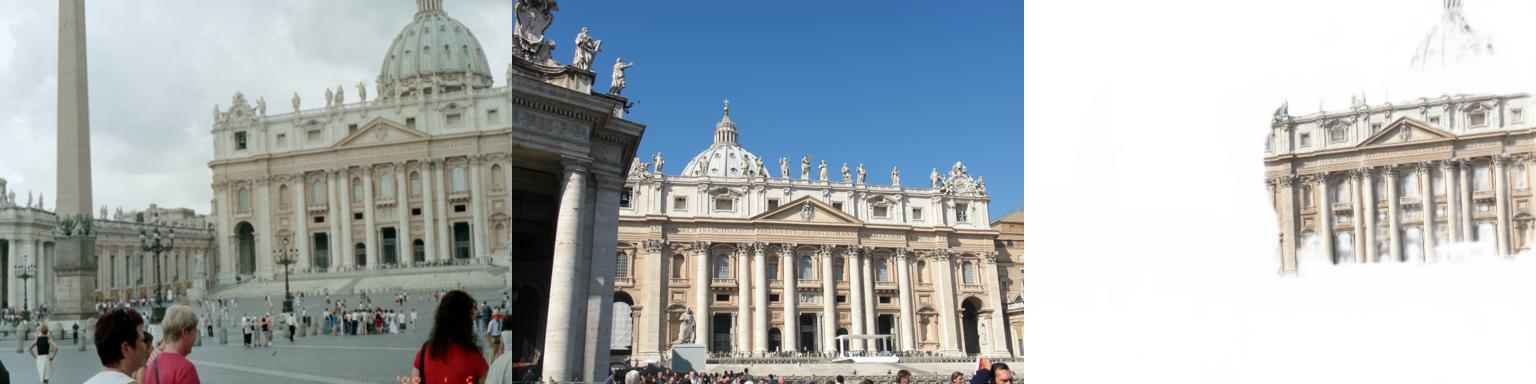}} \\
    \subfloat{\includegraphics[height=2.1cm]{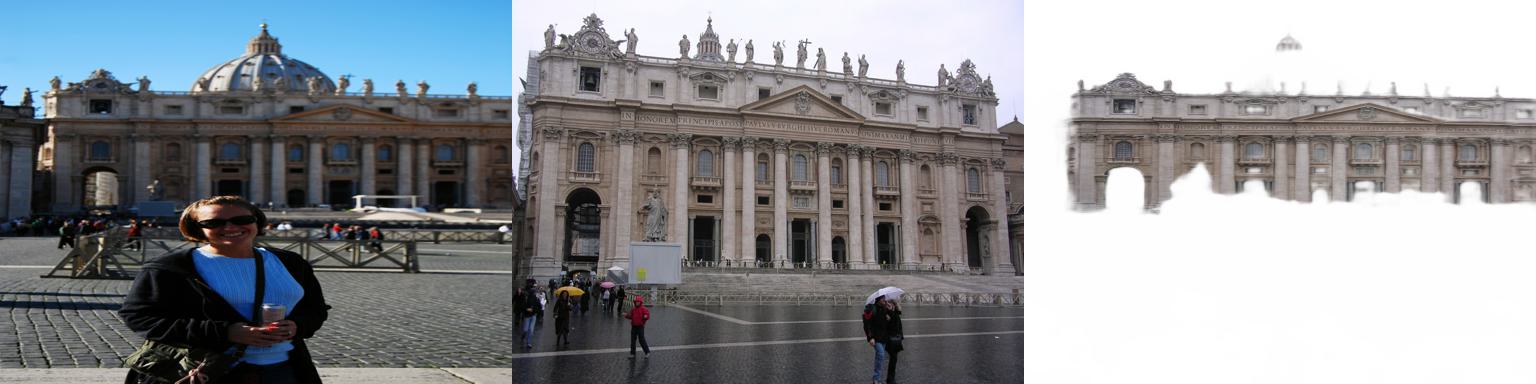}} \,
    \subfloat{\includegraphics[height=2.1cm]{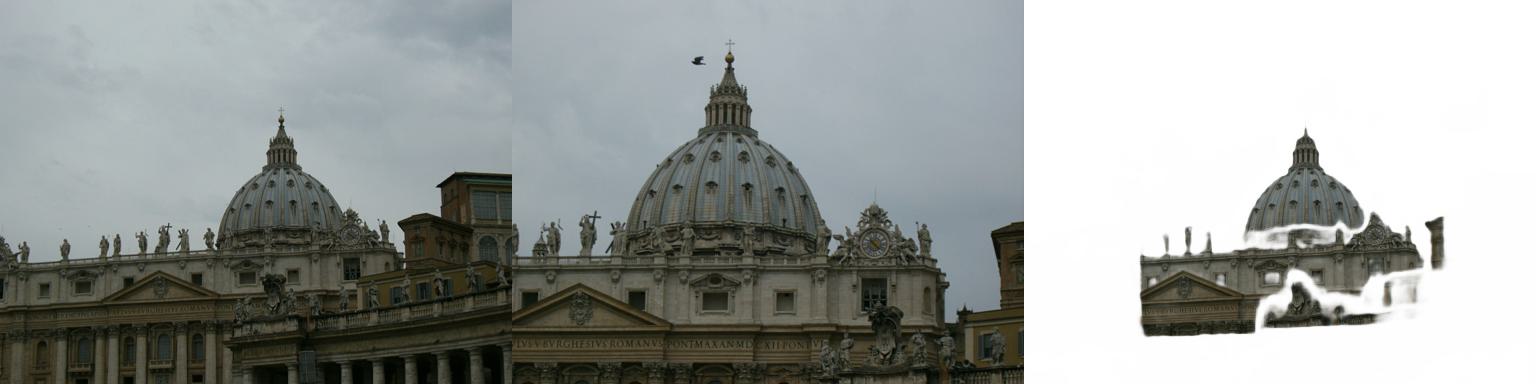}} \\
    \subfloat{\includegraphics[height=2.1cm]{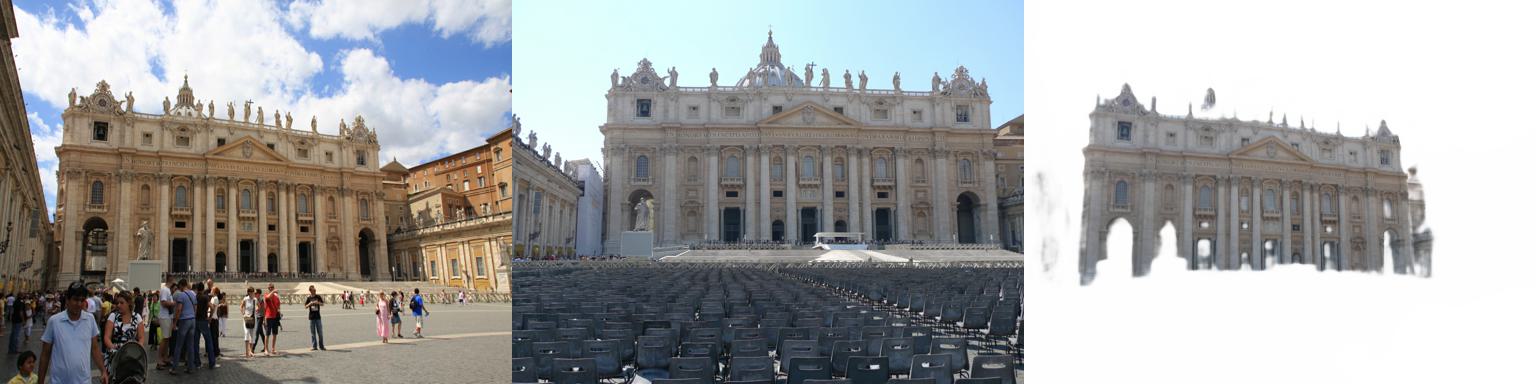}} \,
    \subfloat{\includegraphics[height=2.1cm]{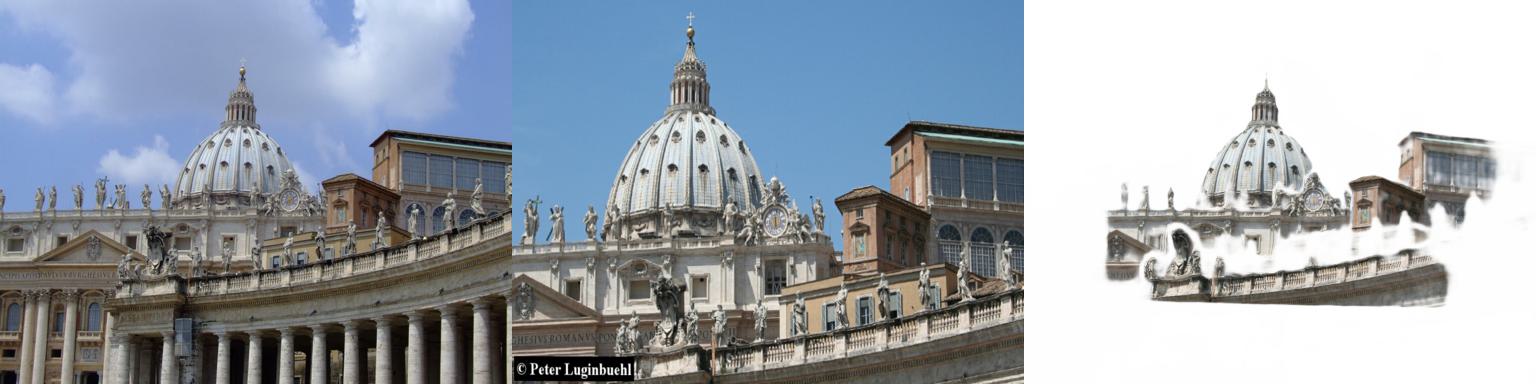}} \\
    \subfloat{\includegraphics[height=2.1cm]{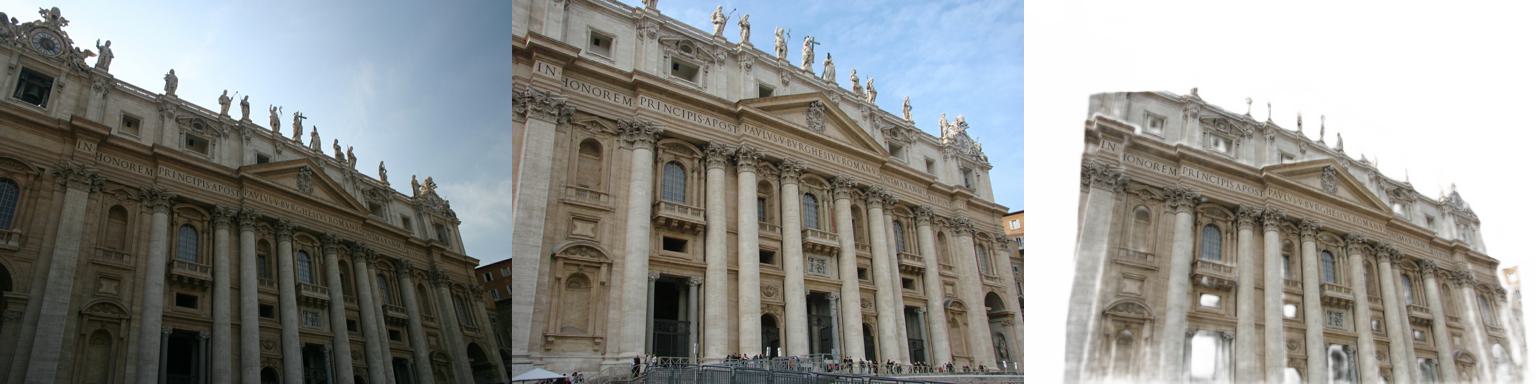}} \,
    \subfloat{\includegraphics[height=2.1cm]{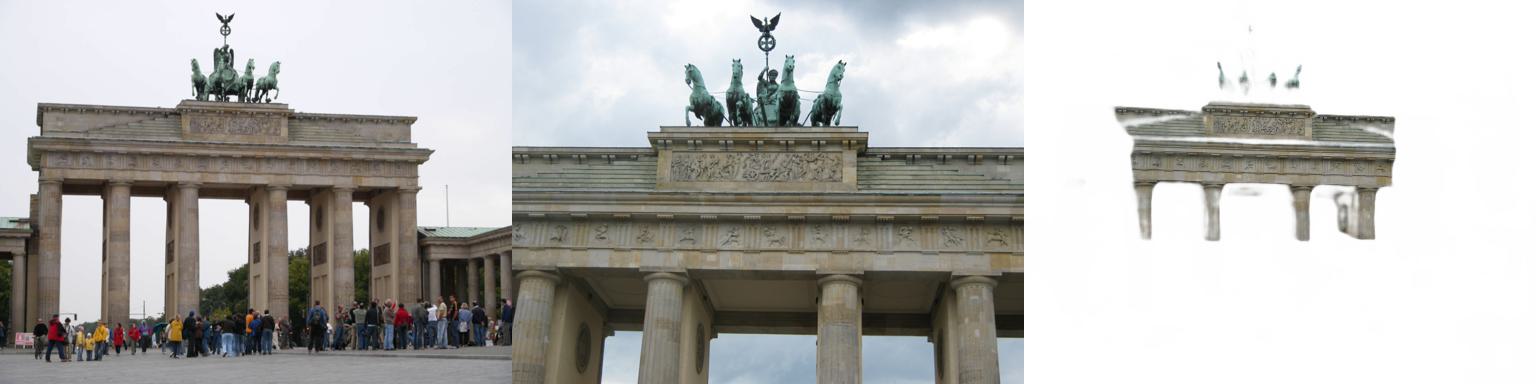}}
    \vspace{0mm}
    \caption{\small
    \textbf{Visual Quality of the Reconstruction on MegaDepth.}
    }
    \vspace{0mm}
    \label{fig:reconstruction2}
\end{figure*}

\begin{figure*}[t!]
    \centering
    \subfloat{\includegraphics[height=2.1cm]{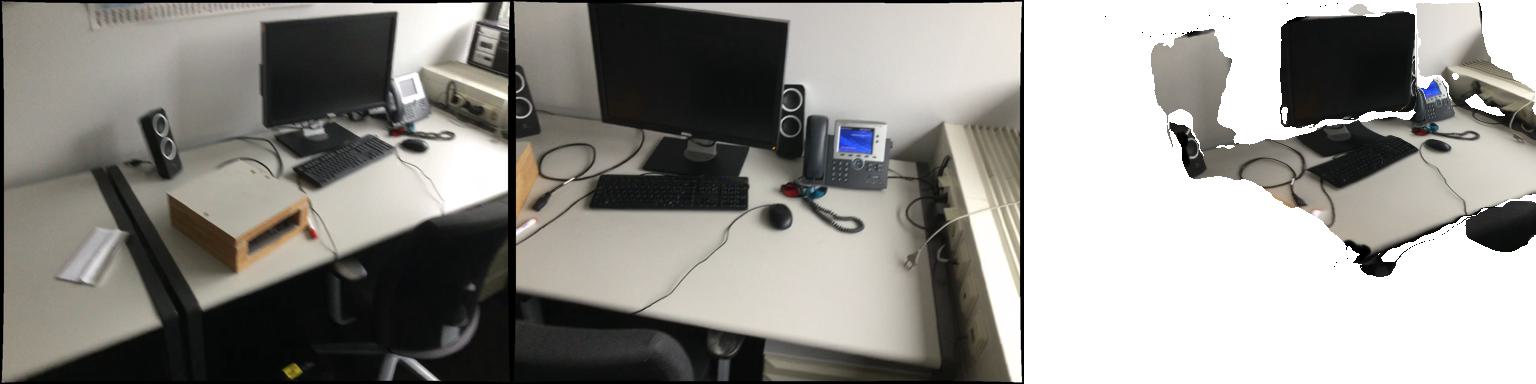}} \,
    \subfloat{\includegraphics[height=2.1cm]{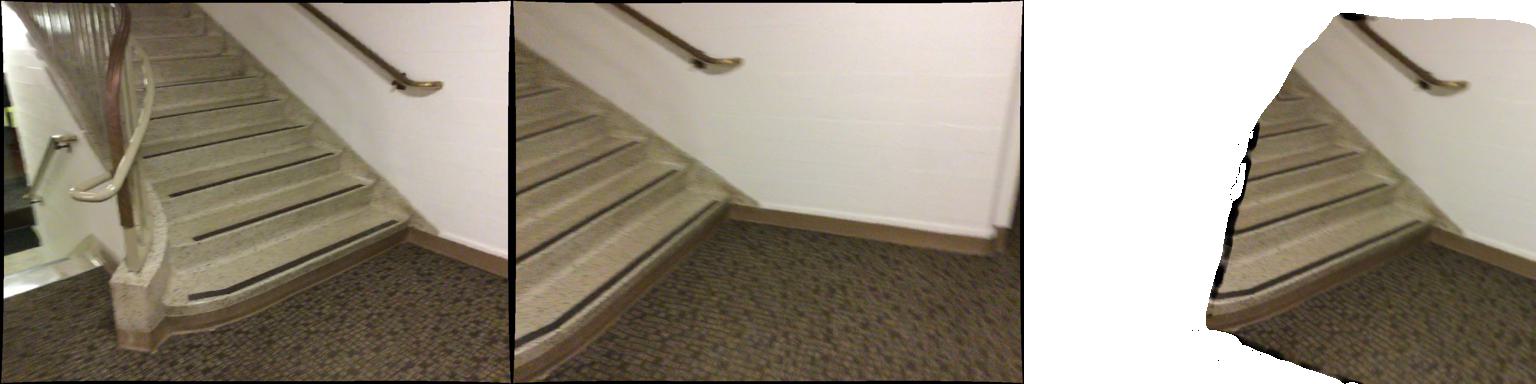}} \\
    \subfloat{\includegraphics[height=2.1cm]{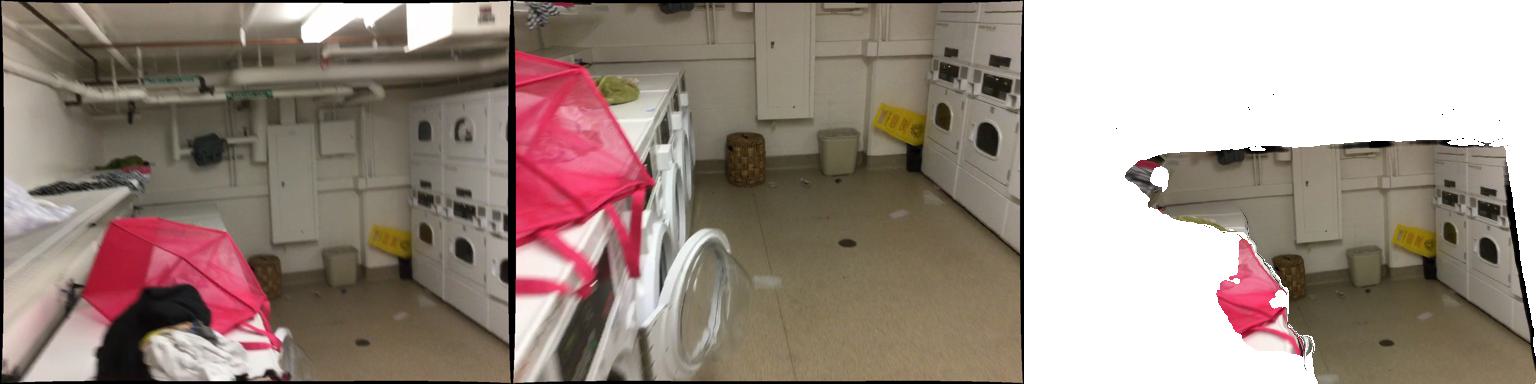}} \,
    \subfloat{\includegraphics[height=2.1cm]{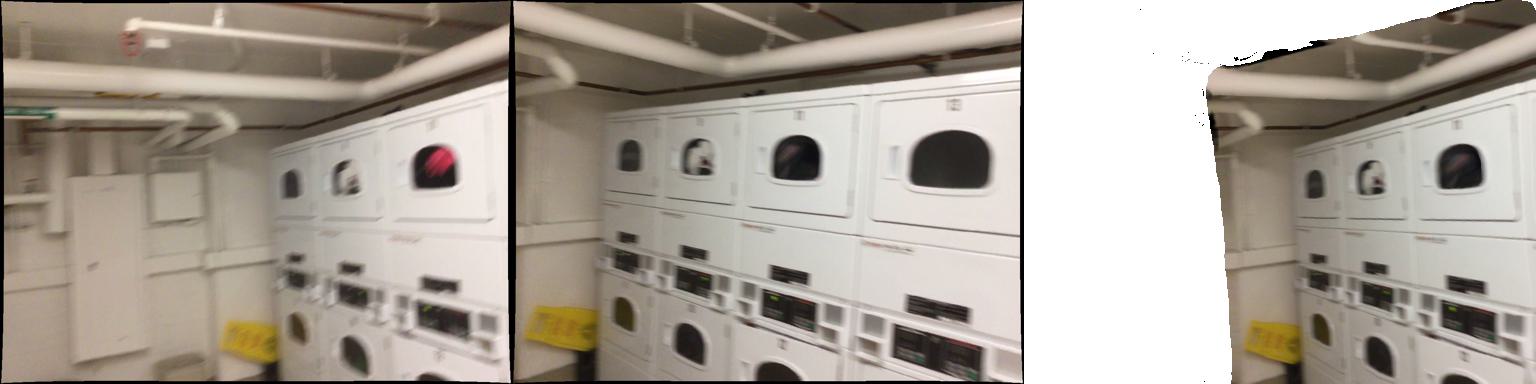}} \\
    \subfloat{\includegraphics[height=2.1cm]{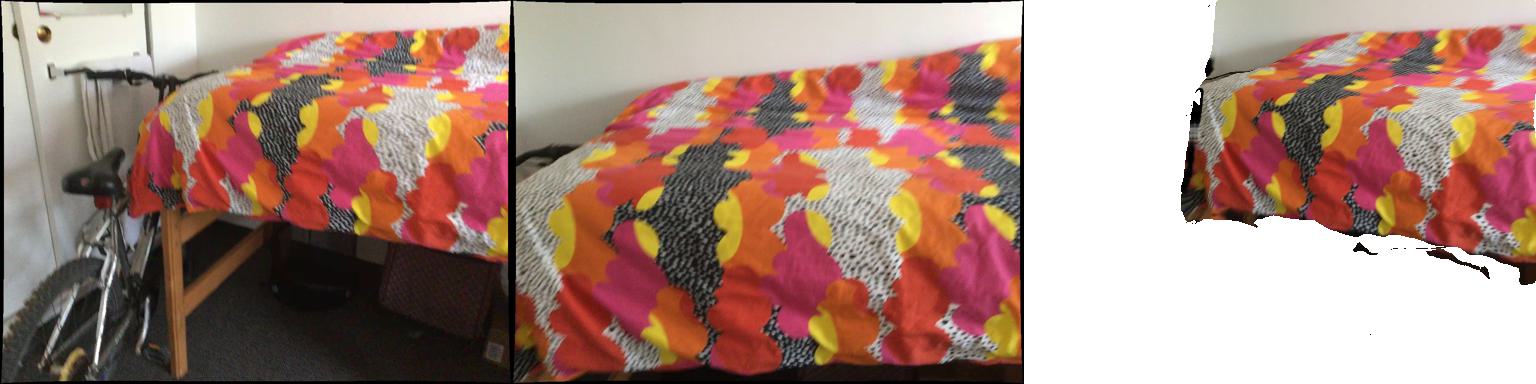}} \,
    \subfloat{\includegraphics[height=2.1cm]{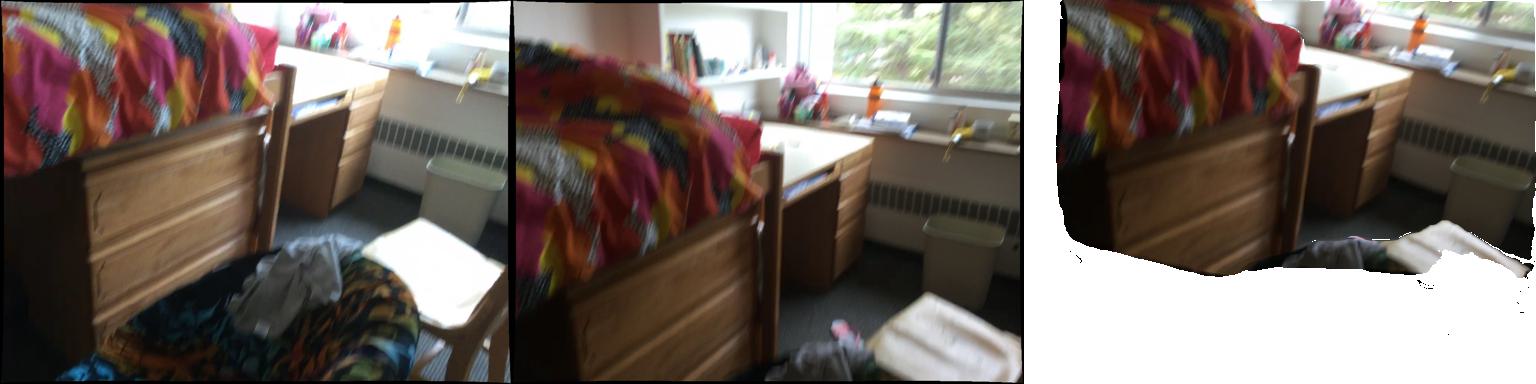}} \\
    \subfloat{\includegraphics[height=2.1cm]{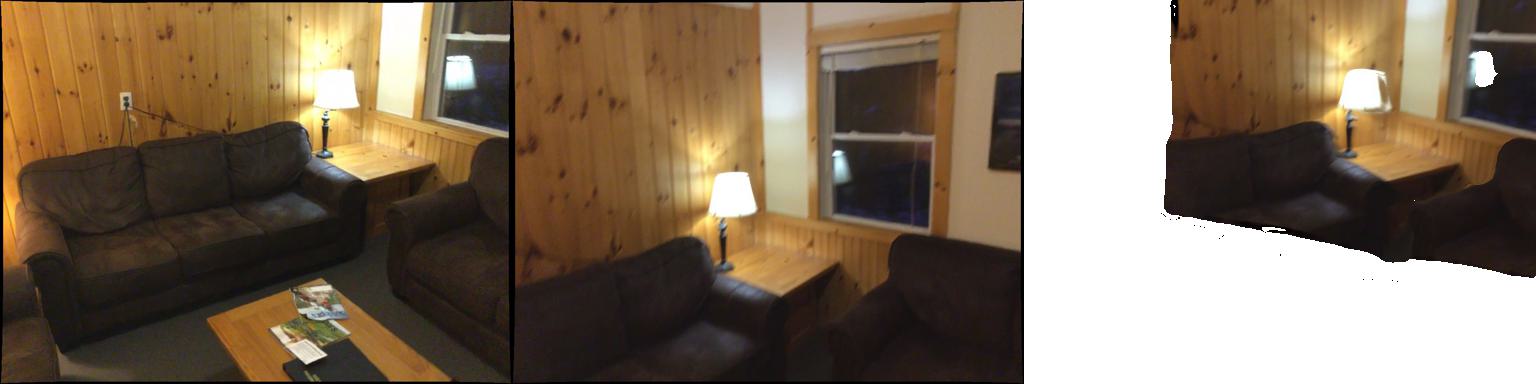}} \,
    \subfloat{\includegraphics[height=2.1cm]{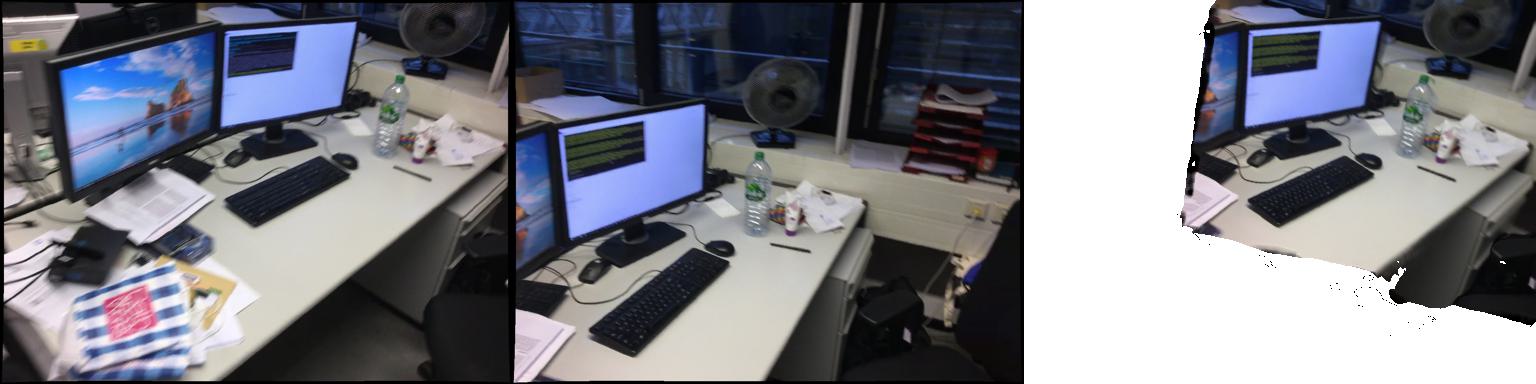}} \\
    \subfloat{\includegraphics[height=2.1cm]{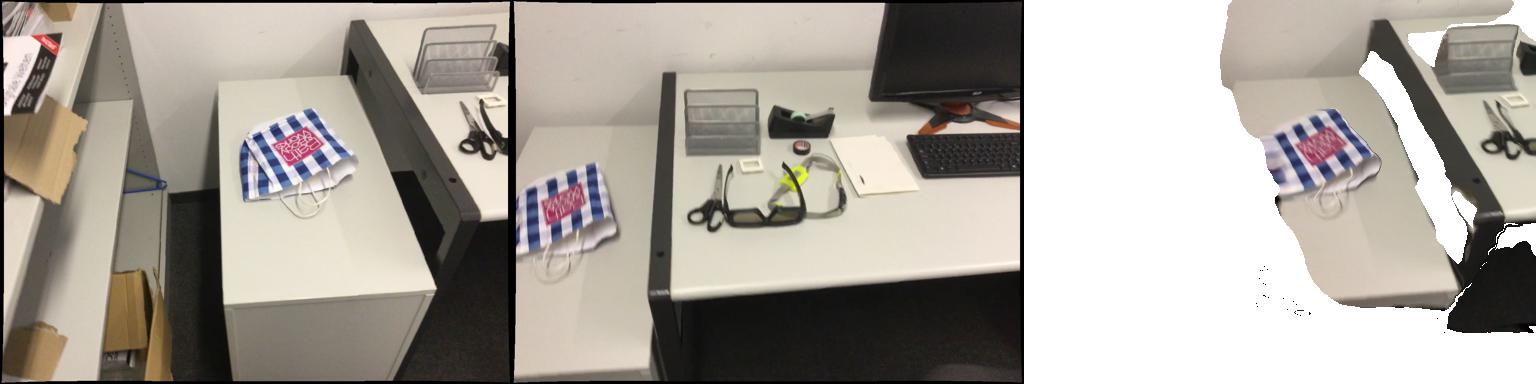}} \,
    \subfloat{\includegraphics[height=2.1cm]{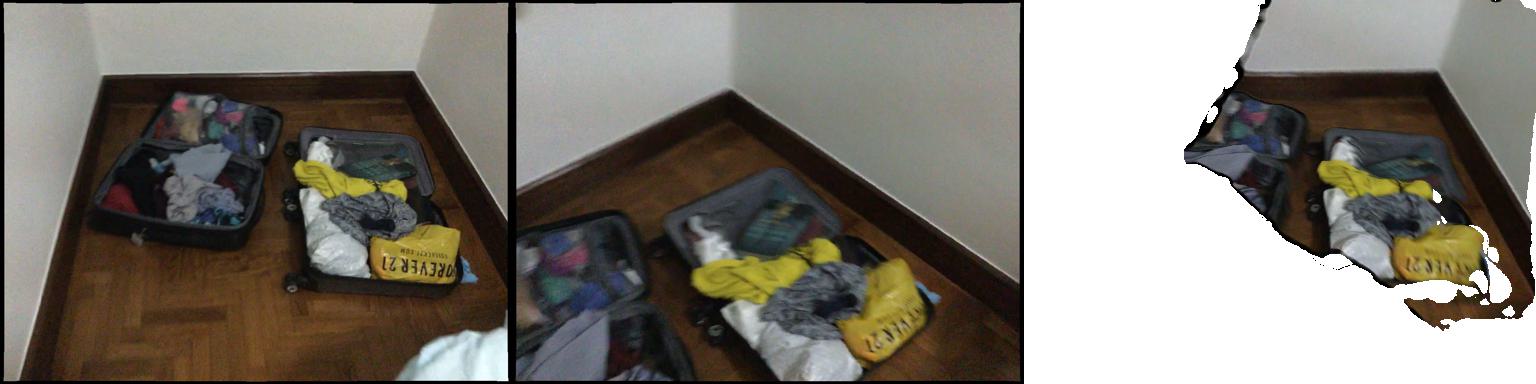}} \\
    \subfloat{\includegraphics[height=2.1cm]{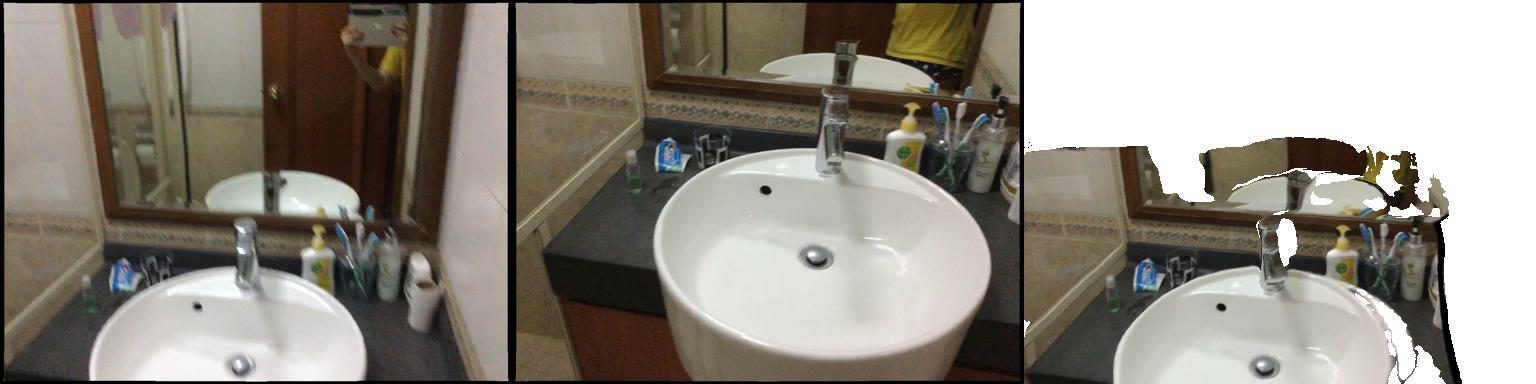}} \,
    \subfloat{\includegraphics[height=2.1cm]{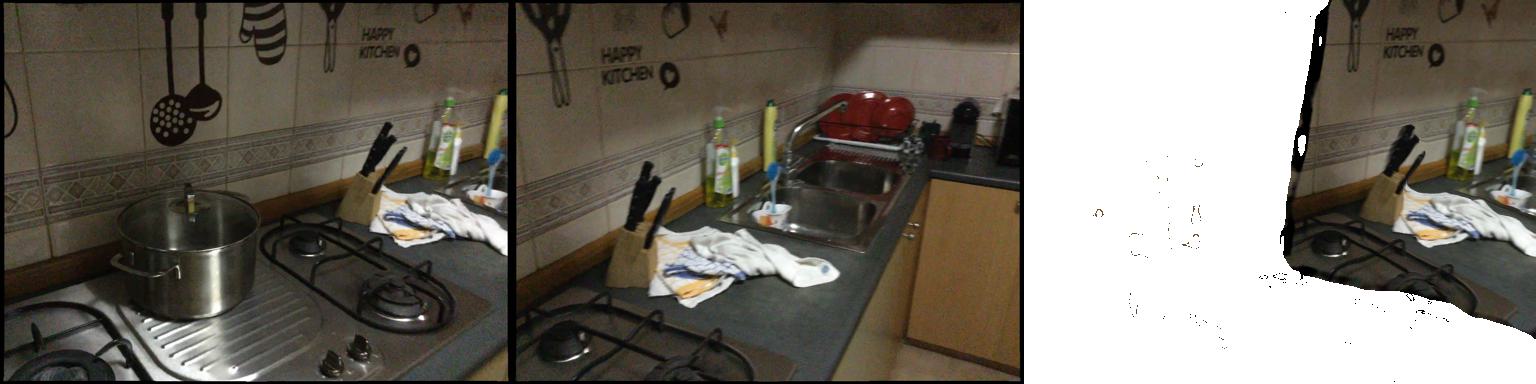}} \\
    \subfloat{\includegraphics[height=2.1cm]{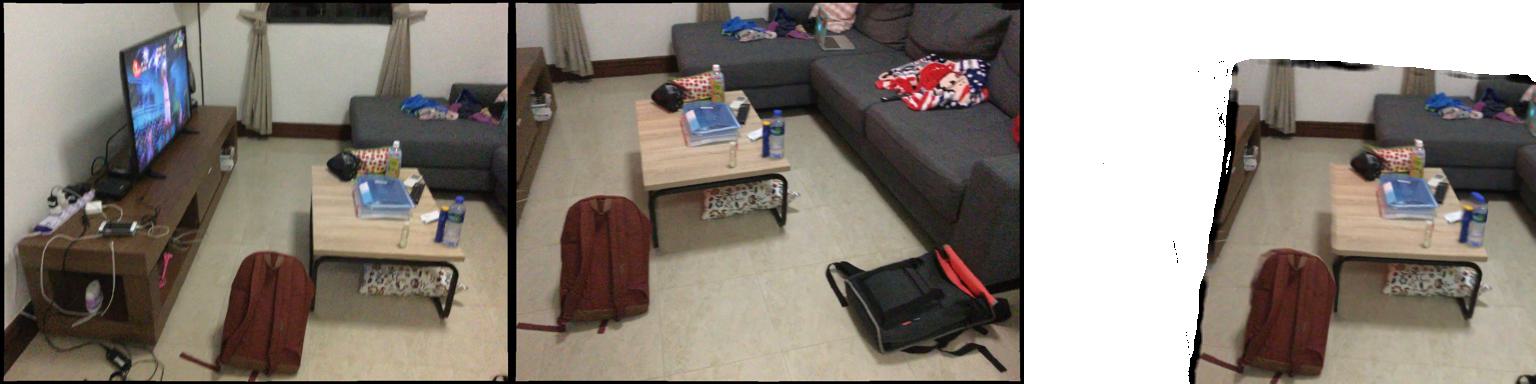}} \,
    \subfloat{\includegraphics[height=2.1cm]{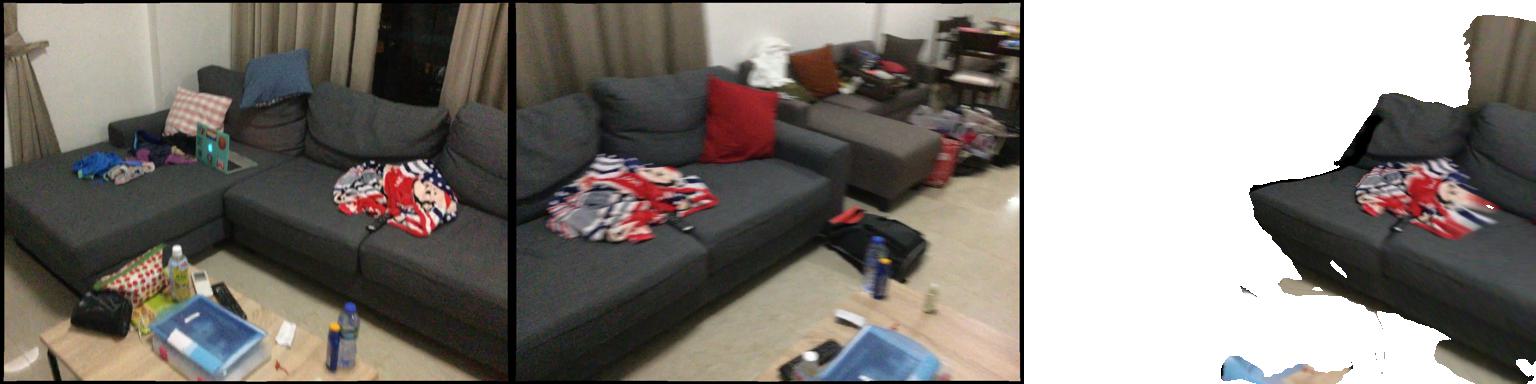}} \\
    \subfloat{\includegraphics[height=2.1cm]{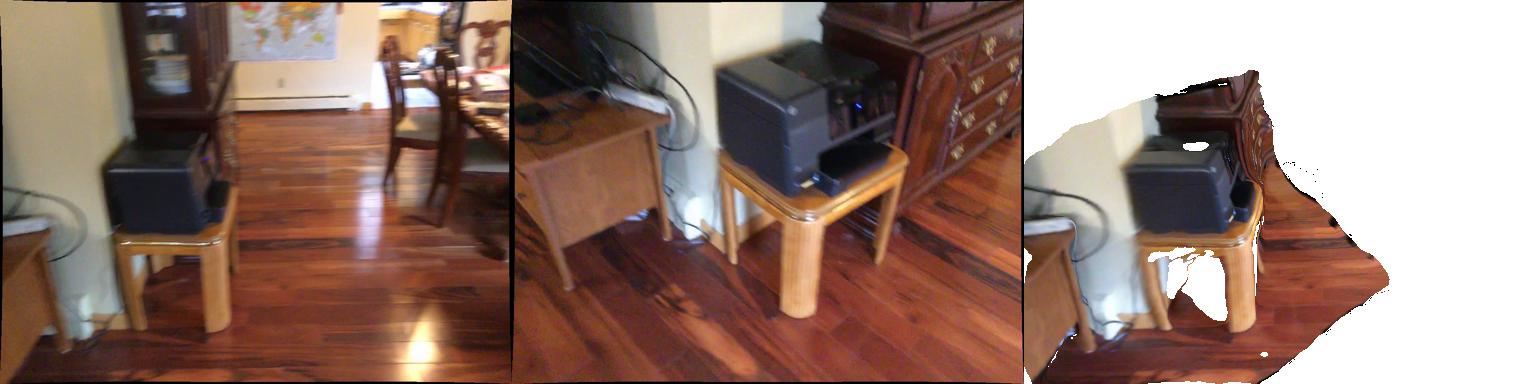}} \,
    \subfloat{\includegraphics[height=2.1cm]{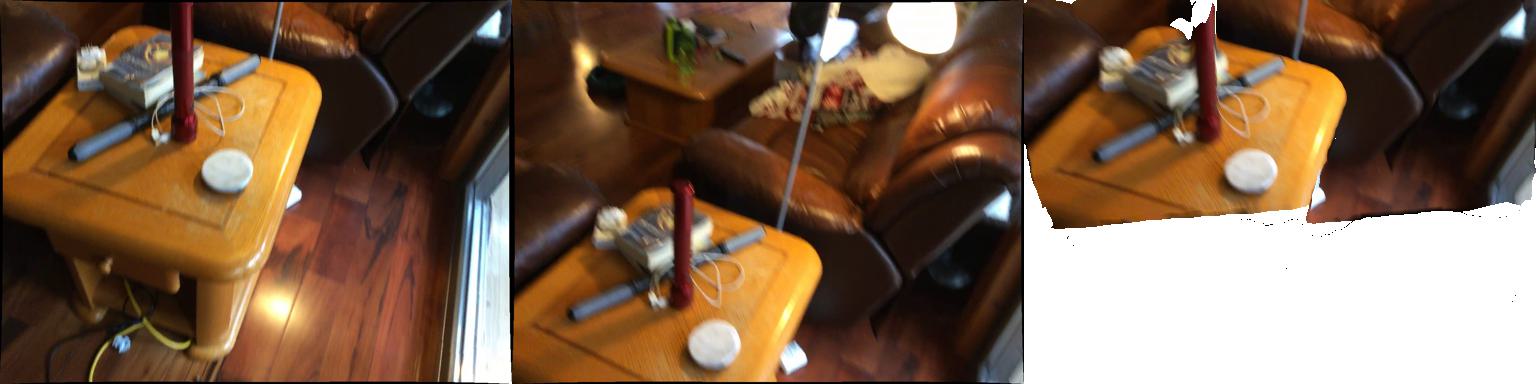}} \\
    \subfloat{\includegraphics[height=2.1cm]{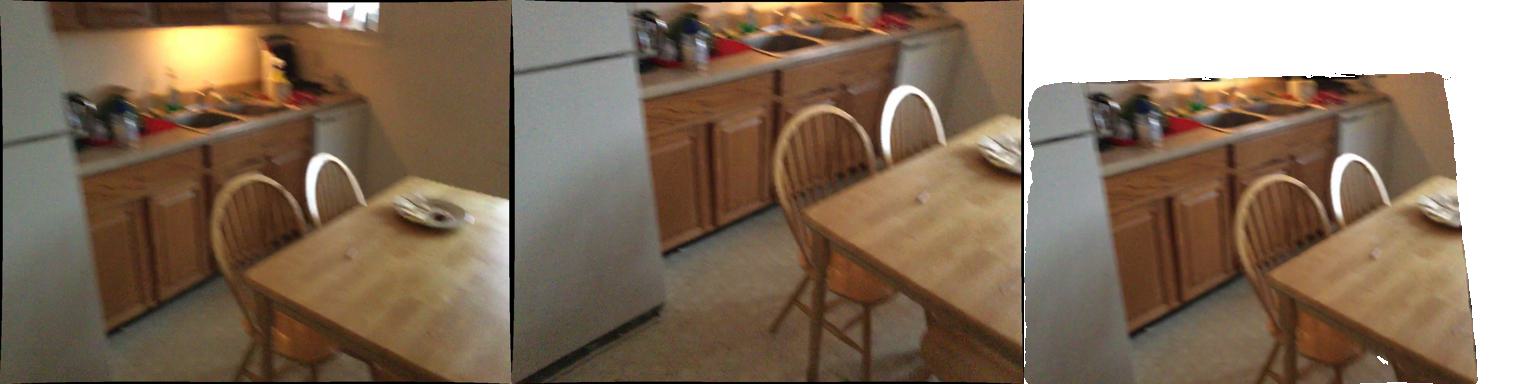}} \,
    \subfloat{\includegraphics[height=2.1cm]{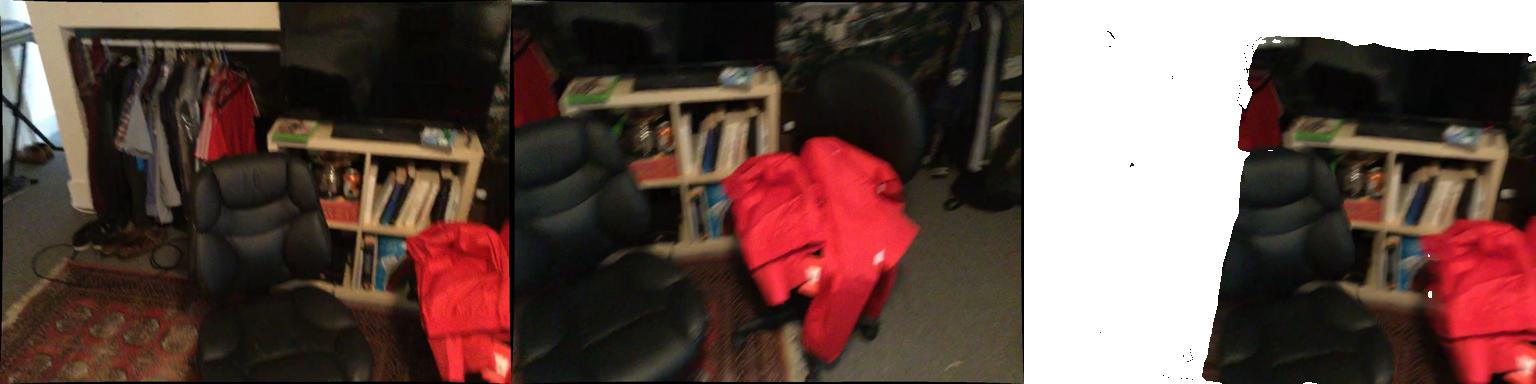}} \\
    \subfloat{\includegraphics[height=2.1cm]{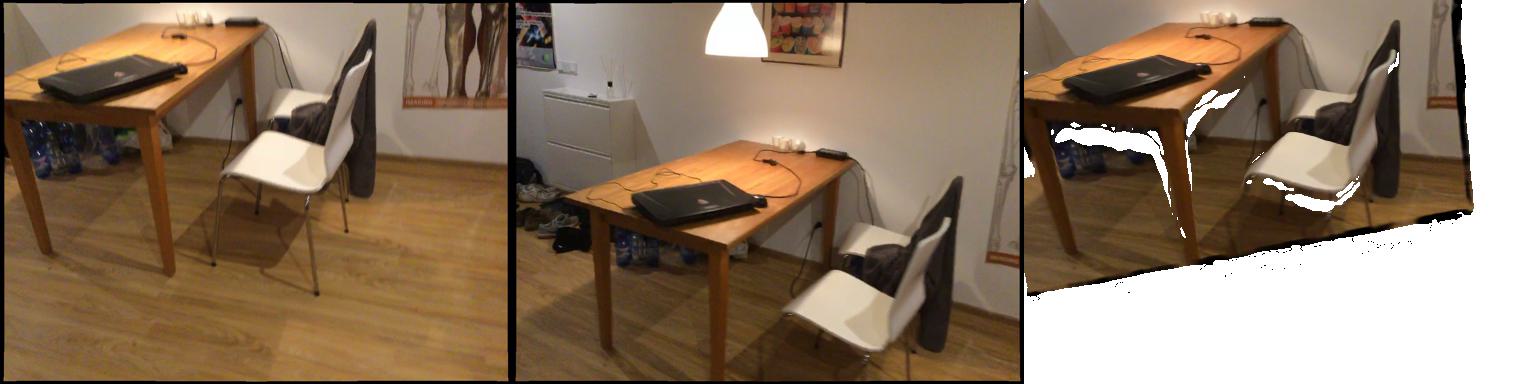}} \,
    \subfloat{\includegraphics[height=2.1cm]{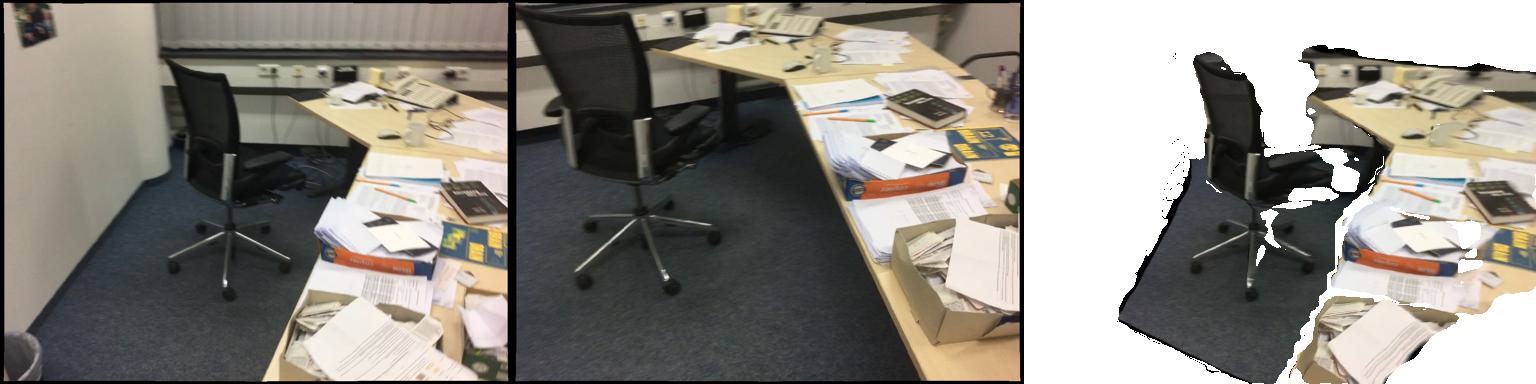}}
    \vspace{0mm}
    \caption{\small
    \textbf{Visual Quality of the Reconstruction on ScanNet.}
    }
    \vspace{0mm}
    \label{fig:reconstruction1}
\end{figure*}

\begin{figure*}[t!]
    \captionsetup[subfigure]{labelformat=empty}
    \centering
    \subfloat{\includegraphics[height=2.3cm]{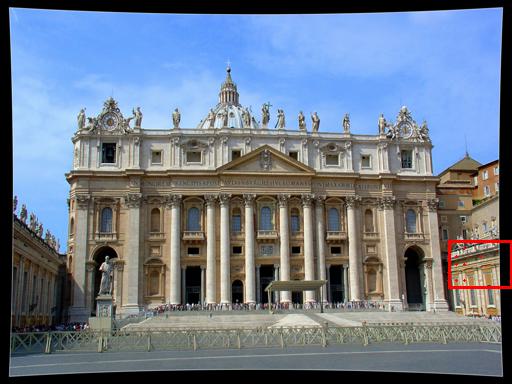}} \,
    \subfloat{\includegraphics[height=2.3cm]{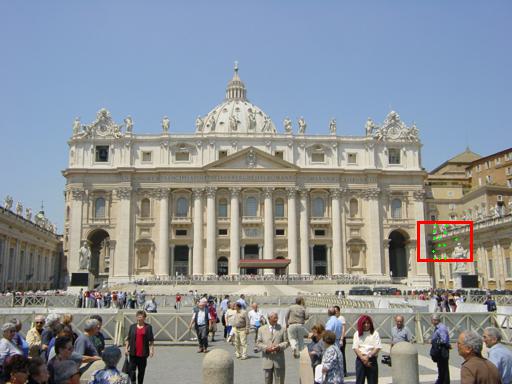}}\,
    \subfloat{\includegraphics[height=2.3cm]{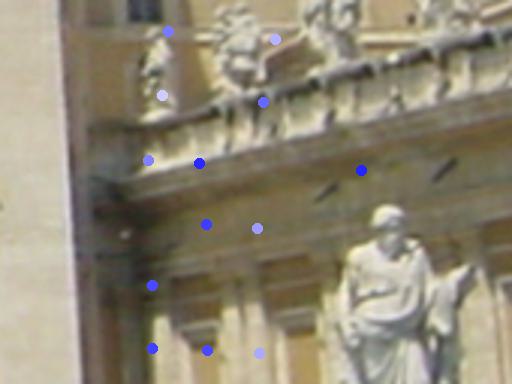}}\,
    \subfloat{\includegraphics[height=2.3cm]{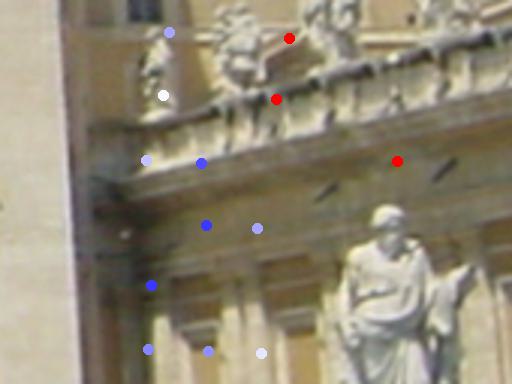}}\,
    \subfloat{\includegraphics[height=2.3cm]{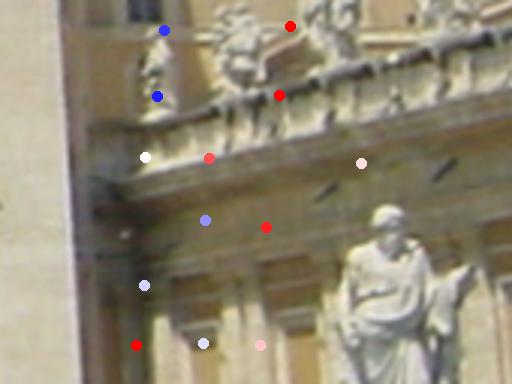}} \\
    \subfloat{\includegraphics[height=2.3cm]{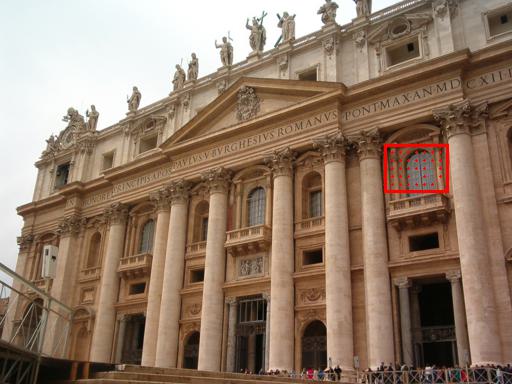}} \,
    \subfloat{\includegraphics[height=2.3cm]{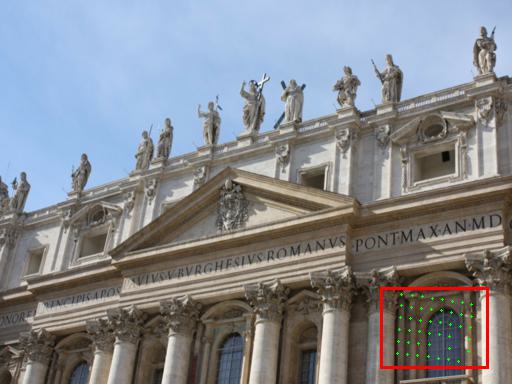}}\,
    \subfloat{\includegraphics[height=2.3cm]{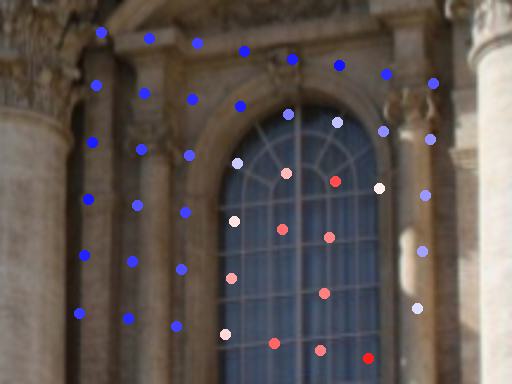}}\,
    \subfloat{\includegraphics[height=2.3cm]{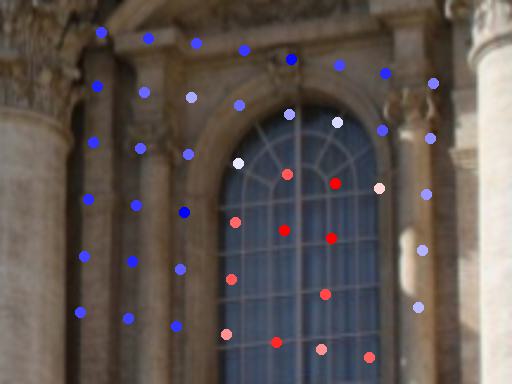}}\,
    \subfloat{\includegraphics[height=2.3cm]{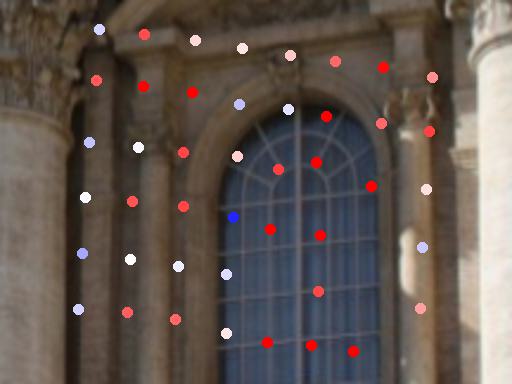}} \\
    \subfloat{\includegraphics[height=2.3cm]{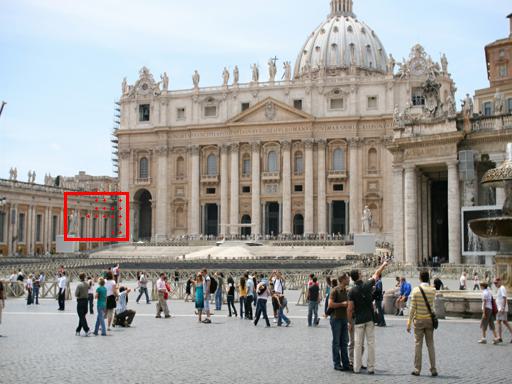}}\,
    \subfloat{\includegraphics[height=2.3cm]{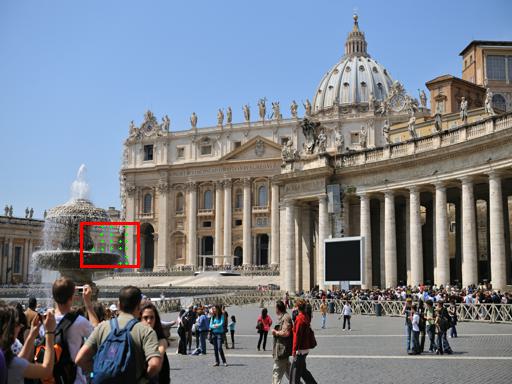}}\,
    \subfloat{\includegraphics[height=2.3cm]{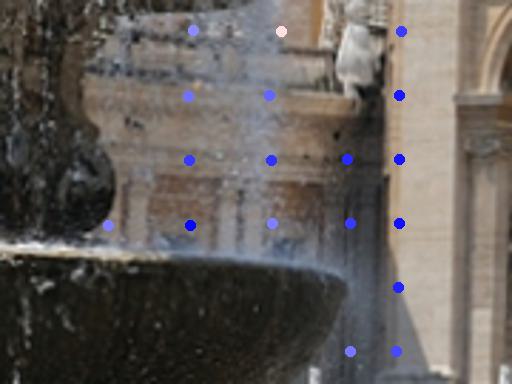}}\,
    \subfloat{\includegraphics[height=2.3cm]{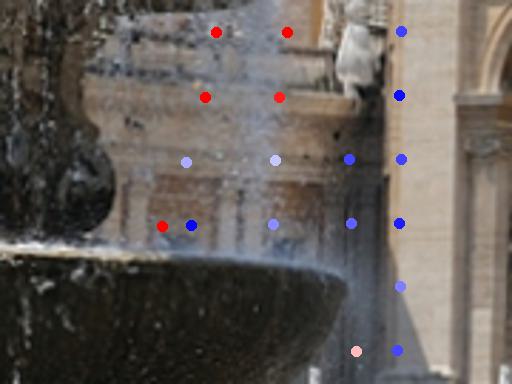}}\,
    \subfloat{\includegraphics[height=2.3cm]{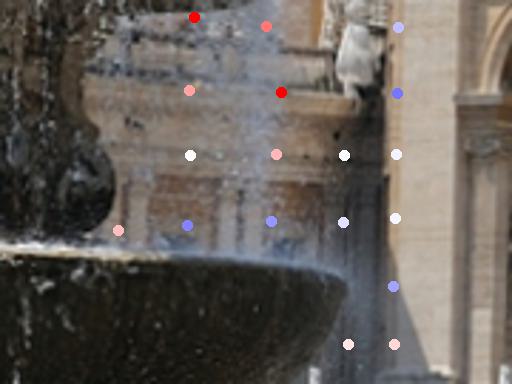}} \\
    \subfloat{\includegraphics[height=2.3cm]{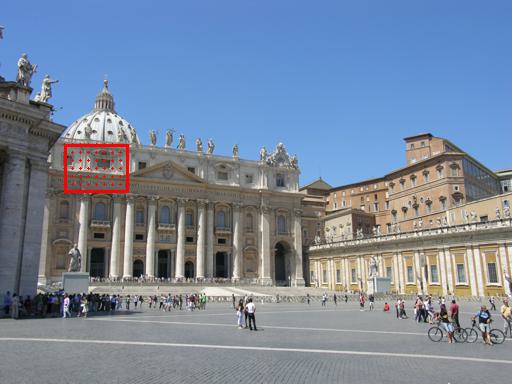}}\,
    \subfloat{\includegraphics[height=2.3cm]{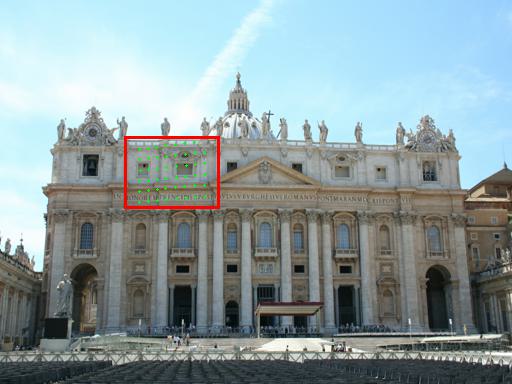}}\,
    \subfloat{\includegraphics[height=2.3cm]{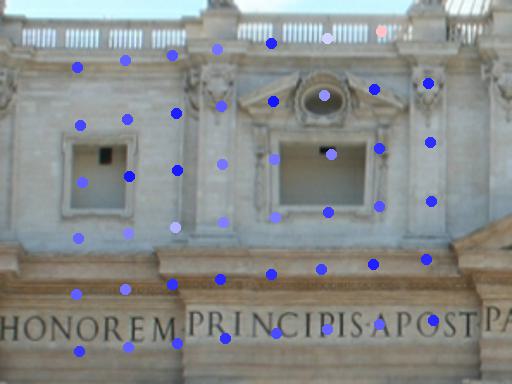}}\,
    \subfloat{\includegraphics[height=2.3cm]{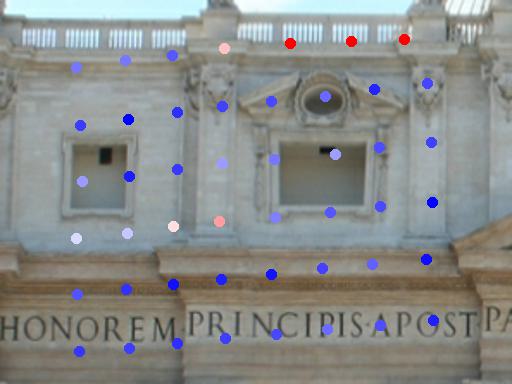}}\,
    \subfloat{\includegraphics[height=2.3cm]{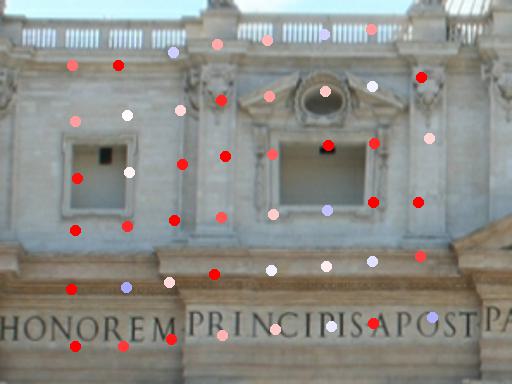}} \\
    \subfloat{\includegraphics[height=2.3cm]{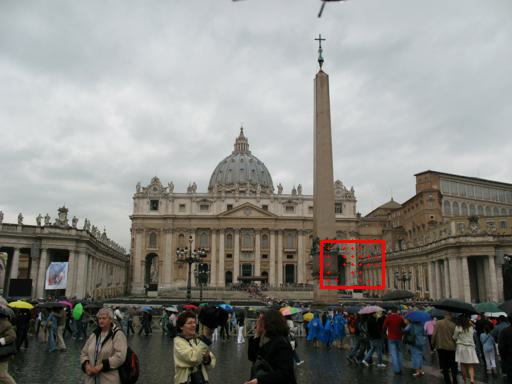}}\,
    \subfloat{\includegraphics[height=2.3cm]{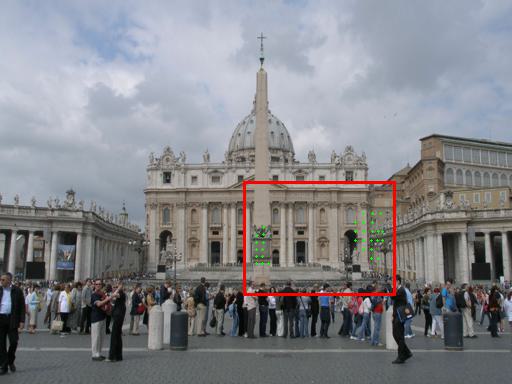}}\,
    \subfloat{\includegraphics[height=2.3cm]{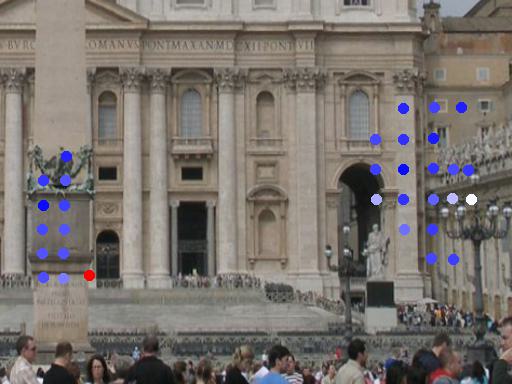}}\,
    \subfloat{\includegraphics[height=2.3cm]{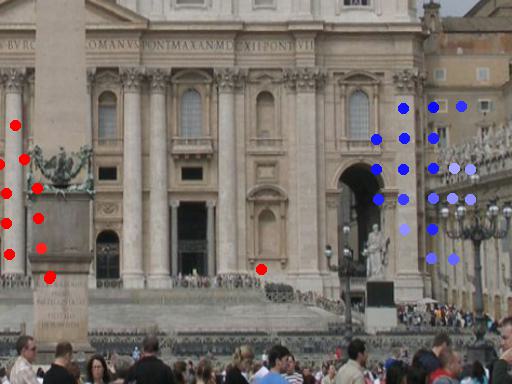}}\,
    \subfloat{\includegraphics[height=2.3cm]{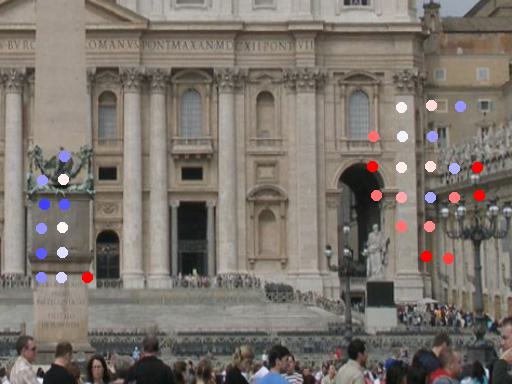}} \\
    \subfloat{\includegraphics[height=2.3cm]{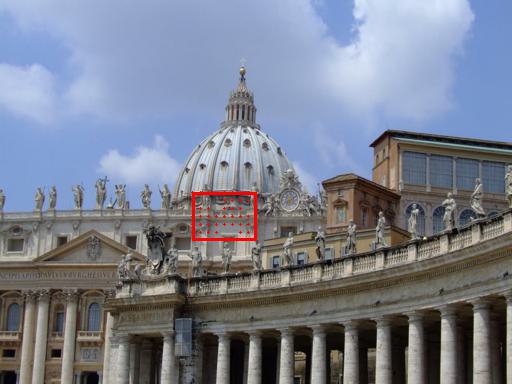}}\,
    \subfloat{\includegraphics[height=2.3cm]{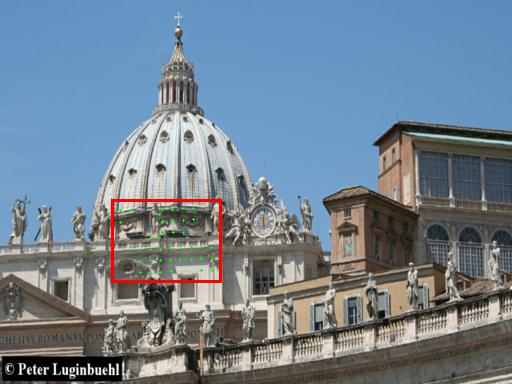}}\,
    \subfloat{\includegraphics[height=2.3cm]{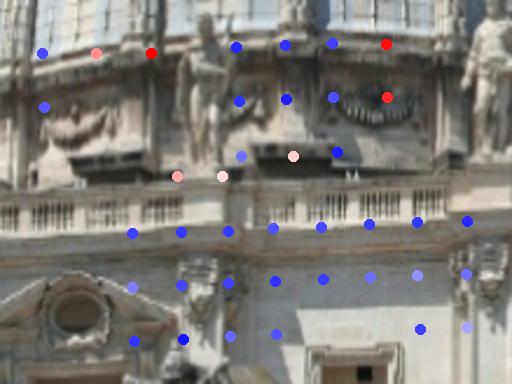}}\,
    \subfloat{\includegraphics[height=2.3cm]{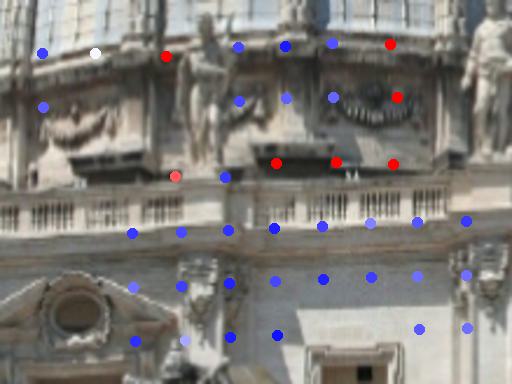}}\,
    \subfloat{\includegraphics[height=2.3cm]{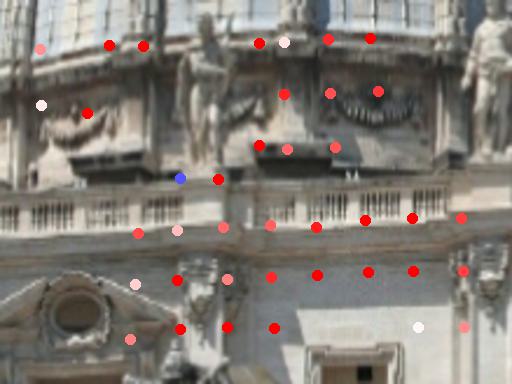}} \\
    \subfloat{\includegraphics[height=2.3cm]{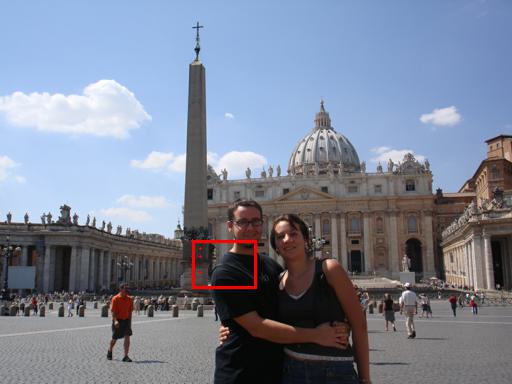}}\,
    \subfloat{\includegraphics[height=2.3cm]{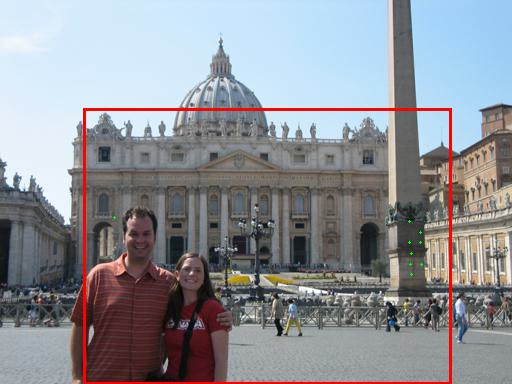}}\,
    \subfloat{\includegraphics[height=2.3cm]{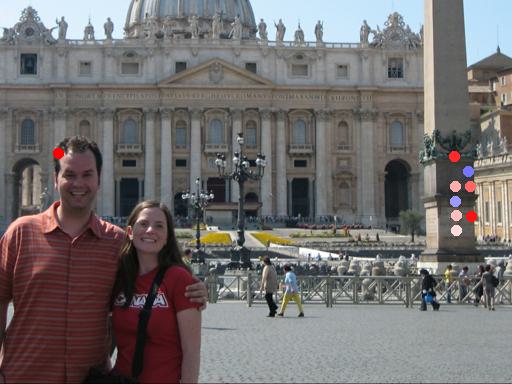}}\,
    \subfloat{\includegraphics[height=2.3cm]{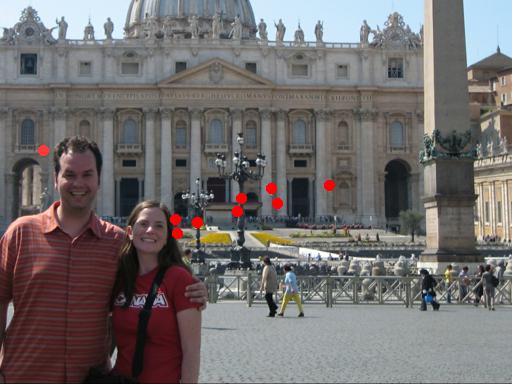}}\,
    \subfloat{\includegraphics[height=2.3cm]{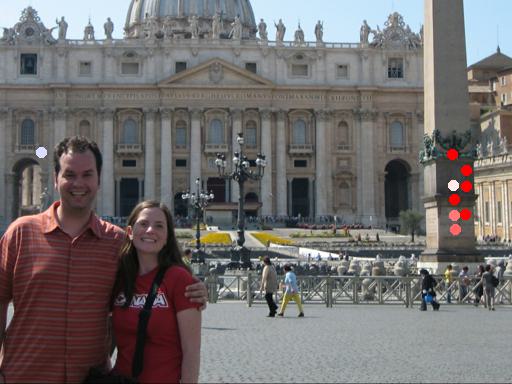}} \\
    \subfloat[(a) Source Frame $\mathbf{I}_1$]{\includegraphics[height=2.3cm]{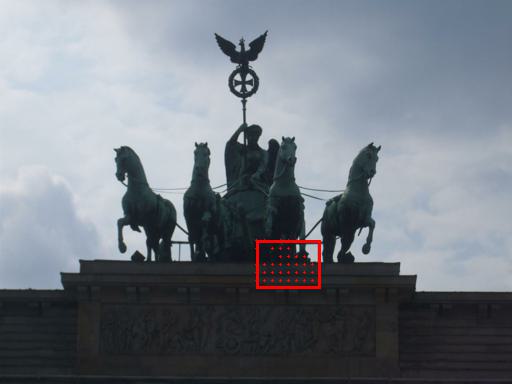}}\,
    \subfloat[(b) Support Frame $\mathbf{I}_2$]{\includegraphics[height=2.3cm]{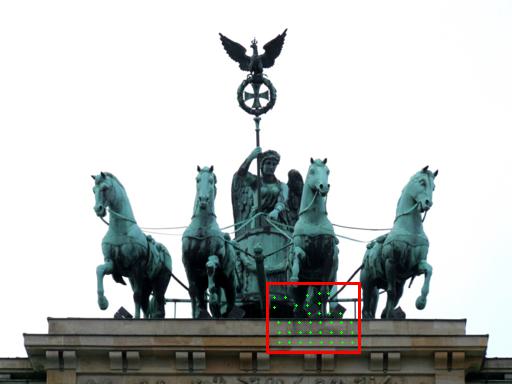}}\,
    \subfloat[(c) PMatch (Ours)]{\includegraphics[height=2.3cm]{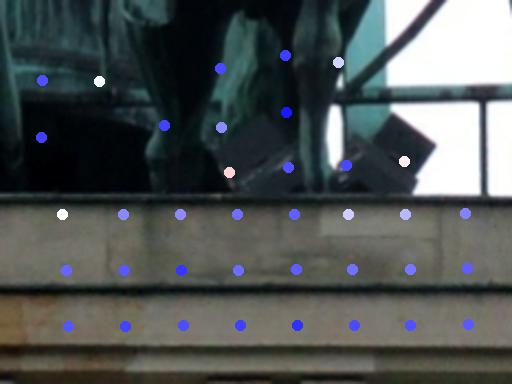}}\,
    \subfloat[(d) DKM~\cite{edstedt2022deep}]{\includegraphics[height=2.3cm]{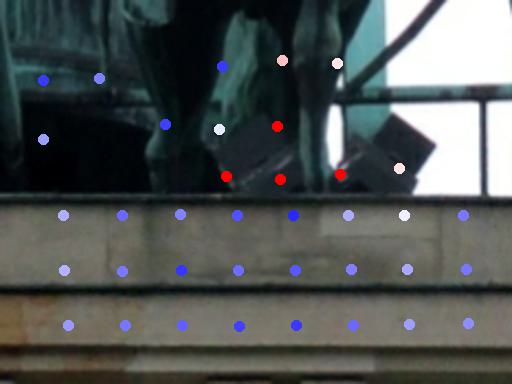}}\,
    \subfloat[(e) LoFTR~\cite{sun2021loftr}]{\includegraphics[height=2.3cm]{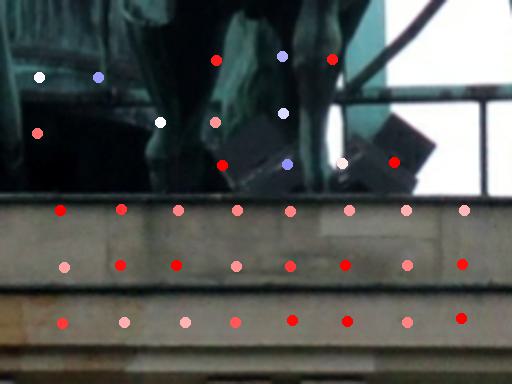}} \\
    \vspace{-1mm}
    \caption{\small 
    \textbf{Visual Comparisons on MegaDepth.}
    We conduct the visual comparison against the SoTA dense~\cite{edstedt2022deep} and sparse~\cite{sun2021loftr} methods on the MegaDepth and the ScanNet datasets.
    The color from \textcolor{blue}{blue} to \textcolor{red}{red} indicates an increment in the end-point-error (L2 error).
    }
    \vspace{-2mm}
    \label{fig:visual_cp1}
\end{figure*}

\begin{figure*}[t!]
    \captionsetup[subfigure]{labelformat=empty}
    \centering
    \subfloat{\includegraphics[height=2.3cm]{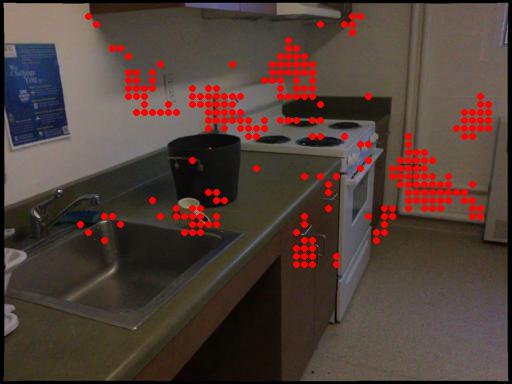}} \,
    \subfloat{\includegraphics[height=2.3cm]{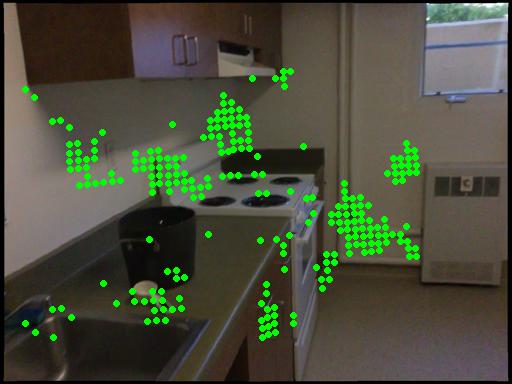}}\,
    \subfloat{\includegraphics[height=2.3cm]{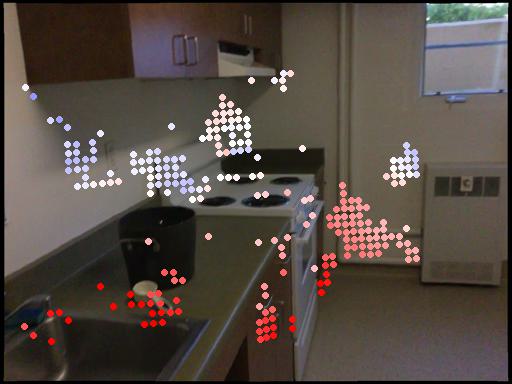}}\,
    \subfloat{\includegraphics[height=2.3cm]{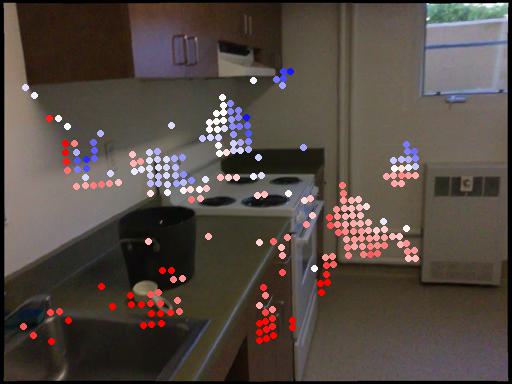}}\,
    \subfloat{\includegraphics[height=2.3cm]{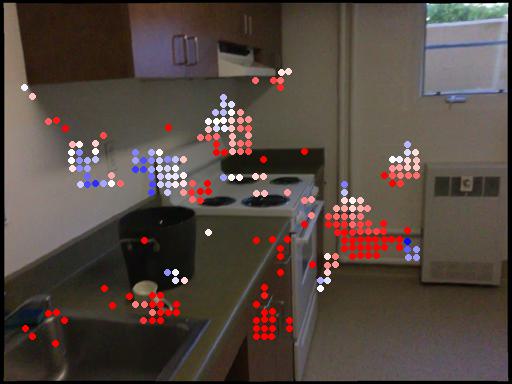}} \\
    \subfloat{\includegraphics[height=2.3cm]{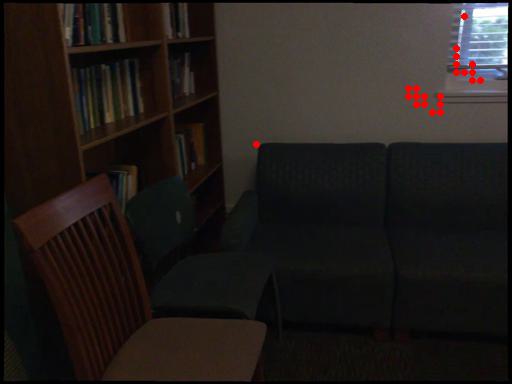}} \,
    \subfloat{\includegraphics[height=2.3cm]{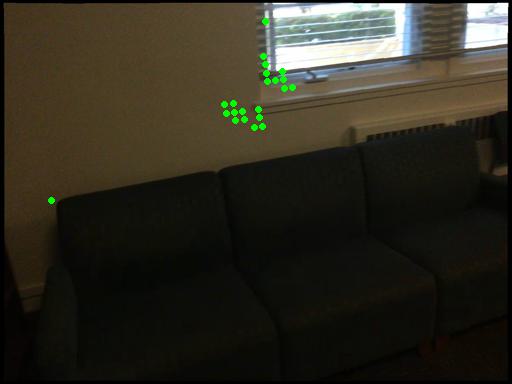}}\,
    \subfloat{\includegraphics[height=2.3cm]{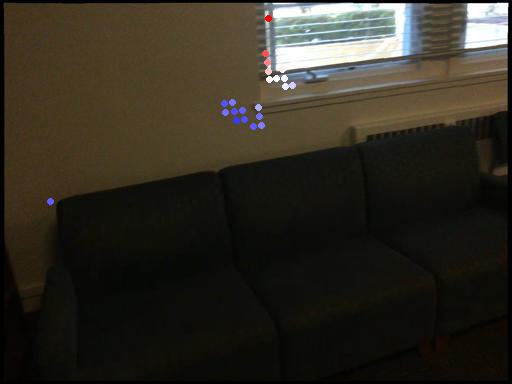}}\,
    \subfloat{\includegraphics[height=2.3cm]{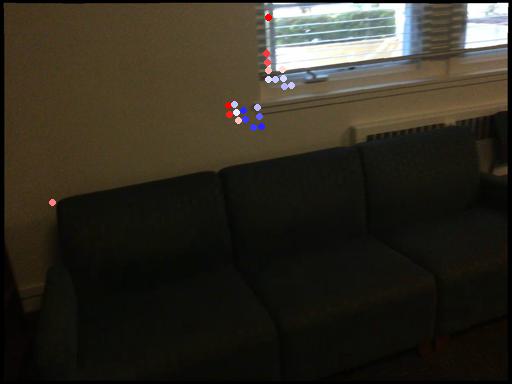}}\,
    \subfloat{\includegraphics[height=2.3cm]{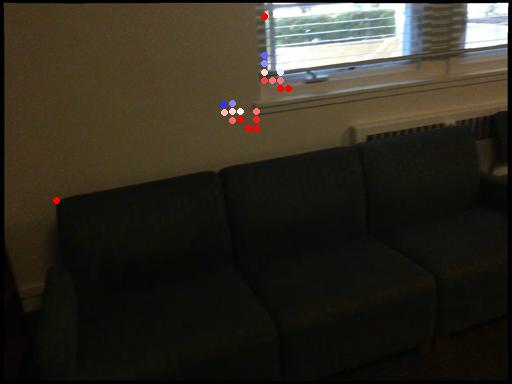}} \\
    \subfloat{\includegraphics[height=2.3cm]{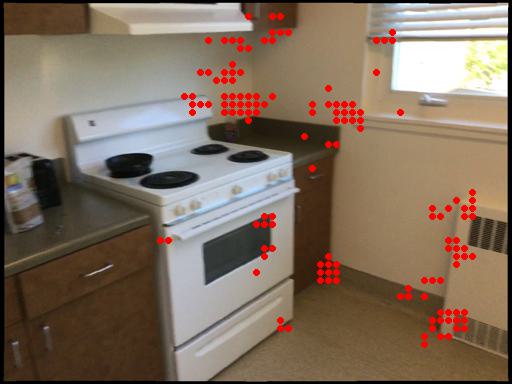}}\,
    \subfloat{\includegraphics[height=2.3cm]{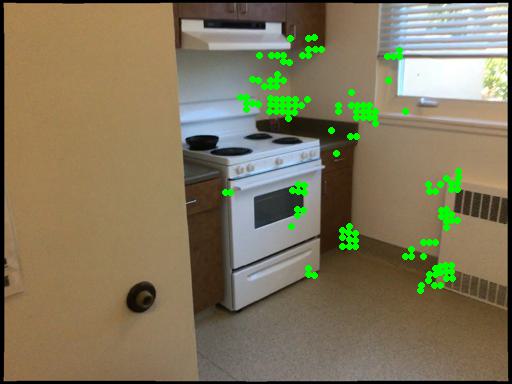}}\,
    \subfloat{\includegraphics[height=2.3cm]{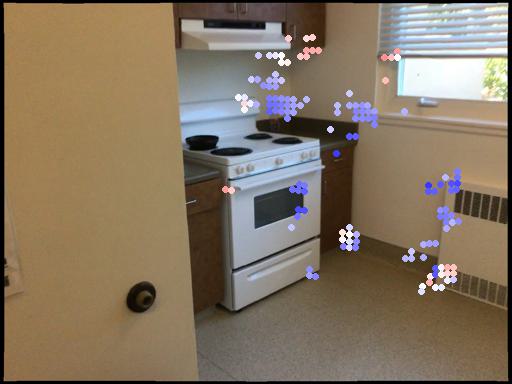}}\,
    \subfloat{\includegraphics[height=2.3cm]{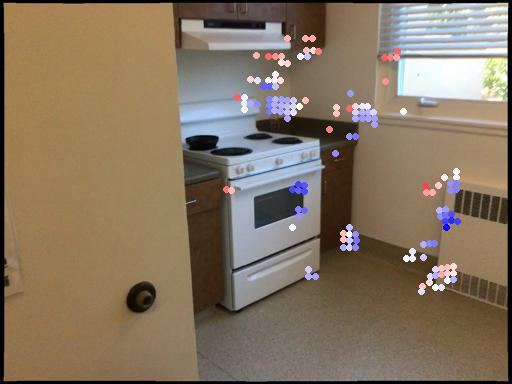}}\,
    \subfloat{\includegraphics[height=2.3cm]{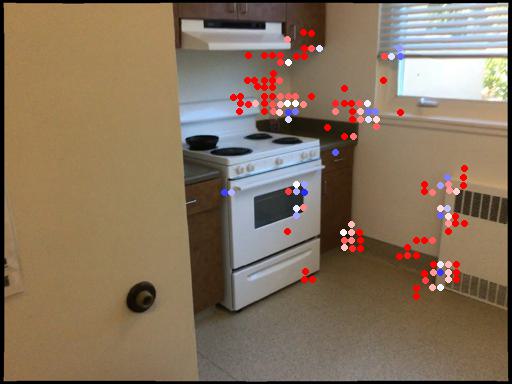}} \\
    \subfloat{\includegraphics[height=2.3cm]{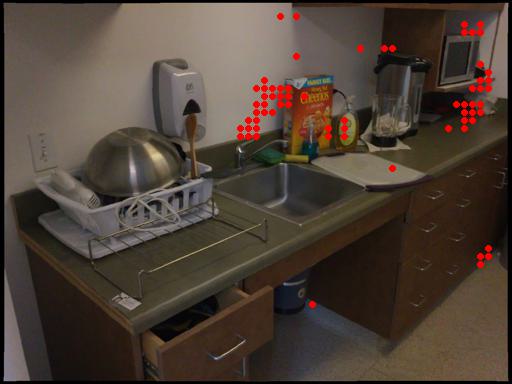}}\,
    \subfloat{\includegraphics[height=2.3cm]{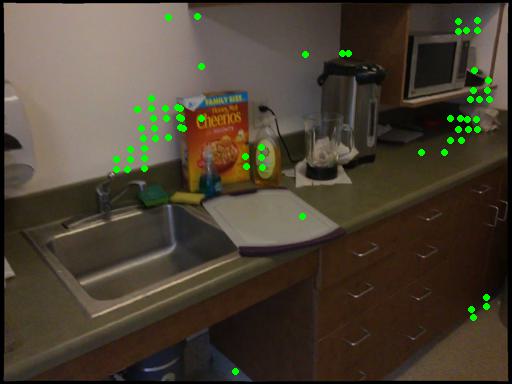}}\,
    \subfloat{\includegraphics[height=2.3cm]{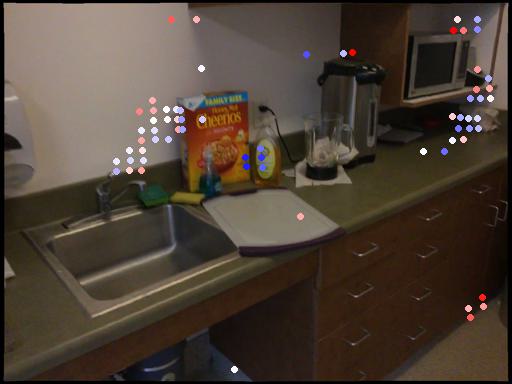}}\,
    \subfloat{\includegraphics[height=2.3cm]{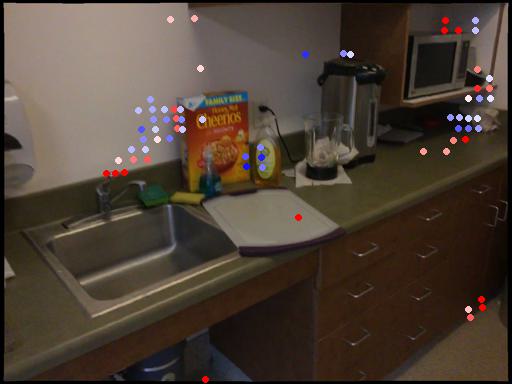}}\,
    \subfloat{\includegraphics[height=2.3cm]{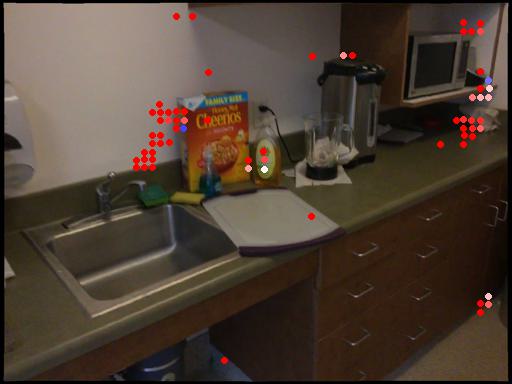}} \\
    \subfloat{\includegraphics[height=2.3cm]{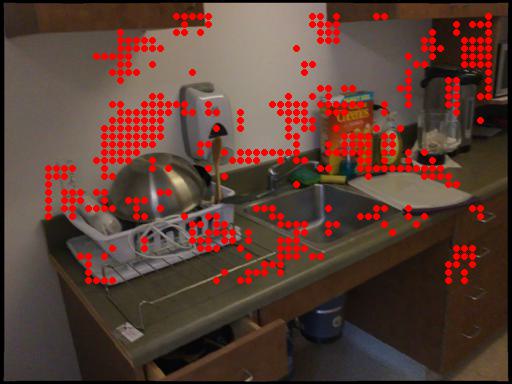}}\,
    \subfloat{\includegraphics[height=2.3cm]{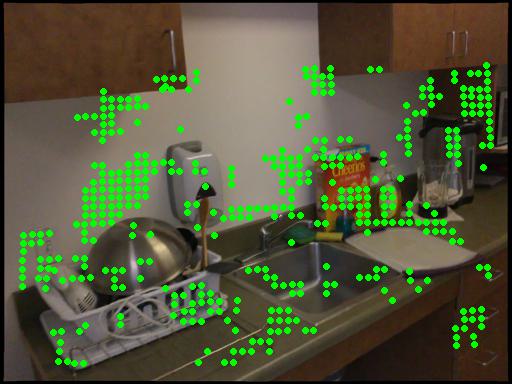}}\,
    \subfloat{\includegraphics[height=2.3cm]{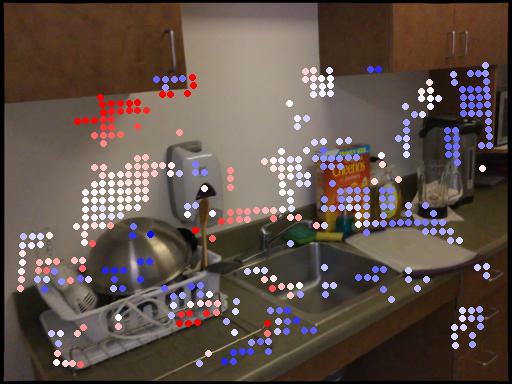}}\,
    \subfloat{\includegraphics[height=2.3cm]{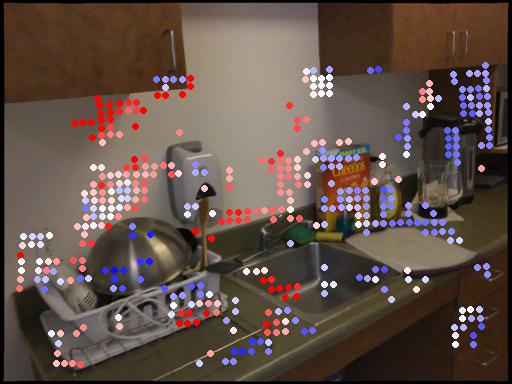}}\,
    \subfloat{\includegraphics[height=2.3cm]{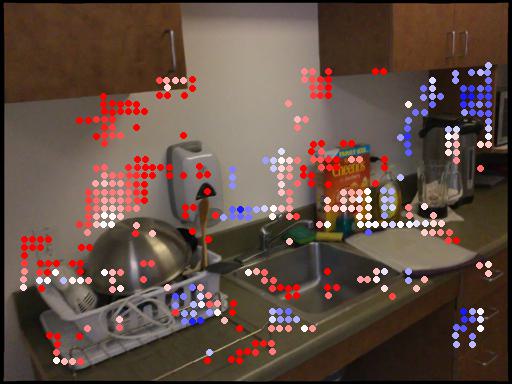}} \\
    \subfloat{\includegraphics[height=2.3cm]{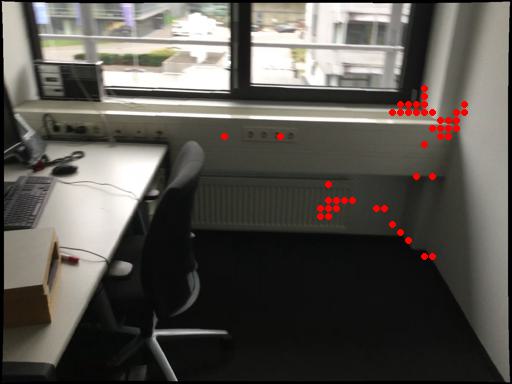}}\,
    \subfloat{\includegraphics[height=2.3cm]{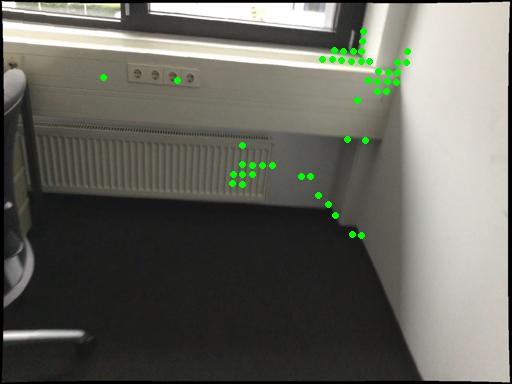}}\,
    \subfloat{\includegraphics[height=2.3cm]{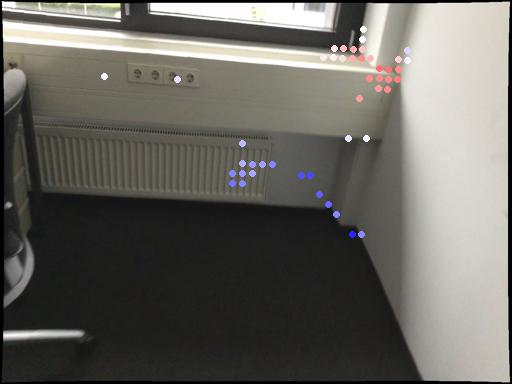}}\,
    \subfloat{\includegraphics[height=2.3cm]{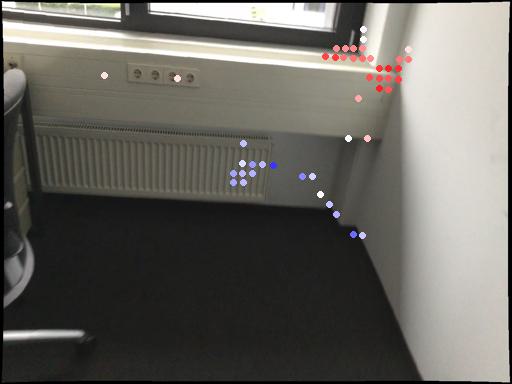}}\,
    \subfloat{\includegraphics[height=2.3cm]{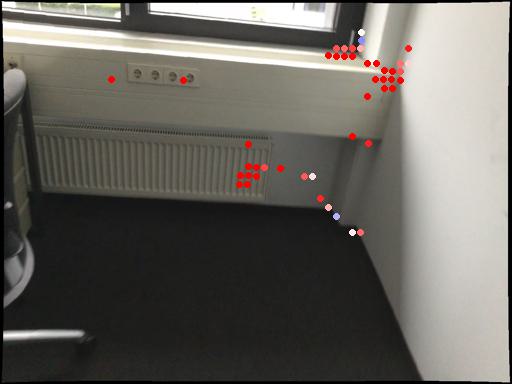}} \\
    \subfloat{\includegraphics[height=2.3cm]{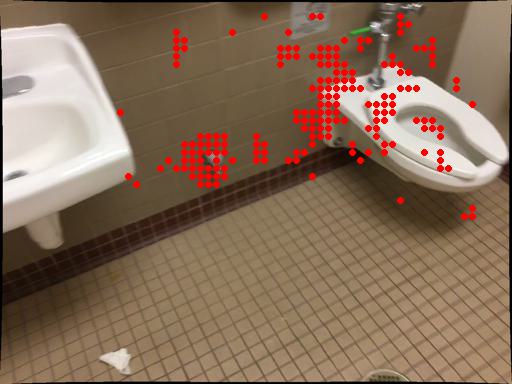}}\,
    \subfloat{\includegraphics[height=2.3cm]{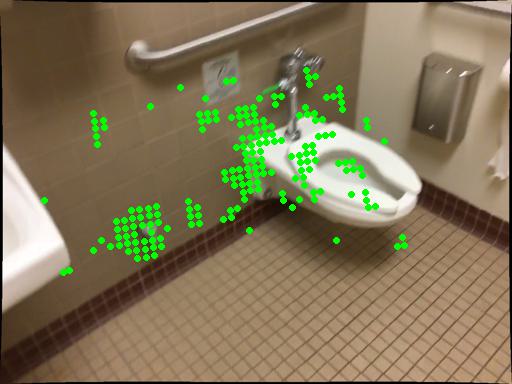}}\,
    \subfloat{\includegraphics[height=2.3cm]{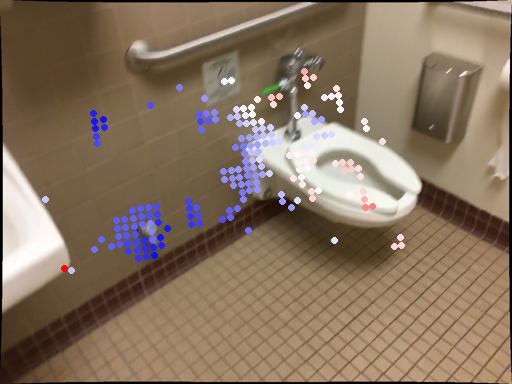}}\,
    \subfloat{\includegraphics[height=2.3cm]{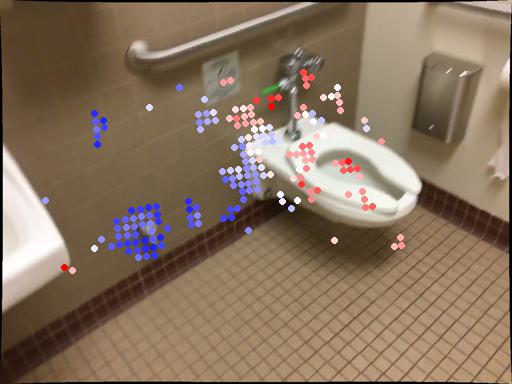}}\,
    \subfloat{\includegraphics[height=2.3cm]{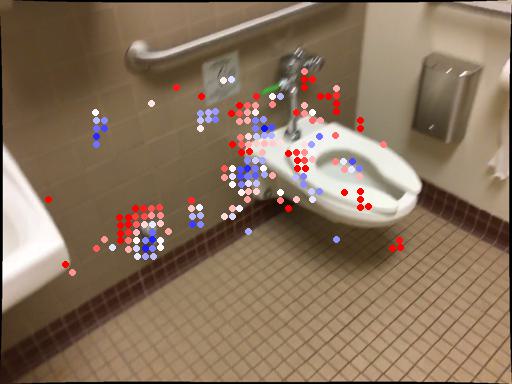}} \\
    \subfloat[(a) Source Frame $\mathbf{I}_1$]{\includegraphics[height=2.3cm]{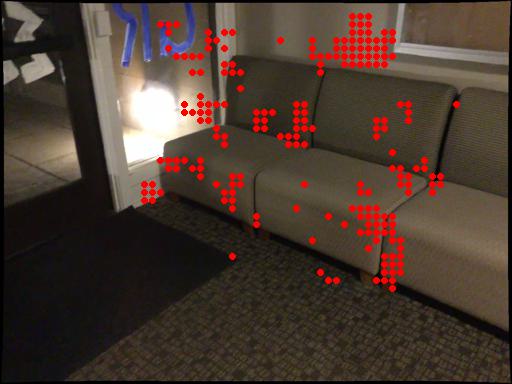}}\,
    \subfloat[(b) Support Frame $\mathbf{I}_2$]{\includegraphics[height=2.3cm]{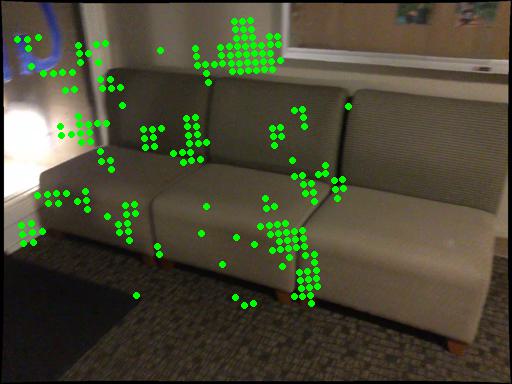}}\,
    \subfloat[(c) PMatch (Ours)]{\includegraphics[height=2.3cm]{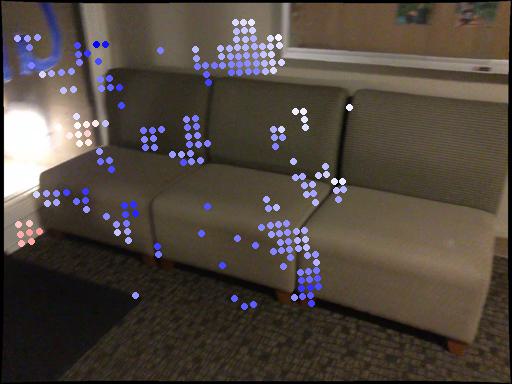}}\,
    \subfloat[(d) DKM~\cite{edstedt2022deep}]{\includegraphics[height=2.3cm]{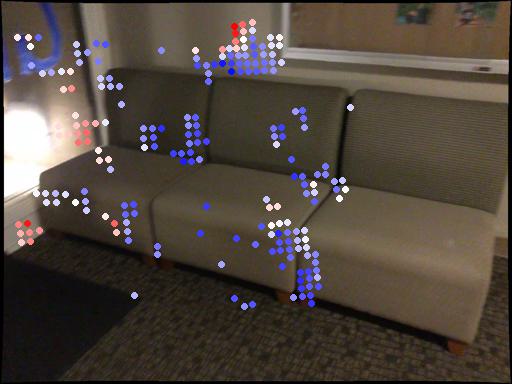}}\,
    \subfloat[(e) LoFTR~\cite{sun2021loftr}]{\includegraphics[height=2.3cm]{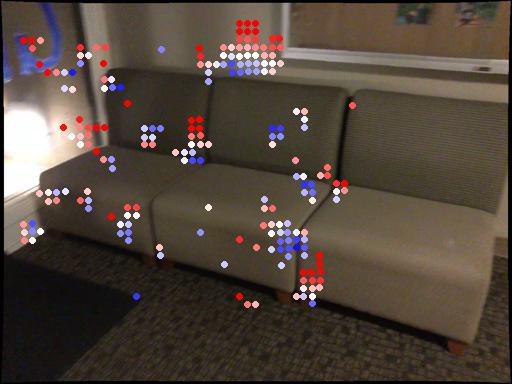}} \\
    \vspace{-1mm}
    \caption{\small 
    \textbf{Visual Comparisons on ScanNet.}
    We conduct the visual comparison against the SoTA dense~\cite{edstedt2022deep} and sparse~\cite{sun2021loftr} methods on the MegaDepth and the ScanNet datasets.
    The color from \textcolor{blue}{blue} to \textcolor{red}{red} indicates an increment in the end-point-error (L2 error).
    }
    \vspace{-2mm}
    \label{fig:visual_cp2}
\end{figure*}

\Paragraph{Additional Visualization.}
We include additional visualization of the paired MIM pretexting task (Fig.~\ref{fig:pretext_vls}), reconstruction visualization (Fig.~\ref{fig:reconstruction1} and Fig.~\ref{fig:reconstruction2}), and visual comparison with other SoTA methods (Fig.~\ref{fig:visual_cp1} and Fig.~\ref{fig:visual_cp2}).

\Paragraph{Negative Societal Impact.} The model can be deployed to malicious odometry system for behavior monitoring.

\begin{table*}[htb!]
\centering
\captionsetup{font=small}
\resizebox{0.6 \linewidth}{!}{
\begin{tabular}[width=\linewidth]{lccc}
\hline
\multicolumn{1}{c}{Ablation} & \multicolumn{3}{l}{Pose Estimation AUC $\uparrow$} \\
& @$5^\circ$           & @$10^\circ$          & @$20^\circ$ \\
\hline
baseline & $51.3$  & $67.5$  & $79.7$  \\
baseline + CFGM & $52.5$ & $68.5$ & $80.7$ \\
baseline + CFGM + $L_H$ & $53.3$  & $69.1$ & $81.1$ \\
baseline + CFGM + $L_H$ + MIM Encoder & $54.2$  & $70.2$ & $\mathbf{82.0}$ \\
baseline + CFGM + $L_H$ + paired MIM Encoder + Decoder & $\mathbf{54.9}$  & $\mathbf{70.6}$ & $\mathbf{82.0}$ \\
\hline
\end{tabular}
}
\vspace{-2mm}
\caption{\small \textbf{Additional Ablation Studies.} 
We extend the Tab.~$6$ in the main paper.
We compare the dense matching performance between pretraining with the MIM and the paired MIM.
In MIM pretraining, we follow the SimMIM~\cite{xie2022simmim}, \textit{i.e.}, pretraining the encoder (the $E_\theta$ and $T_\theta$) and a light decoder.
In the paired MIM pretraining, we include both the encoder and decoder, that is, $E_\theta$, $T_\theta$, $L_\theta$ and, part $R_\theta'$.
The corresponding network component is illustrated in the main paper Fig.~$2$.
The ablation is conducted under a training and testing resolution of $240 \times 480$ on the MegaDepth Dataset.
\label{tab:ablation}
}
\vspace{0mm}
\end{table*}

{\small
\bibliographystyle{ieee_fullname}
\bibliography{egbib} 
}